\documentclass[twoside,11pt]{article}

\usepackage{blindtext}

\usepackage{graphicx}
\usepackage{booktabs}
\usepackage{siunitx}
\usepackage{listings}
\usepackage{amssymb}  
\usepackage{amsmath}
\usepackage{mathtools}
\usepackage{color}
\usepackage{colortbl}
\definecolor{headergray}{gray}{0.88}
\usepackage{multirow}
\usepackage{array}
\usepackage{subcaption}
\usepackage{microtype}
\usepackage{graphics}
 \usepackage[most]{tcolorbox}
\usepackage{lipsum}
\usepackage{pifont}
\usepackage{makecell}
\usepackage[super]{nth}
\usepackage{wasysym}
\usepackage{float}
\usepackage{resizegather}
\usepackage{adjustbox}
\usepackage{multirow}
\usepackage{bm}
\usepackage{adjustbox}
\usepackage{tabularray}
\usepackage{wrapfig}
\usepackage{longtable}

\usepackage{xcolor}

\definecolor{codebg}{rgb}{0.97,0.97,0.97}
\definecolor{codekey}{rgb}{0.13,0.29,0.53}
\definecolor{codecomment}{rgb}{0.50,0.50,0.50}
\definecolor{codestring}{rgb}{0.64,0.08,0.08}

\lstdefinelanguage{yaml}{
  keywords={true,false,null},
  keywordstyle=\color{codekey}\bfseries,
  comment=[l]{\#},
  commentstyle=\color{codecomment}\itshape,
  morestring=[b]",
  stringstyle=\color{codestring},
}

\lstdefinestyle{yamlstyle}{
  language=yaml,
  backgroundcolor=\color{codebg},
  basicstyle=\ttfamily\small,
  frame=single,
  rulecolor=\color{black!30},
  framesep=6pt,
  numbers=left,
  numberstyle=\tiny\color{codecomment},
  numbersep=8pt,
  showstringspaces=false,
  breaklines=true,
  tabsize=2,
}
\usepackage{algorithm}
\usepackage{algorithmic}

\usepackage[preprint]{jmlr2e}

\usepackage{lastpage}
\jmlrheading{23}{2022}{1-\pageref{LastPage}}{1/21; Revised 5/22}{9/22}{21-0000}{Author One and Author Two}

\ShortHeadings{Systems-Aware Scaling for Sparse Mixture-of-Experts}{Systems-Aware Scaling for Sparse Mixture-of-Experts}
\firstpageno{1}

\begin{document}

\title{Compute-Optimal Is Not Cluster-Optimal: Systems-Aware Scaling for Sparse Mixture-of-Experts}

\author{\name Soumajyoti Sarkar \email soumajs@amazon.com \\
       \addr Amazon AGI Foundations \\
       \AND
       \name Yuxin Tang \email tanyuxin@amazon.com \\
       \addr Amazon AGI Foundations\\
       \AND
       \name Sheng Zha \email zhasheng@amazon.com \\
       \addr Amazon AGI Foundations\\
       }

\editor{}

\maketitle
\thispagestyle{empty}

\begin{abstract}
In large-scale pretraining, the algorithm, architecture, and systems decisions are conventionally made in disconnected stages. A scaling law stage selects an architecture and training recipe, optimizing loss under compute constraints, and a separate systems stage then optimizes the implementation for hardware efficiency. In this work, we develop MOSAIC, which formulates model architecture and systems co-design as an optimization problem. MOSAIC couples a predictive scaling law with a calibrated performance model that estimates Model FLOPs Utilization (MFU), communication cost, memory footprint, and the best parallel layout. We instantiate the framework for sparse Mixture-of-Experts (MoE) language models, where expert count, routing sparsity, and other MoE layer dimensions affect both the loss and systems efficiency. We fit a scaling law on sparse MoE models trained on text data, whose scaling dimensions include the sparsity factor, which is the fraction of model parameters inactive per token in a forward pass. The scaling law sweeps in our work span active parameters from $104$ million to $2.7$ billion and total model sizes reaching $79$ billion parameters. We show that, within the calibrated sparsity range, an efficiency-agnostic model-FLOPs budget admits no interior optimal sparsity. The fitted loss decreases monotonically with sparser models and the compute optimum lies at the upper boundary of the data support. An optimal sparsity in MoE models instead emerges under the cluster's systems constraints, as captured by MOSAIC. Our results argue for a shift towards unified architecture and systems co-design for frontier language model training.
\end{abstract}


%
\section{Introduction}
\label{sec:introduction}

Compute-optimal scaling laws~\citep{kaplan2020scaling,hoffmann2022training}
are the dominant tool for translating a fixed compute budget into a model
size and a data count in pretraining. They are the standard recipe in practice
because, once a training compute budget is set, the architecture (model
size $N$) and data (token count $D$) decisions both follow from a
single optimization against the predicted-loss curve, and because the
curve extrapolates far past the regime in which it was fit, making it
a predictable tool even at frontier
scale~\citep{bahri2024explaining,hestness2017deep}. For dense models, \citet{hoffmann2022training} devise a joint scaling law
for loss in the model size $N$ and the number of training tokens $D$,
which admits a closed-form model FLOPs\footnote{We use the term model FLOPs to denote ``compute" as is defined in the view of scaling laws and explicitly use the term deliverable model FLOPs to denote the hardware-delivered FLOPs in training.} optimal allocation. More recent prescriptive scaling laws recommend the optimal scaling dimensions under additional data constraints \citep{lovelace2026prescriptive}, yet none of these constraints price in the systems efficiency of the prescribed architecture. This matters because two architectures of the same predicted quality can realize training throughput that differs several-fold, once their attainable Model FLOPs Utilization (MFU)\footnote{MFU is the fraction of a device's peak throughput that a training run actually sustains. \url{https://cloud.google.com/blog/products/compute/using-cloud-tpu-multislice-to-scale-ai-workloads}} and best parallel layout are accounted for \citep{rajbhandari2022deepspeed}.

What follows is that the systems-side recommendations for architectures have largely evaded the scaling laws, taking the model size as given and optimizing the architecture at that size for hardware efficiency. As an example, \citet{anthony2024codesign}
show that a transformer at a given scale can be made more efficient on a specific accelerator by
aligning its intrinsic dimensions, such as the hidden size, attention-head
count, and vocabulary size, to the structure of the GPU compute kernels,
improving throughput at matched accuracy. A
complementary line of work as FlashAttention~\citep{dao2022flashattention} rearranges the operation arithmetically equivalently and
computes the same attention through an IO-aware, memory-efficient
schedule. Thus, the training recipe is
conventionally built in two disconnected stages, a scaling law stage for the model and data, and a hardware aware tuning stage, as in the Kimi-K2 report~\citep{team2025kimi} that treats sparse MoE scaling laws and systems tuning separately. This separation is consequential at scale, where the model is split across a fleet of GPUs through data, tensor, pipeline, and expert parallelism, as in MegaScale~\citep{jiang2024megascale}. The architecture chosen by a scaling law should account for these systems efficiencies rather than model-FLOPs alone \citep{bian2025inference}.

\begin{figure}[t!]
\centering
\begin{subfigure}{\textwidth}
  \centering
  \includegraphics[width=0.75\textwidth]{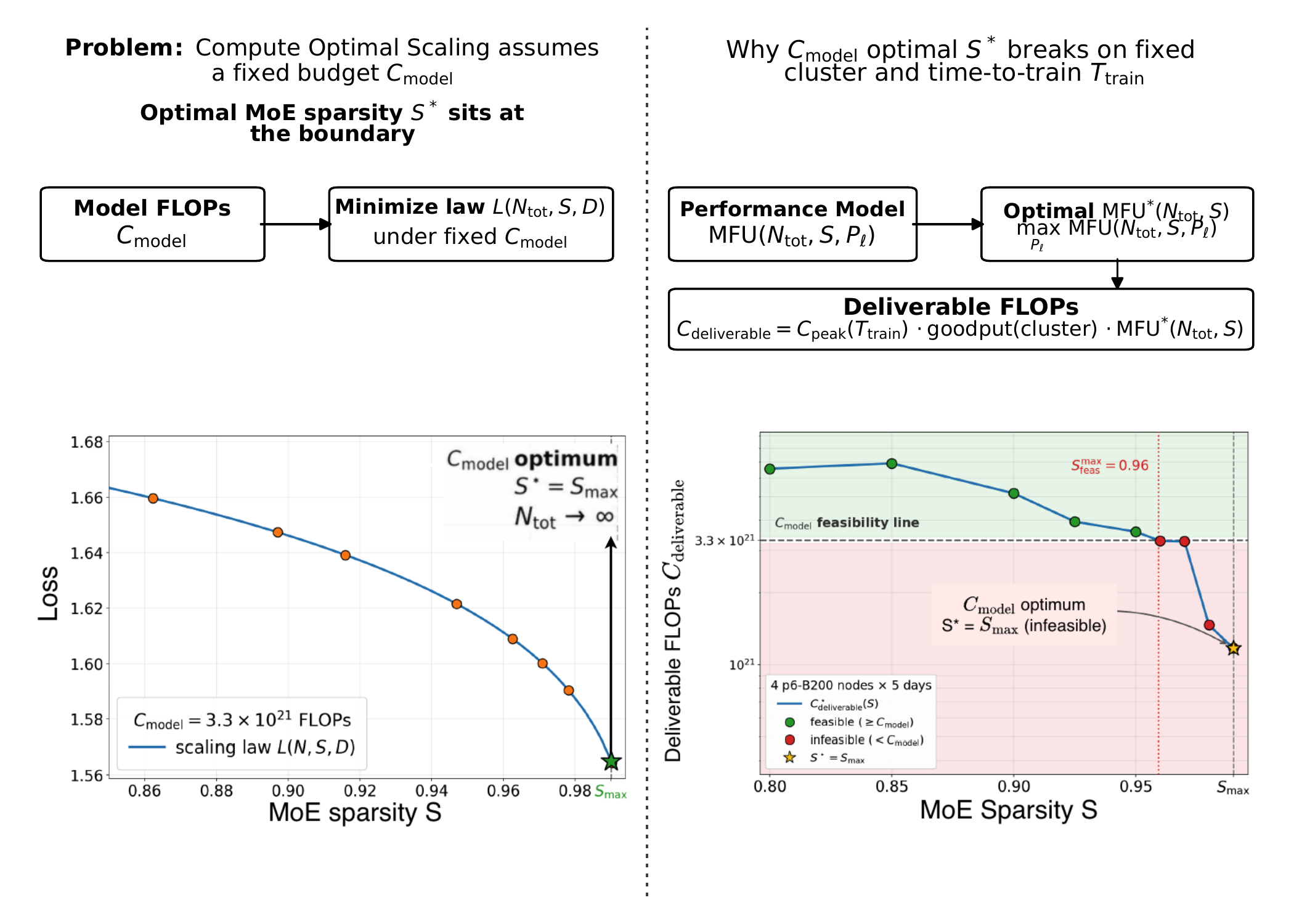}
    \subcaption{\textbf{Model FLOPs selects a boundary MoE sparsity.}
    Under a fixed 
    $C_{\mathrm{model}}=3.3\times10^{21}$, the fitted loss decreases over the
    calibrated sparsity range, so
    $S^{\star}=S_{\max}$
    (left). We do not cap any model dimension, and at each sparsity till $S_{\max}$ (0.985 here), the expert count and activated parameters  are optimized under the budget. Under a hardware budget $C_\text{peak}$ with 4 AWS p6-B200 nodes for 5 days, this
    prescription is infeasible because configurations beyond the feasible
    boundary of $S$=0.96 deliver fewer $C_\text{deliverable}$ FLOPs
    (\textcolor{red}{red region}) than what the fixed prescription $S^{\star}$
    requires (right).}
  \label{fig:mosaic-motivation}
\end{subfigure}

\vspace{0.8em}

\begin{subfigure}{\textwidth}
  \centering
  \includegraphics[width=0.52\textwidth]{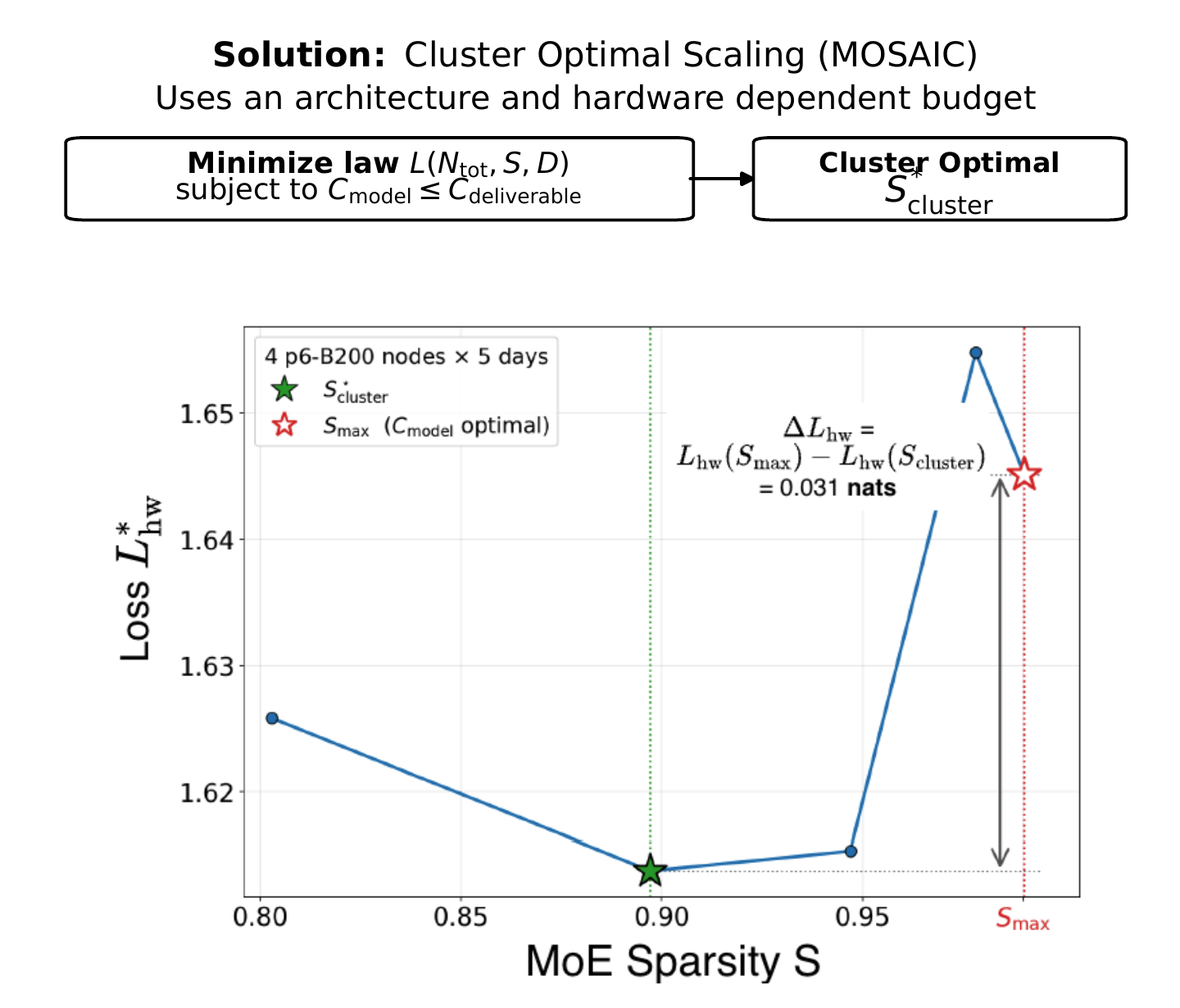}
  \subcaption{\textbf{MOSAIC yields an interior feasible sparsity under the hardware budget.}
The hardware-aware loss $\mathcal{L}^{\star}_{\mathrm{hw}}(S)$ induced by $C_{\mathrm{deliverable}}$ is minimized at the interior sparsity $\textcolor{green!55!black}{S^{\star}_{\mathrm{cluster}}}$ and is 0.031 nats lower than at $\textcolor{red}{S^{\star}=S_{\max}}$.
Each sparsity configuration is evaluated at the token count supported by its own attainable peak MFU under the same 4-node, 5-day hardware envelope.}
  \label{fig:mosaic-solution}
\end{subfigure}
\caption{The systems-aware scaling motivation for MOSAIC.}
\label{fig:mosaic}
\end{figure}

Our goal in this paper is to fold the systems stage into the scaling law stage, so that a single optimization selects the architecture, the token budget, and the cluster parallelism execution layout together, and we instantiate that framework for sparse MoE language models. We focus on sparse MoE architectures adopted previously for language model training \citep{fedus2022switch, sarkar2024revisiting, abnar2025parameters}, where sparsity is obtained by replacing the dense
feed-forward layer of a transformer block with a set of $E$ experts and
routing each token to a subset of them. This
decouples the active parameters $N_{\text{act}}$ that set per-token
model-FLOPs from the total stored parameters $N_{\text{tot}}$, with the expert count $E$ as
one of the knobs controlling total parameter count.  The existing scaling laws for sparse MoE models~\citep{clark2022unified, tian2026towards, krajewski2024scaling}
optimally scale these architecture knobs while ignoring systems efficiency, including \citet{abnar2025parameters}, who show that the sparsity factor $S$, roughly the proportion of  $N_{\text{tot}}$ that is inactive per token, introduces an additional optimum that classical $(N, D)$-only laws cannot see.  We show in  Section~\ref{sec:boundary-optimum} and Figure~\ref{fig:mosaic} that a model-FLOPs budget alone is insufficient to
determine a meaningful MoE sparsity.

A few recent studies begin to surface these systems-side pressures within the MoE scaling law itself. \citet{ludziejewski2025joint}
fit a sparse MoE scaling law and find that scaling
the number of experts improves loss at fixed model-FLOPs per token but inflates stored
parameters and device-memory pressure in a way that a model FLOPs budget
never charges for, so the practically optimal $E$ is where the loss
surface meets a memory-aware budget rather than the $E$ that minimizes
loss at fixed $N_{\text{act}}$.   \citet{wan2026holistic} push this
further, treating total parameters and model-FLOPs per token as independent
degrees of freedom and arguing that model-FLOPs per token alone is an
inadequate fairness metric for choosing a sparse MoE architecture. 

Building on these studies, we fit a new scaling law whose variables are chosen to expose systems efficiency. Conventionally, the optimal sparse MoE geometry under a model-FLOPs budget is derived in two stages, a geometry-free convex program for the target parameters, sparsity, and data budget, followed by a match to the nearest discrete architecture geometry. Our scaling analysis shows that sparser MoE models reach lower loss at any
model-FLOPs budget with the optimal sparsity sitting at the boundary, which forces the binding constraint to move from
model FLOPs to hardware-deliverable model FLOPs, the useful compute a cluster
can actually deliver over the training window. We  
solve a single-step discrete-geometry mixed-integer nonlinear program, which we term MOSAIC. It minimizes the predicted loss directly over the discrete architecture grid, resolving the boundary sparsity problem and realizing a cluster-specific sparse MoE geometry as motivated in Figure~\ref{fig:mosaic}.


We focus this work on the training side of sparse MoE architecture-system co-design, leaving test time scaling laws under hardware costs \citep{sadhukhan2026kinetics} as future work. The closest to our study is the work done in \citet{sun2026hardware} which couple a roofline
analysis to architecture selection. However, roofline alone misses the
overheads that flip MoE recommendations, namely expert dispatch and
combine, kernel-launch overhead under fine-grained experts, all-to-all
bandwidth as a function of EP topology, and data-parallel and
pipeline-parallel coupled exposed collective communications \citep{gale2023megablocks, jiang2024megascale}.

\paragraph{Contributions.}

To run the workflow of MOSAIC and obtain the optimal sparse MoE scaling
dimensions under the constraint of deliverable model FLOPs, we discuss the following contributions and observations:
\begin{itemize}
\item We fit a new four-dimensional joint MoE pretraining scaling law that expose the systems-side knobs. Through our scaling analysis, we show that across model-FLOPs budgets and within the calibrated sparsity range, the fitted loss optimal sparsity lies at the
upper boundary of the data support. This boundary-seeking behavior motivates
the move from a model-FLOPs budget to a deliverable model-FLOPs budget.
\item We develop an operator-level analytical performance model that measures
the per-block forward and backward cost of an MoE transformer. We validate the performance model against training run grids up to 18 billion active parameters, where the mean
absolute percentage error of predicted MFU stays under $15\%$ per sweep.
\item We couple these two components inside MOSAIC and search the
architecture grid on a cluster of NVIDIA B200 GPUs under a fixed
GPU hour budget. We show that the lowest-loss model configuration is not the one
that emits the most model FLOPs. We validate the
ranked selections from MOSAIC optimization, on staged pretraining runs of up to $250$B total parameters on our infrastructure. We show that the loss ordering flips from the model-FLOPs axis
to the peak-equivalent hardware-compute axis, as MOSAIC predicts.
\end{itemize}

We start with describing the abstract formulation of MOSAIC in Section~\ref{sec:mosaic}. To apply MOSAIC with sparse MoE model optimization, we first provide a technical overview of the training recipe used in our runs in Section~\ref{sec:preliminaries}. We then describe the components that instantiate MOSAIC with the sparse MoE architecture: (1) sparse MoE scaling laws in Section~\ref{sec:scaling-law-design} and, (2) the performance model in Section~\ref{sec:performance-model}. We make the application of MOSAIC for sparse MoEs concrete in Section~\ref{sec:hardware-codesign}, where the scaling law and the performance model come together to show why MoE sparsity has no interior optimum under a model FLOPs budget and why the right budget for a sparse MoE is the hardware-deliverable FLOPs. Finally, we discuss the experiments and the results in Section~\ref{sec:results}. We provide a snapshot of the most shared symbols used in the rest of the paper in Table~\ref{tab:symbols}.


\begin{table}[!t]
\centering
\small
\setlength{\tabcolsep}{6pt}
\renewcommand{\arraystretch}{1.15}
\begin{tabular}{@{}p{0.14\textwidth}p{0.78\textwidth}@{}}
\toprule
\textbf{Symbol} & \textbf{Meaning}\\
\midrule
\multicolumn{2}{@{}l}{\textbf{Architecture}}\\
$L_{\text{layers}}$
  & Number of transformer blocks.\\
$d$
  & Residual-stream width ($d_{\text{model}}$).\\
$d_{\text{ff}}$
  & Hidden width of the dense FFN that a sparse MoE block replaces.\\
$d_{\text{expert}}$
  & Hidden width of a single expert.\\
$E$
  & Total number of experts per sparse MoE block.\\
$K$
  & Top-$K$ active experts per token.\\
$G$
  & expert split factor, $G = d_{\text{ff}}/d_{\text{expert}}$. Figure~\ref{fig:g-geometry} in Appendix~\ref{app:scaling-law-coeffs} visualizes the construction.\\
$N_{\text{act}}$ 
  & Active parameters per token (FLOPs-relevant).\\
$N_{\text{tot}}$ 
  & Total stored parameters of the model (model-capacity).\\
$S$
  & Sparsity, $S = 1 - N_{\text{act}}/N_{\text{tot}}$ (fraction of
    parameters not activated per token).\\
\midrule
\multicolumn{2}{@{}l}{\textbf{Routing and load balancing}}\\
$H\in\mathbb{R}^{T\times d}$
  & Pre-MoE hidden states for a batch of $T$ tokens (Algorithm~\ref{alg:routing}).\\
$\phi\in\mathbb{R}^E$
  & Per-expert routing bias (LFLB controller).\\
$\mu_b$
  & Bias-update rate of the LFLB controller.\\
$\rho$
  & Proposal-to-selection ratio for Gumbel routing, giving proposal size
    $M = \min(\rho K, E)$.\\
$s'_i(x)$, $s_i(x)$
  & Clean sigmoid router score and the selection score $s_i(x)=s'_i(x)+\phi_i$
    that adds the balancing bias (Section~\ref{sec:routing}).\\
$\mathcal{T}_K(x)$
  & Set of top-$K$ experts selected for token $x$.\\
\midrule
\multicolumn{2}{@{}l}{\textbf{MOSAIC decision variables}}\\
$Z$
  & Architecture decision variable of MOSAIC, a realizable sparse MoE geometry
    with induced scaling law dimensions $(N_{\text{tot}}, S, G)$
    (Section~\ref{sec:mosaic}).\\
$P_{\ell}$
  & Execution plan,
    $P_{\ell} = (P_{\text{TP}}, P_{\text{EP}}, P_{\text{CP}}, P_{\text{PP}}, P_{\text{DP}},
    B_{\text{micro}}, A_{\text{ckpt}})$, giving the parallelism degrees along tensor (TP),
    expert (EP), context (CP), pipeline (PP), and data parallel (DP) groups together with
    the micro-batch size and the activation-checkpointing mode.\\
$B$
  & Batch envelope, $B = (B_{\text{global}}, T_{\text{seq}})$, giving
    $n_{\text{mb}} = B_{\text{global}}/(P_{\text{DP}}\,B_{\text{micro}})$
    microbatches per optimizer step.\\
$D$
  & Number of training tokens.\\
$C_{\text{model}}$
  & Model FLOPs, $C_{\text{model}} = 6\,N(Z)\,D$, where the activated
    parameters $N(Z)$ are $N_{\text{act}}$ for sparse MoEs.\\
$T_{\text{train}}$
  & Wall-clock training window on the cluster.\\
$C_{\text{peak}}$
  & Raw hardware ceiling, $C_{\text{peak}} = F_{\text{peak}}\,N_{\text{dev}}\,T_{\text{train}}$,
    where $N_{\text{dev}}$ is the device count.\\
$F_{\text{peak}}$
  & Per-device peak throughput.\\
$C_{\text{deliverable}}$
  & Deliverable model FLOPs,
    $C_{\text{deliverable}}(Z, P_{\ell}, B) = C_{\text{peak}}\cdot\mathrm{MFU}(Z, P_{\ell}, B)\cdot\eta_{\text{good}}(N_{\text{dev}})$,
     MFU is defined in Equation~\eqref{eq:mfu} and $\eta_{\text{good}}$ is the goodput fraction.\\

\bottomrule
\end{tabular}
\caption{Symbols used in the paper.}
\label{tab:symbols}
\end{table}

\begin{figure}[t!]
\centering
\vspace{-0.4em}
\includegraphics[width=0.85\textwidth]{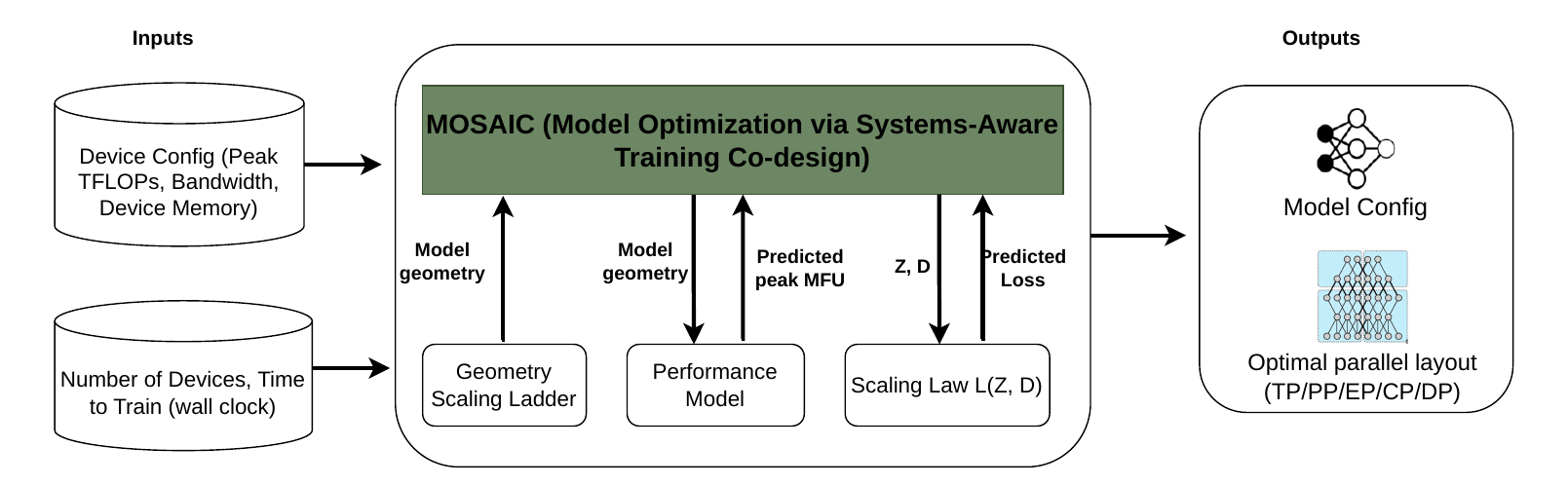}
\vspace{-0.6em}
\caption{The MOSAIC workflow. It takes the device configuration and the cluster
size and training window as inputs, queries the geometry scaling ladder for
candidate geometries, the performance model for predicted peak MFU, and the
scaling law for predicted loss, and returns the model configuration together with
its parallel layout.}
\vspace{-0.8em}
\label{fig:mosaic-workflow}
\end{figure}

\section{MOSAIC: \underline{M}odel \underline{O}ptimization via \underline{S}ystems-\underline{A}ware Tra\underline{I}ning \underline{C}o-design}
\label{sec:mosaic}

MOSAIC selects a model architecture, training-token budget, and distributed execution layout jointly for a fixed cluster and training window. Its central departure from conventional model FLOPs optimal scaling is that the available model-FLOPs budget is not treated as an architecture-independent constant. In Section~\ref{sec:why-flops-is-not-time}, we define the resulting architecture-dependent deliverable budget, we formulate the joint optimization in Section~\ref{sec:mosiac-formal}, and we instantiate it for sparse MoEs in Section~\ref{sec:mosaic-instantiation}. Figure~\ref{fig:mosaic-workflow} shows the workflow end to end, from the cluster inputs through the geometry ladder that fixes how the model dimensions co-scale, the performance model, and the scaling law, to the model configuration and parallel layout it returns.

\subsection{From Hardware Budget to Deliverable Model FLOPs}
\label{sec:why-flops-is-not-time}
A model FLOPs budget treats every allocation with the same
$C_{\text{model}}(Z, D) = 6\,N(Z)\,D$ as equally expensive, where $N(Z)$
counts the parameters the architecture activates per token (equals the full
parameter count for a dense model). On real clusters, model FLOPs omit a multiplicative systems-efficiency
factor that itself depends on the architecture and its execution plan. A cluster of $N_{\text{dev}}$ devices, each with peak per-device throughput $F_{\text{peak}}$, provides the raw hardware ceiling
$C_{\text{peak}} = F_{\text{peak}}\, N_{\text{dev}}\, T_{\text{train}}$ over a
training window
$T_{\text{train}}$. It is the physical maximum available over that window and only a
fraction of this ceiling becomes useful model computation. 

Before discussing how we define useful model computation, we define Model FLOPs Utilization (MFU) as is used in our work.  It is the fraction of the cluster's peak
throughput that a training run converts into the FLOPs mathematically
required by the model \citep{chowdhery2023palm}. An execution plan $P_{\ell}$
specifies how the model is sharded and its computation scheduled across the
GPU fleet \citep{meta2025llama3isca}. For architecture $Z$,
execution plan $P_{\ell}$, and batch envelope $B$, let
$t_{\text{iter}}(Z, P_{\ell}, B)$ denote the wall-clock time of one
optimizer step processing $B_{\text{global}}\, T_{\text{seq}}$ tokens, where
$T_{\text{seq}}$ is the sequence length. We define MFU as
\begin{equation}
\mathrm{MFU}(Z, P_{\ell}, B)
\;=\;
\frac{6\, N(Z) \; B_{\text{global}}\, T_{\text{seq}}}
     {t_{\text{iter}}(Z, P_{\ell}, B)\; N_{\text{dev}}\; F_{\text{peak}}}
\label{eq:mfu}
\end{equation}
It is the model FLOPs per token executed per second,
normalized by the peak FLOPs rate $F_{\text{peak}} N_{\text{dev}}$ the
cluster could sustain. The numerator counts only the FLOPs the model
mathematically requires, so activation recomputation, though executed on
the hardware, earns no credit. This distinguishes MFU from hardware FLOPs
utilization (HFU), and MFU is the metric that tracks time-to-loss.
MFU captures per-step execution efficiency, including kernel quality,
compute and communication overlap, pipeline bubbles, and expert-parallel
all-to-all. The goodput factor
$\eta_{\text{good}}(N_{\text{dev}})$ captures the fraction of allocated
wall-clock spent training rather than checkpointing, recovering from failures,
or waiting on the input pipeline. We  define the
deliverable model FLOPs as 
\begin{equation}
C_{\text{deliverable}}(Z, P_{\ell}, B)
\;=\;
C_{\text{peak}}
\;\cdot\;
\mathrm{MFU}(Z, P_{\ell}, B)
\;\cdot\;
\eta_{\text{good}}(N_{\text{dev}}).
\label{eq:deliverable-flops}
\end{equation}
Unlike a conventional compute
budget, $C_{\text{deliverable}}$ depends on the architecture and the layout $P_{\ell}$ used
to execute it. A candidate training configuration $(Z, D)$ is feasible to train
on the cluster within the window $T_{\text{train}}$ only if it
satisfies the constraint
\begin{equation}
C_{\text{model}}(Z, D)
\;\leq\;
\max_{P_{\ell} \in \mathcal{P}_{\text{feas}}(Z, B)}\, C_{\text{deliverable}}(Z, P_{\ell}, B)
\label{eq:feasible-budget}
\end{equation}
where $\mathcal{P}_{\text{feas}}(Z, B)$ collects the execution plans that fit the cluster constraint, defined in Section~\ref{sec:mosiac-formal}.

\subsection{Joint Architecture-Systems Optimization}
\label{sec:mosiac-formal}
MOSAIC formalizes hardware-aware model optimization as a mixed-integer
nonlinear program (MINLP) that couples the architecture to the systems layout it will execute under. It optimizes three coupled choices, the architecture $Z$, drawn from a
finite set $\mathcal{Z}_{\text{disc}}$ of realizable architectures, each
carrying the scaling law dimensions its geometry induces, the training tokens $D$,
and the execution plan $P_{\ell}$ of
Section~\ref{sec:why-flops-is-not-time}. The hardware inputs are
$(F_{\text{peak}}, N_{\text{dev}}, T_{\text{train}}, M_{\text{dev}})$, where
$M_{\text{dev}}$ is the per-device memory cap, and the
batch envelope $B = (B_{\text{global}}, T_{\text{seq}})$ fixes the tokens per
optimizer step. The execution plans feasible for an architecture are collected
in
\begin{equation}
\mathcal{P}_{\text{feas}}(Z, B)
\;=\;
\bigl\{\, P_{\ell} \in \mathcal{P}_{\text{disc}}(N_{\text{dev}})
\;:\;
\mathrm{Mem}(Z, P_{\ell}, B) \le M_{\text{dev}} \,\bigr\}
\label{eq:p-feas}
\end{equation}
where $\mathcal{P}_{\text{disc}}(N_{\text{dev}})$ is the finite set of
realizable execution plans the target training stack exposes on
$N_{\text{dev}}$ devices, comprising the parallelism degrees that tile the
cluster, the micro-batch size $B_{\text{micro}}$, and the
activation-checkpointing mode $A_{\text{ckpt}}$. Over this feasible plan
set, we solve the following optimization
\begin{tcolorbox}[ams align, colback=white, colframe=black, boxrule=0.4pt,
sharp corners, left=2pt, right=2pt, top=2pt, bottom=2pt]
\min_{Z \in \mathcal{Z}_{\text{disc}},\; D}\quad
& \mathcal{L}\bigl(Z, D\bigr)
\label{eq:hw-minlp}\\
\text{subject to}\quad
& C_{\text{model}}(Z, D)
\;\leq\;
\max_{P_{\ell} \in \mathcal{P}_{\text{feas}}(Z, B)}\,
C_{\text{deliverable}}(Z, P_{\ell}, B).
\tag{C-FLOPs}\label{eq:c-flops}
\end{tcolorbox}


The objective $\mathcal{L}$ is the scaling law, which predicts loss from the architecture specific
dimension $Z$ and the data $D$. The loss functional form evaluates $Z$ through the
scaling law dimensions it induces, whereas
$\mathrm{MFU}(.)$ and $\mathrm{Mem}(.)$ evaluate the concrete architecture itself, so all
three are well-defined functions of $Z$ even though several architectures can
share the same dimensions. The execution plan enters only through the inner
maximum, so the outer problem optimizes $(Z, D)$ alone. All systems
constraints sit inside $\mathcal{P}_{\text{feas}}(Z, B)$, which means that the maximum
never returns an infeasible plan.
Equation~\eqref{eq:c-flops} is the feasibility
condition of Equation~\eqref{eq:feasible-budget}. Equivalently, MOSAIC chooses for each architecture
the best feasible plan as
\begin{equation}
P_{\ell}^{\star}(Z)
\;=\;
\arg\max_{P_{\ell} \in \mathcal{P}_{\text{feas}}(Z, B)}\;
C_{\text{deliverable}}(Z, P_{\ell}, B).
\label{eq:best-layout}
\end{equation}

We note that Equation~\eqref{eq:hw-minlp} is a
bilevel non-convex mixed-integer nonlinear program, an optimization problem
that is bilevel because of the embedded
maximization over $\mathcal{P}_{\text{feas}}$ in Equation~\eqref{eq:c-flops}. It is mixed-integer nonlinear since the layout
degrees are integer and the architecture is drawn from the discrete set
$\mathcal{Z}_{\text{disc}}$. The constraint functions
$\mathrm{MFU}(\cdot)$, $\mathrm{Mem}(\cdot)$ are nonlinear, and non-convex
because of the discrete variables of the realizable plan set
$\mathcal{P}_{\text{disc}}(N_{\text{dev}})$ together with the non-convexity of MFU and memory in
$P_{\ell}$. We collect in Appendix~\ref{sec:minlp-structure} three structural
properties of this program, including a closed-form elimination of $D$, that
make its exact solution by enumeration tractable.

\subsection{Instantiating MOSAIC  for Sparse MoEs}
\label{sec:mosaic-instantiation}
We put Equation~\eqref{eq:hw-minlp} to be architecture-agnostic. Applying it requires
three components, a realizable architecture space induced by $Z$, a predictive loss
function $\mathcal{L}(Z, D)$, and predictors for $\mathrm{MFU}(Z, P_{\ell}, B)$
and $\mathrm{Mem}(Z, P_{\ell}, B)$. We define the sparse MoE realization of $Z$ in Section~\ref{sec:preliminaries}, under which $\mathcal{Z}_{\mathrm{disc}}$ becomes the finite set of realizable sparse MoE
geometries and the activated parameter count becomes $N(Z)=N_{\mathrm{act}}$. The execution plan additionally
includes expert parallelism, and in our Megatron-Core realization, expert
parallelism repartitions the data-parallel domain rather than introducing an
independent world-size factor. We give the
exact mesh factorization and divisibility constraints used to construct
$\mathcal{P}_{\text{disc}}(N_{\text{dev}})$ for our MoE models in Appendix~\ref{app:plan-realization}. We discuss each of these three
components applied to sparse MoEs in the following sections.
Figure~\ref{fig:mosaic} instantiates the constraint of Equation~\ref{eq:feasible-budget} with real numbers where
a $C_{\mathrm{model}}=3.3\times10^{21}$ prescription on 4 p6-B200 nodes
for 5 days becomes infeasible beyond $S_{\mathrm{feas}}^{\max}=0.96$.
Re-optimizing under the same hardware envelope yields an interior
cluster-optimal sparsity whose predicted loss is 0.031 nats lower than
the model FLOPs optimal boundary configuration.
These constraints let us treat the model dimensions as decision variables of
the optimization rather than quantities capped in advance, which we argue is
the more natural formulation for MoE-like models.
\section{Sparse MoE Pretraining: Scaling Recipe}
\label{sec:preliminaries}
In this section, we discuss the dimensions $Z$ for sparse MoEs in the context of scaling laws and describe the training recipe used to collect the scaling law data. We lay out the details of the architecture that is shared by all the runs in our study.
Our architecture and training recipe follows the broader work from existing sparse MoE scaling studies of~\citet{krajewski2024scaling} and~\citet{abnar2025parameters} with some adjustments, and we describe those differences in the following sections.

\subsection{Training Setup}
\label{sec:training-setup}

We experiment with decoder-only transformer language models in which every feed-forward layer is replaced by an MoE layer.   The first transformer block is kept dense and all remaining blocks
are sparse MoE layers. Following the architecture design in DeepSeek-V3
\citep{deepseekai2024v3,dai2024deepseekmoe}, every MoE layer in the transformer block carries one
shared expert that is always active in addition to the $K$
routed experts.  Each training run is optimized for
text-to-text with the next-token-prediction language modeling objective and all runs use bf16 weights and activations. We focus on the MoE layer in this section since that forms the basis of the scaling laws in the rest of the paper. We use the standard parametrization recipe for weight initializations and we adopt AdamW optimizer \citep{loshchilov2017decoupled} for all the runs. Similar to the Chinchilla training runs \citep{hoffmann2022training}, we consider a single pass over the tokens and we use the smoothed training loss of the final checkpoint of each run for the scaling law fits. The rest of the technical details of the data, training recipe and the hyper-parameter choices including the optimizer choices and learning rate scheduler are described in Appendix~\ref{sec:train_recipe}.

\subsection{MoE Layer}

We use $L_{\text{layers}}$ to denote the number of transformer blocks,
$d \!\equiv\! d_{\text{embd}}$ as the residual stream width, and
$d_{\text{ff}}$ as the hidden width of the corresponding dense feed-forward layer
that a sparse MoE layer replicates into $E$ copies \footnote{$d_{\text{ff}}$ is conventionally a fixed multiple of the residual width $d$. For example, GPT-2~\citep{radford2019language} uses $d_{\text{ff}} = 4d$.}. Each MoE layer consists of $E$ experts
$\{f_i\}_{i=1}^{E}$, where every expert is itself a two-layer feed-forward
network of the form
$f_i(x) = W^{(2)}_i \,\sigma\!\left(W^{(1)}_i x\right)$
with $W^{(1)}_i \in \mathbb{R}^{d_{\text{expert}}\times d}$,
$W^{(2)}_i \in \mathbb{R}^{d \times d_{\text{expert}}}$, and $\sigma(\cdot)$ a
gated activation,  in our runs being SwiGLU. We keep $d_{\text{ff}}$ and
$d_{\text{expert}}$ as separate symbols because a sparse MoE layer can
split the reference FFN of width $d_{\text{ff}}$ into narrower experts of
width $d_{\text{expert}}$, so the two coincide only when each expert is a
full-width dense FFN, something we will explore in Section~\ref{sec:scaling-law-design}. We follow the Switch-style token-choice routing
formulation~\citep{fedus2022switch,liu2023sparsemixer}. A learnable router
$W_r \in \mathbb{R}^{E \times d}$ produces a per-token score
vector $\boldsymbol s(x) \in \mathbb{R}^{E}$, and the top-$K$ experts under
$\boldsymbol s(x)$ are selected for each token in a forward pass. The sparse MoE layer output is the gating-weighted
sum of their outputs,
\begin{equation}
y(x) \;=\; \sum_{i \,\in\, \mathcal{T}_K(x)} \tilde\pi_i(x)\, f_i(x),
\qquad
\mathcal{T}_K(x) \;=\; \operatorname{Top}\!K\bigl(\boldsymbol s(x)\bigr)
\label{eq:moe-output}
\end{equation}
where $\tilde\pi_i(x)$ are the probabilities the selected set
$\mathcal{T}_K(x)$. In the following section, we describe the routing operations that result in the $\operatorname{Top}\!K(.)$ outputs and the associated probabilities $\tilde\pi_i(x)$ and we treat
Equation~\ref{eq:moe-output} as the canonical sparse MoE layer definition.


\begin{algorithm}[!t]
\caption{Sparse MoE token-choice routing for a single MoE layer, batch of $T$ tokens routed to $E$ experts}
\label{alg:routing}
\renewcommand{\algorithmicrequire}{\textbf{Input:}}
\renewcommand{\algorithmicensure}{\textbf{Output:}}
\begin{algorithmic}[1]
\REQUIRE Pre-MoE hidden states $H \in \mathbb{R}^{T \times d}$
         (attention output of the current block added to the previous-layer
         residual stream); router weights $W_r \in \mathbb{R}^{E \times d}$;
         per-expert bias $\phi \in \mathbb{R}^{E}$; hyperparameters
         $K$ (top-$K$),  Gumbel proposal size $M = \min(\rho K, E)$,
         Gumbel temperature $\tau$.
\vspace{0.1em}\noindent\rule{\linewidth}{0.4pt}\vspace{0.1em}
\ENSURE  Top-$K$ expert indices
         $\mathcal{T}_K \in \{1,\dots,E\}^{T \times K}$, where row
         $\mathcal{T}_K(x_t) = \operatorname{Top}\!K(\boldsymbol s(x_t))$ over the
         proposal set $\mathcal{C}_t$ is the
         selected set for token $t$ in Eq.~\ref{eq:moe-output}; normalized gate
         probabilities $\tilde\pi \in \mathbb{R}_{\geq 0}^{T \times K}$ with
         $\sum_{i \in \mathcal{T}_K(x_t)} \tilde\pi_{t,i} = 1$;
         per-expert token counts $n \in \mathbb{N}^{E}$. Throughout,
         $i \in \{1,\dots,E\}$ identifies an expert and $t \in \{1,\dots,T\}$ a token.
\vspace{0.1em}\noindent\rule{\linewidth}{0.4pt}\vspace{0.1em}
\STATE $\tilde W_r \gets \textsc{RowNorm}(W_r, c_r)$
       \COMMENT{$\ell_2$-normalize each router row to target norm $c_r$}
\STATE $\Theta \gets H\, \tilde W_r^{\!\top}$
       \COMMENT{router logits}
\STATE $s' \gets \sigma(\Theta)$
       \COMMENT{clean scores, no bias}
\STATE $s \gets s' + \phi$
       \COMMENT{LFLB bias-added selection scores, Eq.~\ref{eq:sigmoid-score}
       in Appendix~\ref{app:routing-details}}

\FOR{each token $t \in T$ (parallel)}
  \STATE $g_{t,i} \sim \mathrm{Gumbel}(0, 1)$ for $i = 1, \dots, E$
         \COMMENT{i.i.d.\ standard Gumbel per expert}
  \STATE $g_t \gets g_t - \frac{1}{T}\!\sum_{t'=1}^{T} g_{t'}$
         \COMMENT{per-expert noise centering}
  \STATE $\mathcal{C}_t \gets \operatorname*{Top-}\!M_{i}\bigl(\log s'_{t,i} + \tau\, g_{t,i}\bigr)$
         \COMMENT{proposal on clean scores}
  \STATE $\mathcal{T}_{K,t} \gets \operatorname*{Top-}\!K_{i \in \mathcal{C}_t}\bigl(s_{t,i}\bigr)$
          \COMMENT{bias-aware correction}
  \STATE $\zeta_t \gets \sum_{j \in \mathcal{T}_{K,t}} s'_{t,j}$
  \STATE $\tilde\pi_{t,i} \gets \frac{s'_{t,i}}{ \zeta_t},\quad i \in \mathcal{T}_{K,t}$
         \COMMENT{gates use clean scores, $\sum=1$}
\ENDFOR

\STATE $n_i \gets \sum_{t=1}^{T} \mathbb{1}\!\left[i \in \mathcal{T}_{K,t}\right],\quad i = 1, \dots, E$
       \COMMENT{entries of the per-expert token-count vector $n$}
\STATE \textbf{return} $\mathcal{T}_K,\ \tilde\pi,\ n$
\end{algorithmic}
\end{algorithm}
\subsection{Routing and Load Balancing}
\label{sec:routing}

Stable large-scale sparse MoE training hinges on the routing recipe that
selects the active experts per token, since imbalances in token-to-expert
assignment compound into load-balancing issues \citep{hu2025sigma,wang2024lossfree}, loss
spikes, and a degraded effective parameter count. We describe the routing choices
that kept training stable as we scaled up, a prerequisite for reliable scaling law
extrapolation. There have been myriad studies focused on designing such routing algorithms that improve both the training dynamics and the end model accuracy \citep{muqeeth2023soft, zoph2022st, zhou2022mixture, hu2025sigma}. We build on the
auxiliary-loss-free load balancing (LFLB) strategy of
\citet{wang2024lossfree}, in which a per-expert bias is updated online to
counteract observed load imbalance, but augment it with several modifications
that we found necessary for stable training across our scaling sweeps.

In its original form, LFLB scores each token-expert pair by a sigmoid
affinity $s'_i(x) = \sigma(\theta_i(x))$ with $i$ denoting an expert and $\theta_i(x) = \tilde{w}_{r,i}^{\!\top} x$ denoting
the router logit. It selects the top-$K$ experts by a bias-adjusted score
$s_i(x) = s'_i(x) + \phi_i$, and combines their outputs with the clean scores $s'_i(x)$
renormalized over the selected set~\citep{wang2024lossfree,deepseekai2024v3}.
The per-expert bias $\phi_i$  is a controller state
nudged once per training step toward under-used experts from the
observed per-expert token counts $n_i$, so the bias fixed at the end of one
step steers selection on the next. We keep this sigmoid-plus-bias score
(Eq.~\ref{eq:sigmoid-score} in Appendix~\ref{app:routing-details}) but adjust the
selection rule and the bias controller, as summarized next.

We introduce a few additions to the base LFLB recipe, laid out in
Algorithm~\ref{alg:routing} and detailed in
Appendix~\ref{app:routing-details}. The algorithm takes as input a batch
of $T$ tokens with their pre-MoE hidden states $H$, the router weights $W_r$, and the current
per-expert bias $\phi$, and returns the top-$K$ expert indices
$\mathcal{T}_K$ selected per token, their normalized gate probabilities
$\tilde\pi$, and the per-expert token counts $n$ that drive the bias
update for the next step. The main additions to LFLB are as follows.
\begin{itemize}
\item The expert
bias is updated online by an RMS-normalized imbalance controller and
re-centered each step, so the correction tracks the shape of the
imbalance rather than the batch size.
\item The top-$K$ selection involves a Gumbel proposal-correction sampler. It entails a stochastic top-$M$ proposal on the clean sigmoid scores followed by a bias-aware top-$K$ correction ($M > K$, with $M$ a pre-selected hyperparameter fixed across all runs), with the per-expert noise centered across the batch. The top-$M$ proposal followed by the top-$K$ correction separates exploration from final routing. We use Gumbel scores only to expand the candidate set to $M$, then choose the final $K$ using clean scores inside that set. Since applying Gumbel noise over all experts can cause large arbitrary flips, we restrict it to a top-$M$ candidate pool among $E$ experts. It lets near-boundary experts get exposure while preventing far-tail experts from winning due to noise. Lines 5\textendash9 of Algorithm~\ref{alg:routing} follow this procedure.
\item The router rows are
$\ell_2$-normalized every forward pass so routing depends on direction
rather than norm drift. Load balancing is thus carried by the bias controller and
the noise, with no batch-level load-balancing loss added to the objective. We do
retain the sequence-level auxiliary loss of \citet{deepseekai2024v3}, described
in Appendix~\ref{app:routing-details}.
\end{itemize}

Together, these changes provide a stable routing recipe across the sweep without a batch-level
load-balancing loss. That stability is what lets us treat
the final-checkpoint loss of every run as a comparable fit target for the scaling
laws of Section~\ref{sec:scaling-law-design}.

\subsection{Sparsity $S$ and Expert Split Factor $G$}

In considering the geometry of MoE models, the two architectural dimensions of interest are the total expert count $E$ and the per-token activated experts $K$. Together with $L_{\text{layers}}$, $d$, and
$d_{\text{ff}}$, these determine an active parameter budget
$N_{\text{act}}$ (the model FLOPs-relevant parameter count seen by any individual
token) and a total parameter budget $N_{\text{tot}}$ (the memory
footprint of the model). In this section, we discuss two architectural knobs, the sparsity $S$ and the expert split factor $G$, that we later use to parametrize the sparse MoE scaling laws. We recall from Table~\ref{tab:symbols} that the two parameter budgets set the sparsity $S = 1 - N_{\text{act}}/N_{\text{tot}}$, the fraction of parameters left inactive per token. To compare architectures whose experts differ in size, we use
\textbf{expert split factor} defined as:
\begin{equation}
G \;\;\coloneqq\;\; \frac{d_{\text{ff}}}{d_{\text{expert}}}
\label{eq:granularity}
\end{equation}

It is the ratio of the dense FFN width $d_{\text{ff}}$ that an MoE block
replaces to a single expert's hidden width $d_{\text{expert}}$,
measuring how finely that dense FFN has been partitioned into experts. We visually show the construction of this expert split factor $G$ in an MoE layer in Figure~\ref{fig:g-geometry} in Appendix~\ref{app:scaling-law-coeffs}. $d_{\text{ff}}$ is the intermediate (hidden) width of the
reference full FFN, a fixed multiple of the residual-stream width
$d_{\text{model}}$, and is distinct from $d_{\text{model}}$ itself and
$d_{\text{expert}} = \frac{d_{\text{ff}}}{G}$ is the per-expert slice of it.

The $G\!=\!1$ case recovers the 
setting studied in \cite{clark2022unified} in which each expert is itself
a full-width FFN. $G \!>\! 1$ corresponds to fine-grained experts whose width
shrinks proportionally as their count grows. Equivalently, at a fixed
total parameter count (or fixed total expert width), a higher $G$ means
smaller, but more finer-grained experts in the MoE layer. The ratio $\frac{E}{G}$ is the
effective expansion rate of total to dense FFN parameters used
by~\citet{krajewski2024scaling}, so increasing $G$ at fixed $\frac{E}{G}$
yields more, smaller experts at constant total parameter count. 

\section{ Sparse MoE Scaling Laws $\mathcal{L}(.)$}
\label{sec:scaling-law-design}

Building on the sparse MoE architecture discussed in Section~\ref{sec:preliminaries}, we elaborate on what $Z$ entails and we lay out the desiderata behind the scaling law formulations we target in our study. We want a loss law where  $Z$ meets three
requirements. First, a concrete discrete MoE geometry must map to the law's
dimensions, so a downstream optimizer can realize the fitted optimum as a
real architecture rather than an abstract parameter count, which is why
we parameterize in total parameters and geometry knobs. Second, those dimensional knobs must expose how the model geometry acts on the
systems side including communication, and memory and not on loss
alone. Third, we prefer that the law stay low-dimensional, because if each of the $u$ free dimensions in a scaling law is evaluated at $v$ discrete values using a full factorial design, the number of configurations scales as $O(v^u)$. The laws fitted in the study by
\citet{abnar2025parameters} and by \citet{tian2026towards} are on
essentially the same axes, differing mainly in the shared-expert ratio. We build on both while adding the geometry dimension the systems side
needs. We utilize the sparsity $S$ and expert split factor $G$ described in Section~\ref{sec:preliminaries}, and argue that $G$ is a necessary axis beyond those considered in \cite{abnar2025parameters}. We then review the \cite{abnar2025parameters} law on our training runs in Section~\ref{sec:law-with-sparsity}, following which we discuss the final scaling function form we use for the optimization in Section~\ref{sec:joint-NtSDG-law}.

\begin{figure}[!t]
\centering
\begin{subfigure}[t]{0.35\textwidth}
  \centering
  \includegraphics[width=\linewidth]{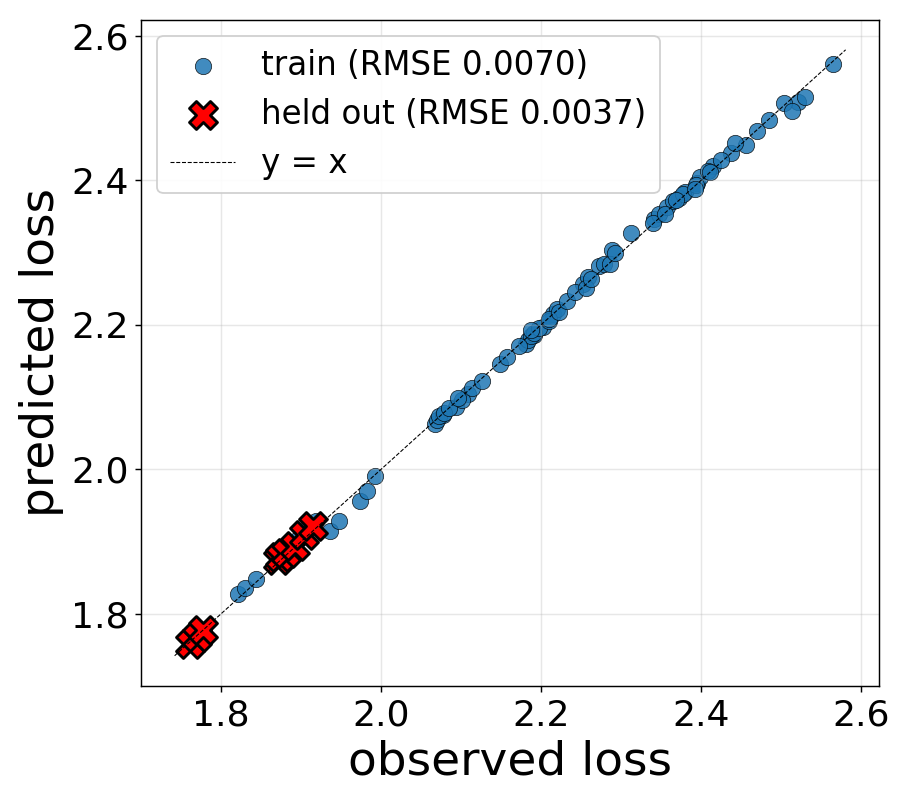}
  \subcaption{$G=4$.}
  \label{fig:parity-G4}
\end{subfigure}\hspace{0.03\textwidth}
\begin{subfigure}[t]{0.35\textwidth}
  \centering
  \includegraphics[width=\linewidth]{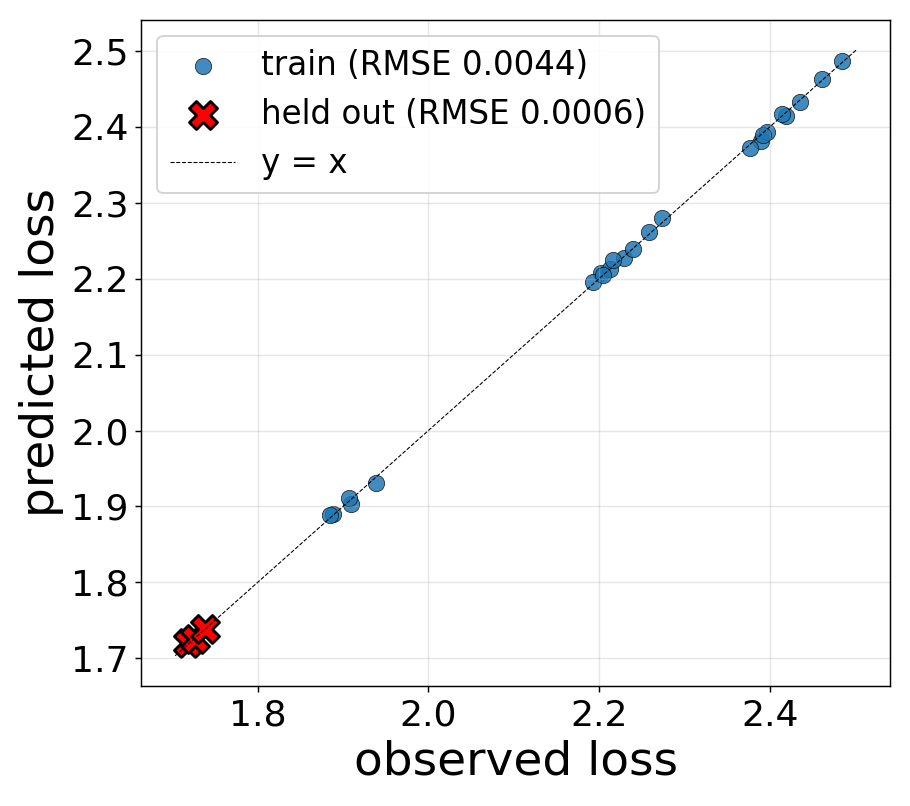}
  \subcaption{$G=8$.}
  \label{fig:parity-G8}
\end{subfigure}
\caption{Predicted-vs-observed training loss for the
$\mathcal{L}(N_{\text{tot}}, S, D)$ fit in Equation~\eqref{eq:abnar-law},
fit per expert split factor. The panels for every stratum we fit,
$G\!\in\!\{1, 2, 4, 8\}$, are in Figure~\ref{fig:per-G-parity-appendix} of
Appendix~\ref{app:scaling-law-coeffs}.}
\label{fig:per-G-parity}
\end{figure}

\subsection{Scaling Laws with Sparsity}
\label{sec:law-with-sparsity}

Before fitting the functional form following the desiderata above, we follow the functional form studied in ~\citet{abnar2025parameters}, and parameterize per-token training
loss as a separable Chinchilla-style law in total parameters and tokens, with
an additional sparsity term, so the induced scaling law dimensions are
$(N_{\text{tot}}, S)$,
\begin{equation}
\mathcal{L}(N_{\text{tot}}, S, D)
\;=\;
\underbrace{\frac{a}{N_{\text{tot}}^{\alpha}}}_{\text{capacity}}
\;+\;
\underbrace{\frac{b}{D^{\beta}}}_{\text{data}}
\;+\;
\underbrace{\frac{c}{(1-S)^{\lambda}}}_{\text{sparsity floor}}
\;+\;
\underbrace{\frac{j}{(1-S)^{\delta}\, N_{\text{tot}}^{\gamma}}}_{\text{interaction}}
\;+\;
e
\label{eq:abnar-law}
\end{equation}
with sparsity $S$, which differs from how sparsity ($=1-\frac{K}{E}$) is defined in that study. The first
two terms recover the Chinchilla form~\citep{hoffmann2022training}
in the limit $S\!\to\!0$ (dense model). 

\begin{figure}[t!]
\centering
\includegraphics[width=\linewidth]{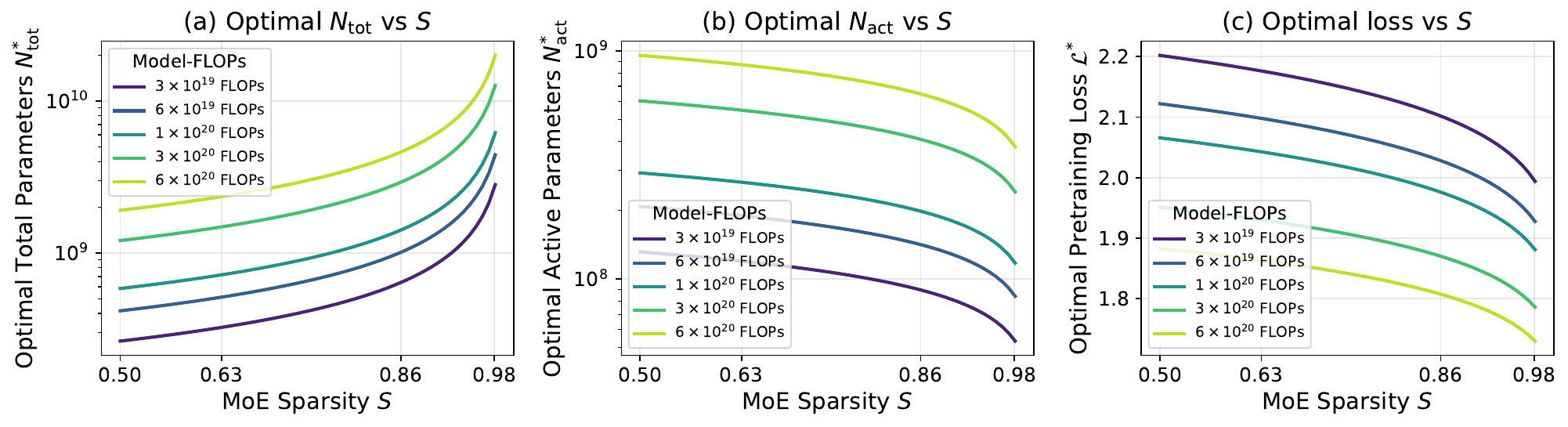}
\caption{Model FLOPs optimal $N_{\text{tot}}^{\star}$, $N_{\text{act}}^{\star}$,
and $\mathcal{L}^{\star}$ as a function of sparsity $S$, at five
budgets $C$. The curves are computed from our own $G\!=\!4$ fit of
Equation~\eqref{eq:abnar-law} on our runs. Larger MoE sparsity $S$ increases optimal total parameters, decreases optimal active
parameters, and lowers achievable loss at every model-FLOPs budget.}
\label{fig:figure3-abnar-repro}
\end{figure}

We fit the functional form in Equation~\eqref{eq:abnar-law} separately for each expert split factor $G$,
holding out the largest models by active parameters to test upward
extrapolation. The fits are statistically tight. The predicted and observed
loss collapse onto the identity line with held-out RMSE between
$0.0002$ and $0.0037$ in loss units across the four expert split factors,
under half a percent relative error. Figure~\ref{fig:per-G-parity} shows $G\!=\!4$ and
$G\!=\!8$, and the full $G\!\in\!\{1, 2, 4, 8\}$ panel is given in
Appendix~\ref{sec:sl_sparsity}. What the parity
plots also reveal is that the fitted coefficients change substantially
across $G$, so no single set of $(N_{\text{tot}}, S, D)$
coefficients describes all the expert split factors at once. This is what leads us to build $G$ into the law in
Section~\ref{sec:joint-NtSDG-law}. In Figure~\ref{fig:figure3-abnar-repro},  we find that three trends are monotone over $S \in [0.5, 0.981]$. These
three trends agree with those \citet{abnar2025parameters} report, recovered here
on our own runs and under our sparsity definition.
We return to this figure and its consequence
for the model-FLOPs optimum in Section~\ref{sec:boundary-optimum}.

\subsection{$G$-augmented Joint Scaling Law }
\label{sec:joint-NtSDG-law}

Before adding the expert split factor $G$ to the law, we note
that $G$ and the sparsity $S$ are distinct knobs that happen to share
the parameter counts. The two move different geometry variables and appear
coupled only because both these dimensions modulate $N_{\text{tot}}$ and $N_{\text{act}}$.
Empirically they are close to orthogonal on our grid, where the Pearson
correlation between $\log G$ and each of $\log N_{\text{tot}}$, $S$, and
$\log N_{\text{act}}$ is at most $0.02$ in magnitude, because the sweep varied
$d_{\text{expert}}$ independently of $(E, K)$ and every
$G\!\in\!\{1, 2, 4, 8\}$ is observed across nearly the full
$N_{\text{tot}}$ and $S$ ranges.

\begin{figure}[!t]
\centering
\includegraphics[width=0.85\linewidth]{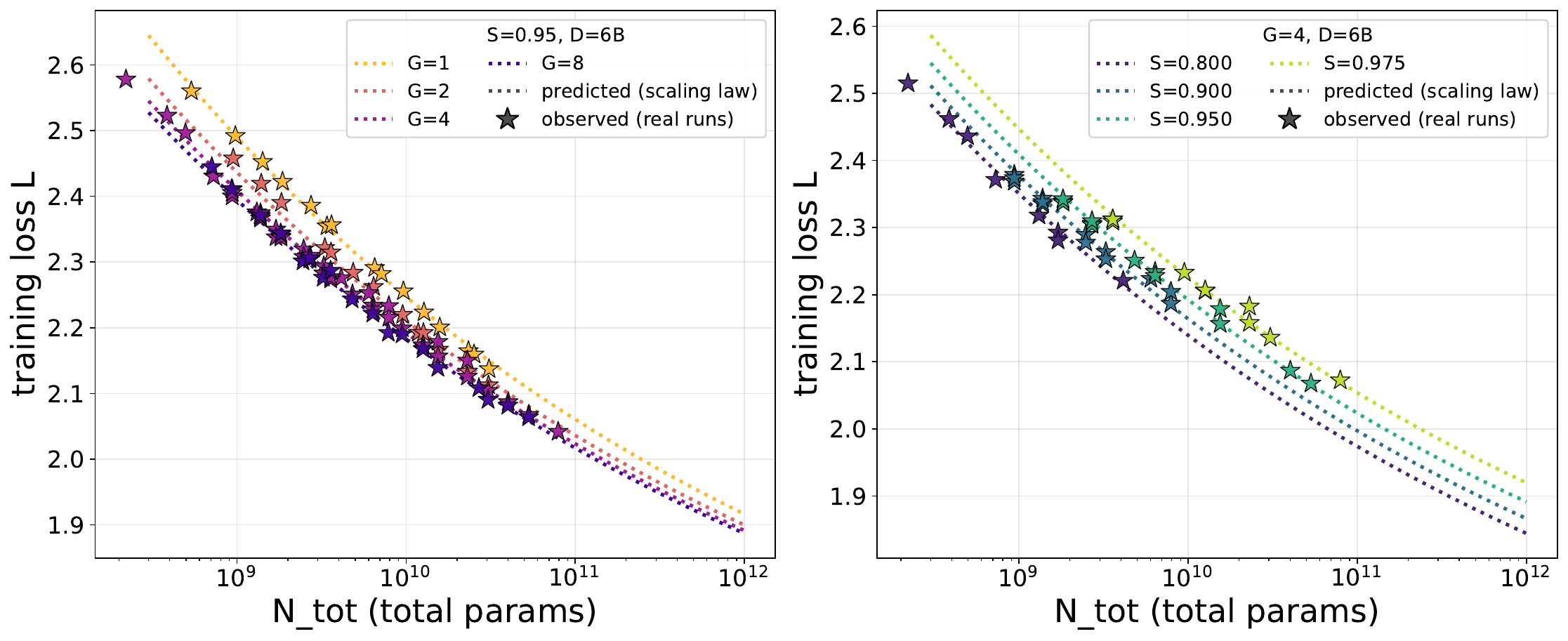}
\caption{Loss against total MoE parameters $N_{\text{tot}}$. In the left, one curve per expert split factor $G$ at fixed $S\!=\!0.95$,
$D\!=\!6$\,B. In the right, one curve per sparsity band at fixed $G\!=\!4$,
$D\!=\!6$\,B. The curves fan out rather than staying parallel (left), which is why $G$
enters the law as its own axis instead of being absorbed into an effective
parameter count, and Table~\ref{tab:joint-law-coeffs} bears this out, where the
$G$-exponent $\eta$ is identified rather than sitting on the coefficient ridge.}
\label{fig:motivation-G-sweep}
\end{figure}

A starting point to understand how to augment the scaling functional form
with the expert split factor $G$ is the conventional model FLOPs optimal
recipe. Figure~\ref{fig:motivation-G-sweep} shows that the loss curves in
$N_{\text{tot}}$ fan out rather than stay parallel as $G$ or $S$ varies, so
the effect of the expert split factor and of sparsity is not a rigid shift but depends
on $N_{\text{tot}}$, which is a coupling a separable law in
$(N_{\text{tot}}, S, D)$ cannot express. Following this observation, we tie $G$, $S$, and
$N_{\text{tot}}$ together in a single interaction term. The joint scaling law
we use, on the scaling law dimensions $(N_{\text{tot}}, S, G)$ induced by $Z$, is
\begin{align}
\mathcal{L}(N_{\text{tot}}, S, D, G)
&\;=\;
\frac{a}{N_{\text{tot}}^{\alpha}}
\;+\;
\frac{b}{D^{\beta}}
\;+\;
\frac{c}{(1-S)^{\lambda}}
\notag\\[2pt]
&\quad+\;
\frac{j}{(1-S)^{\delta}\,N_{\text{tot}}^{\gamma}\,G^{\eta}}
\;+\; e.
\label{eq:joint-law}
\end{align}
Here $G$ appears only in the denominator of the interaction term, so raising $G$
shrinks the sparsity penalty and lowers the predicted loss at fixed
$(N_{\text{tot}}, S, D)$. Placing $G$ in the
joint coupling rather than folding it into a single effective parameter
count is what keeps the fit balanced across $G$ values.

\begin{table}[h]
\centering
\small
\setlength{\tabcolsep}{6pt}
\renewcommand{\arraystretch}{1.1}
\begin{tabular}{@{}lrll@{}}
\toprule
\textbf{Coeff.} & \textbf{Value} & \textbf{$95\%$ BCa CI} & \textbf{Identifiability} \\
\midrule
$a$        & $8.462$  & $[8.24,\, 8.85]$    & ridge \\
$b$        & $88.32$  & $[87.8,\, 89.9]$    & ridge \\
$c$        & $0.1828$ & $[0.181,\, 0.184]$  & ridge \\
$j$        & $0.7123$ & $[0.710,\, 0.713]$  & ridge \\
$e$        & $0.3129$ & $[0.307,\, 0.316]$  & ridge \\
$\alpha$   & $0.1048$ & $[0.104,\, 0.106]$  & ridge \\
$\beta$    & $0.2070$ & $[0.206,\, 0.208]$  & identified \\
$\lambda$  & $0.1249$ & $[0.119,\, 0.138]$  & ridge \\
$\delta$   & $0.5567$ & $[0.550,\, 0.577]$  & ridge \\
$\gamma$   & $0.1702$ & $[0.167,\, 0.174]$  & ridge \\
$\eta$     & $\mathbf{+0.9513}$ & $\mathbf{[+0.911,\, +0.954]}$ & identified \\
\bottomrule
\end{tabular}
\caption{Fitted coefficients of the joint scaling
law in Equation~\eqref{eq:joint-law}. $95\%$ BCa CIs come
from a warm-started wild bootstrap in the deployed basin.
We work through which coefficients the data
identify and which sit on the flat ridge in Appendix~\ref{app:scaling-law-coeffs}.}
\label{tab:joint-law-coeffs}
\end{table}

\paragraph{Fitting Results.} We fit the form in Equation~\eqref{eq:joint-law} on the scaling grid spanning $G\!\in\!\{1, 2, 4, 6, 8\}$. We hold out the
top-$10\%$ of runs by $N_{\text{act}}$ within each $G$-stratum to validate the observations
we do not use in training. We minimize the Huber objective on $\log\mathcal{L}$
($\delta_{\text{Huber}}=0.1$) by L-BFGS-B in log-coefficient space. Table~\ref{tab:joint-law-coeffs} reports the point estimates together
with $95\%$ confidence intervals, and marks each coefficient as identified or ridge.
\begin{figure}[t!]
\centering
\includegraphics[width=0.55\linewidth]{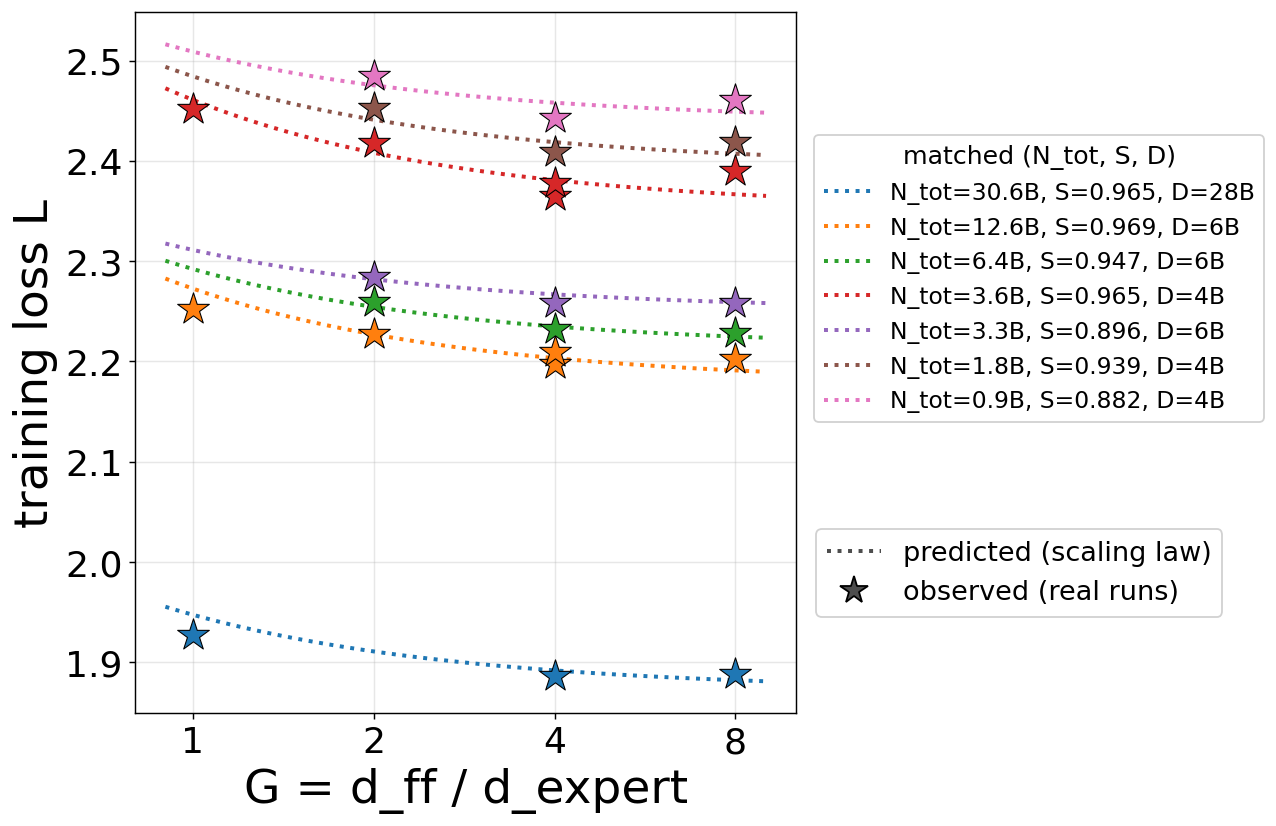}
\caption{Training loss against expert split factor
$G\!=\!d_{\text{ff}}/d_{\text{expert}}$ at matched
$(N_{\text{tot}}, S, D)$ operating points. The predicted loss decreases
monotonically with $G$. The observed runs decrease up to $G\!=\!4$ and flatten
or slightly reverse at $G\!=\!8$ within run noise. The gap between the coarsest
and finest experts widens along the plotted series, tracking their rising
sparsity.}
\label{fig:granularity-not-sparsity}
\end{figure}
A coefficient is identified when the data pin its value uniquely, and it
sits on a ridge when only a combination of several coefficients is fixed. Table~\ref{tab:joint-law-coeffs} shows that
$\eta$ and $\beta$ are identified here, because $G$ and $D$ each enter the
law in a single place that no other coefficient can absorb. The
$G$ exponent $\eta\!\approx\!0.95$ is therefore our robust observation, and
since it multiplies the joint sparsity and capacity term, its reading is that
larger $G$ discounts the loss penalty that sparsity imposes at fixed capacity. We do not claim a universal expert split factor law with our functional form fits, since our scaling data traverses the discrete space
$G\!\in\!\{1, 2, 4, 6, 8\}$.  We check this against the runs in Figure~\ref{fig:granularity-not-sparsity}, where the
predictions of Equation~\eqref{eq:joint-law} decrease monotonically with $G$ at
matched $(N_{\text{tot}}, S, D)$ while the observed losses decrease up to $G\!=\!4$
and flatten or slightly reverse at $G\!=\!8$ within run-to-run noise.
The law attributes the widening gap between the coarsest and finest experts
to the rising sparsity of the larger operating points through the $(1-S)^{-\delta}$ factor rather than to
$N_{\text{tot}}$ itself, since at matched $S$ the
$N_{\text{tot}}^{-\gamma}$ decay of the interaction term shrinks the gap as
models grow.
From the
systems viewpoint of this paper, $\eta$ quantifies the rate at
which finer-grained experts soften the sparsity penalty at a fixed
memory footprint, a benefit that the performance model in
Section~\ref{sec:performance-model} will price against sparse MoE layer communications overhead.

\begin{figure}[!t]
\centering
\includegraphics[width=0.42\linewidth]{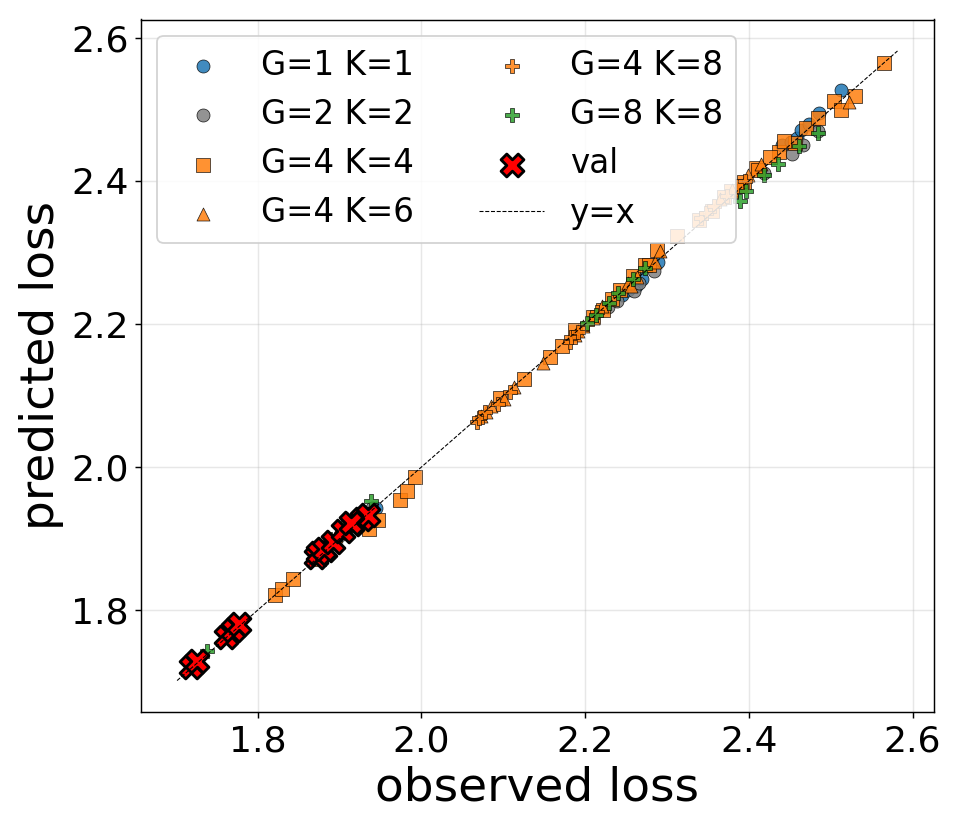}
\caption{Joint-law parity plot over all runs, $R^2=0.99877$. The crosses are
the held-out validation runs, the largest $10\%$ by $N_{\text{act}}$ within
each $G$ stratum, with a held-out RMSE of $0.00779$ in loss units. Marker
shape and color pick out a sample of the $(G, K)$ pairs the sweep covers.}
\label{fig:joint-law-fits}
\end{figure}

We further validate the fit quality in
Figure~\ref{fig:joint-law-fits}, which collapses a subset of all
runs across $G\!\in\!\{1, 2, 4, 6, 8\}$ onto the identity line, and the joint law
attains a held-out RMSE of $0.00779$ in loss units on the largest $10\%$ of
runs by $N_{\text{act}}$ within each $G$ stratum. The fitted form also
behaves as we expect
of a well-formed scaling law. It is monotone-decreasing in $N_{\text{tot}}$,
$D$, and $G$, and convex in $S$ with a sharp penalty as $S\!\to\!1$. As
$G\!\to\!\infty$ the joint term drops out and the loss settles onto
$a/N_{\text{tot}}^{\alpha} + b/D^{\beta} + c\,(1-S)^{-\lambda} + e$. As
$N_{\text{tot}}, D\!\to\!\infty$ at fixed MoE sparsity $S$, the loss approaches
$e + c\,(1-S)^{-\lambda}$, so the sparsity floor persists at infinite scale and
the irreducible term is $e_{\infty}\!=\!e+c\!\approx\!0.50$ in the dense limit
$S\!\to\!0$ rather than $e$ alone. 


\section{Estimating $MFU(.)$ with a Performance Model}
\label{sec:performance-model}
This section specifies the design of a performance estimator that predicts
$\mathrm{MFU}(Z, P_{\ell}, B)$ and $\mathrm{Mem}(Z, P_{\ell}, B)$ for MOSAIC in
Section~\ref{sec:mosaic}. This estimator takes only the model specification and the
physical device layout as input, and produces a parallelism layout, single-iteration
running time, MFU, and memory footprint as output. A closed-form scaling rule cannot
stand in for MFU predictions, because systems scaling is nonlinear and MFU is therefore nonlinear
in the device count. So, doubling the GPUs neither guarantees doubling the throughput nor preserves
the layout that was optimal before, so we cost the forward and backward passes of a
single iteration analytically against measured collective timings instead.

\subsection{Design Principle}
The core principle behind our performance model is to minimize MFU prediction error
while keeping the model simple and physically-grounded: every term
must trace to a hardware specification, a measured kernel or collective timing, or a
scheduling equation, never a hand-tuned constant. We pair a lightweight analytical
skeleton (FLOPs, collective-algorithm equations, and the 1F1B pipeline schedule)
with a set of carefully designed microbenchmarks run on the target hardware, so that the
model consumes \emph{measured} kernel efficiencies and collective bandwidths rather than
attempting to reproduce the intrinsic complexity of runtime kernel behavior from first
principles. Two effects cannot be captured by an uncontended microbenchmark,
namely CPU-dispatch-bound execution on small models and cross-node straggler and
overlap in large collectives. We isolate each into a small, explicitly labeled
correction rather than spreading fudge factors across every component.

In the large, this design yields a practical calculator that guides performance
engineering toward the optimal MFU through an efficient parallelism-layout search. In the
small, it also offers a fine-grained per-component time breakdown of a single MoE
training or inference iteration, which supports bottleneck diagnosis.
\subsection{Functionality}
The performance model is driven by a model-architecture specification that fully
describes a training run, as shown in Appendix~\ref{app:performance-model-input}. It
groups settings into seven blocks. The \textit{model} block defines the dense transformer geometry
(depth, widths, attention heads, vocabulary, and architectural switches such as SwiGLU,
rotary embeddings, and weight tying), along with the compute precision, which selects
among floating-point formats (\texttt{bf16}, \texttt{fp16}, \texttt{fp8}, \texttt{fp4},
\texttt{tf32}, \texttt{fp32}, and others) so that GEMMs and attention are costed against
the matching device peak-FLOPs lane. The \textit{model.moe} block adds the mixture-of-experts layer
(expert count, top-$k$ routing, expert width, shared-expert width, and the capacity factor
that sizes the routed token budget and the all-to-all traffic). The \textit{search} block declares
the device budget. The \textit{performance} block sets the activation-checkpointing strategy
(\texttt{none}, \texttt{selective}, \texttt{super-selective}, or \texttt{full}). The \textit{optimizer} block selects the optimizer family
(\texttt{adam}, \texttt{adamw}, \texttt{muon}, \texttt{scion}, and more), which sets the
optimizer-state footprint and update FLOPs. The \textit{data} block gives the batch and sequence
shape, and the \textit{hardware} block names the target node type.

From this specification, the tool builds an internal model and hardware representation,
then enumerates the feasible execution plans $P_{\ell} =$ ($P_{\text{TP}}$, $P_{\text{EP}}$,
$P_{\text{CP}}$, $P_{\text{PP}}$, $P_{\text{DP}}$, $B_{\text{micro}}$)
that tile the device budget and
estimates the compute, memory, and communication cost of each. The output is a per-layout
prediction of MFU and memory usage, used to rank parallelism configurations against
realistic training jobs. We give a worked
example of both the specification and the returned output in Appendix~\ref{app:performance-model-input}, where we also describe the
layout search that produces the ranking.

\subsection{Microbenchmarks}
 \label{sec:microbenchmarks}
Closed-form FLOPs divided by device peak FLOPs over-predict throughput: real
kernels reach only a fraction of peak that depends on operand shape, and memory-bound
operations do not scale with FLOPs at all. We therefore benchmark the dominant
computation and communication components on the target hardware and feed the measured
timings into the model as interpolated lookup tables. We interpolate within the measured
range and refuse to extrapolate silently outside it. The benchmarked components are:
\begin{itemize}
  \item \textbf{Dense GEMMs}: achieved utilization as a function of $(M, N, K)$ and
        precision, for the attention, MLP, shared-expert, and LM-head projections.
  \item \textbf{Grouped GEMMs} (the MoE expert MLP): the fused
        \texttt{GroupedLinear} operator, keyed by the number of local experts
        (\texttt{num\_gemms}), the per-expert token load (the grouped tile $M$), and the
        expert shape. This is essential because the expert MLP runs many small
        per-expert tiles rather than one large GEMM, where its efficiency is set by the
        \emph{per-group} tile
        $M_{\text{expert}} = \lceil T_{\text{seq}}\,B_{\text{micro}}\,K / E \rceil$
        (sequence length $T_{\text{seq}}$, micro-batch $B_{\text{micro}}$, top-$K$,
        $E$ experts), which is invariant to expert and data
        parallelism, not by the aggregate token count. Costing it as a single dense GEMM
        over-credits utilization by up to an order of magnitude in the token-dropless regime.
  \item \textbf{Scaled dot-product attention} (SDPA) with the Flash-Attention
        backend~\citep{zadouri2026flashattention}, forward and backward, indexed by the
        TP-sharded head counts that actually execute on one GPU.
  \item \textbf{MoE top-$k$ routing} and \textbf{element-wise} operations (\texttt{silu},
        \texttt{residual\_add}), forward and backward.
  \item \textbf{Cross-entropy over the vocabulary}: the fused vocab-parallel loss on
        \texttt{fp32}-upcast logits. This is far more expensive than a naive count of
        logit passes suggests, and it lands entirely on the last pipeline stage.
  \item \textbf{NCCL collectives}: \texttt{all\_reduce}, \texttt{reduce\_scatter},
        \texttt{all\_gather}, \texttt{all\_to\_all}, and \texttt{send\_recv}, as effective
        bus-bandwidth curves swept over message size \emph{and} group size, covering both
        the intra-node (NVSwitch) and inter-node (EFA) regimes and both contiguous
        (dense-DP) and strided (expert-DP) rank layouts. These curves differ from the
        analytic link-peak-times-protocol-efficiency estimate by up to $8\times$ in both
        directions and exhibit a strong size ramp that specifications alone do not
        predict.
  \item \textbf{Per-kernel CPU dispatch gap}: the steady-state CPU enqueue stall between
        kernels in an eager execution stream, which bounds throughput when kernels are
        small (see below).
\end{itemize}\subsection{Iteration-Time Model}
\label{sec:iteration-time-model}
The predicted per-iteration time sums a transformer-block compute term, a pipeline-bubble
term, a vocabulary-stage (embedding and LM-head/cross-entropy) term, the exposed
data-parallel and pipeline-parallel collective times, and the optimizer step. Three parts
of this decomposition depart from a naive roofline and warrant explanation.

\paragraph{Compute versus dispatch (per-microbatch wall).}
For small models with a micro-batch size of one, the transformer-block kernels are tiny
(e.g.\ $<0.1$~ms per GEMM) and a single microbatch launches tens of thousands of kernels. 
The CPU cannot enqueue them fast enough and the GPU starves. We therefore model the
per-microbatch wall as
\begin{equation}
  t_{\text{mb}} = \max\!\big(t_{\text{cmp}},\ t_{\text{disp}}\big),
  \qquad
  t_{\text{block}} = n_{\text{mb}}\, t_{\text{mb}},
  \qquad
  n_{\text{mb}} = \frac{B_{\text{global}}}{P_{\text{DP}}\, B_{\text{micro}}}
\end{equation}
where $n_{\text{mb}}$ is the number of microbatches per optimizer step, $t_{\text{cmp}}$ is
the summed measured GPU-compute time of the stage's layers, and
$t_{\text{disp}} = L_{\text{stage}}\,\kappa(E_{\text{loc}})\,t_{\text{gap}} +
t_{\text{host}}$ is the CPU-dispatch floor over $L_{\text{stage}}$ layers per stage, with a
measured per-kernel gap $t_{\text{gap}}$, a fixed per-microbatch host term
$t_{\text{host}}$, and a kernel-count law
$\kappa(E_{\text{loc}}) = \kappa_0 + \kappa_1\,E_{\text{loc}}$ that grows with the number of
local experts $E_{\text{loc}}$ (the surrounding framework kernels, namely permutation,
copies, and bias-adds, dominate the launch count, not the fused expert GEMM). The $\max$
makes this term \emph{self-limiting}, where large-model kernels are large enough to hide
dispatch, so the term vanishes and those configurations remain compute-bound.

\paragraph{Pipeline bubble.}
With 1F1B scheduling, the pipeline fill and drain phases leave each stage idle for an
effective fraction of the iteration. For $n_{\text{mb}}$ microbatches per pipeline,
$P_{\text{PP}}$ pipeline stages, and $P_{\text{VPP}}$-way interleaving \footnote{https://docs.nvidia.com/nemo/megatron-bridge/0.2.0/parallelisms.html\#interleaved-pipeline-parallel-schedule}, the bubble fraction
is
\begin{equation}
  f_{\text{bubble}} = \frac{P_{\text{PP}} - 1}{P_{\text{VPP}} \, n_{\text{mb}}}
  \label{eq:bubble}
\end{equation}
so the pipeline bubble time is $t_{\text{bubble}} = f_{\text{bubble}}\, t_{\text{block}}$.
Each rank computes $n_{\text{mb}}$ microbatches of work and idles
$(P_{\text{PP}}-1)/P_{\text{VPP}}$ microbatch-slots during warmup and cooldown. That ramp is
the entire bubble. This form replaces an earlier weighted-average expression that added a
separate fill/drain term, charging each of the $2(P_{\text{PP}}-1)$ warmup/cooldown
microbatches a $(P_{\text{PP}}-1)/P_{\text{PP}}$ stage bubble on top of the
$(P_{\text{PP}}-1)/n_{\text{mb}}$ steady-state term, and clamped the result at
$(P_{\text{PP}}-1)/P_{\text{PP}}$. That expression double-counted the fill/drain phase, and
its clamp under-counted the small-batch regime it was meant to bound.

\paragraph{Vocabulary-stage imbalance.}
The token embedding is placed on the first pipeline stage and the LM head plus
cross-entropy is placed on the last, in addition to those stages' transformer layers. Because
1F1B is paced by the slowest stage, the surplus $t_{\text{last}}$ on the last stage (the
measured LM-head GEMM forward and backward plus the measured cross-entropy) is exposed
once per microbatch, and it further lengthens the fill/drain ramp. The stage-imbalance
contribution is
\begin{equation}
  t_{\text{vocab}}
  = (n_{\text{mb}}-1)\,\max(t_{\text{first}}, t_{\text{last}})
    + t_{\text{first}} + t_{\text{last}}
\end{equation}
with an added $t_{\text{last}}\,n_{\text{mb}}\,f_{\text{bubble}}$ term for the ramp. Here $t_{\text{first}}$
is the (bandwidth-bound) embedding gather/scatter. This term is essential for
deep-pipeline MoE configurations, where the vocabulary-parallel LM head is a large
fraction of the critical path and was previously unpriced.

The exposed data-parallel and pipeline-parallel collective times are taken directly from
the measured bandwidth curves of Section~\ref{sec:microbenchmarks}, with the DP grad
reduce-scatter and param all-gather overlapped against compute where windows allow.

\subsection{Calibration}
Because the compute and wire-time terms above are read from measured lookup tables, most
of the model requires no correction factor at all. What remains, and cannot be recovered
from an uncontended microbenchmark, is a single \emph{runtime} effect, namely cross-node
straggler and overlap. When an expert-parallel or pipeline-parallel group spans multiple
nodes, many collective rings run concurrently and are paced by the slowest rank (traces
show per-call inter-node latencies with a tight median but tails two to three orders of
magnitude larger), so the realized cost exceeds the uncontended microbenchmark. When a
deep pipeline fits within a single node, conversely, the heavy last stage overlaps other
stages over NVLink better than a straggler-free 1F1B accounting assumes, so the realized
step is \emph{faster} than predicted.

We isolate this into one explicitly labeled correction, the model's only fitted term, which
multiplies the analytical iteration time,
\begin{equation}
  t^{\text{cal}}_{\text{iter}} = \chi_{\text{sys}}\, t^{\text{analytic}}_{\text{iter}}
  \label{eq:cal-iter}
\end{equation}
where the systems calibration multiplier is
\begin{equation}
\begin{split}
  \chi_{\text{sys}}
  &= 1
  + c_{\text{A2A}}\,\frac{t_{\text{A2A}}}{t_{\text{iter}}}\,\mathbb{1}\!\left[N^{\text{EP}}_{\text{node}} > 1\right]
  + c_{\text{PP}}\,\big(N^{\text{PP}}_{\text{node}} - 1\big) \\
  &\quad + c_{\text{ovl}}\,\mathbb{1}\!\left[P_{\text{PP}} \ge N_{\text{gpu/node}},\ N^{\text{PP}}_{\text{node}} = 1\right].
\end{split}
\label{eq:desync}
\end{equation}
Here $N^{\text{EP}}_{\text{node}} = \lceil P_{\text{EP}}\,P_{\text{TP}} /
N_{\text{gpu/node}}\rceil$ and
$N^{\text{PP}}_{\text{node}} = \lceil P_{\text{PP}} / N_{\text{gpu/node}}\rceil$ count how
many nodes of $N_{\text{gpu/node}}$ GPUs each the expert-parallel and pipeline-parallel
groups span, and $t_{\text{A2A}}$ is the modeled per-step expert all-to-all wire time. Each
driver is tied to a physical quantity, where (i) the inter-node A2A straggler scales the
penalty by how large a fraction of the step the expert A2A already is, and applies only when
the EP group crosses a node boundary. (ii) The deep-pipeline term scales with the number of
inter-node hops the pipeline crosses. (iii) The intra-node overlap \emph{credit}
($c_{\text{ovl}} < 0$) applies only to a deep pipeline confined to one node. Single-node,
shallow-pipeline configurations incur no correction ($\chi_{\text{sys}} = 1$).

Together, the microbenchmarks, the iteration-time model, and this single
calibration term constitute the estimator we release as ScalePlan, whose layout
search supplies the $\mathrm{MFU}$ and $\mathrm{Mem}$ predictors that MOSAIC
consumes. We validate these predictions against
measured MFU on real training sweeps in Section~\ref{sec:perf-validation}.
\section{Optimal MoE Sparsity Scaling}
\label{sec:hardware-codesign}

In Section~\ref{sec:mosaic}, we defined deliverable model FLOPs $C_\text{deliverable}$ as the systems-aware constraint used by MOSAIC. We now develop the motivation sketched in Figure~\ref{fig:mosaic} in more detail for sparse MoEs. We bring together the two pieces, the sparse MoE scaling laws and the hardware-deliverable model FLOPs, to explain why the deliverable model FLOPs $C_{\text{deliverable}}$ of Equation~\eqref{eq:deliverable-flops} is the budget that makes sense for sparse MoEs before we turn to results. A practical use of scaling laws such as ours is to
pick the scaling dimensions that minimize predicted loss under a model FLOPs budget, $C_{\text{model}}\!=\!6\,N_{\text{act}}D$. Simply said, we want the
$(N_{\text{tot}}, S, D, G)$ that minimizes our law under the budget.
Conventionally, solutions of this kind proceed in two stages. In the
first stage, we solve an optimization for the model FLOPs optimal
$(N_{\text{tot}}^{\star}, S^{\star}, D^{\star}, G^{\star})$, treating
these as continuous variables. In the second stage, we map that optimum
to a discrete, deployable model configuration, since $S$ and $G$ are not free
knobs one can set directly. $S$ is a derived ratio, and $G$ moves only by changing the
discrete geometry $(E, K, L_{\text{layers}}, d)$ of Section~\ref{sec:preliminaries}. To that end, Sections~\ref{sec:boundary-optimum} and~\ref{sec:law-to-allocation} discuss two limitations of model-FLOPs based optimality for sparse MoE scaling that make such a two-stage optimization the wrong choice. The next section shows that the sparse MoE dimensions in our scaling law admit no interior optimum under a model-FLOPs budget, which forces the budget to be recast as an architecture-dependent deliverable model-FLOPs budget.

\subsection{Model-FLOPs Optimal Sparsity Is Boundary-Seeking}
\label{sec:boundary-optimum}

\begin{figure}[!t]
\centering
\includegraphics[width=0.8\linewidth]{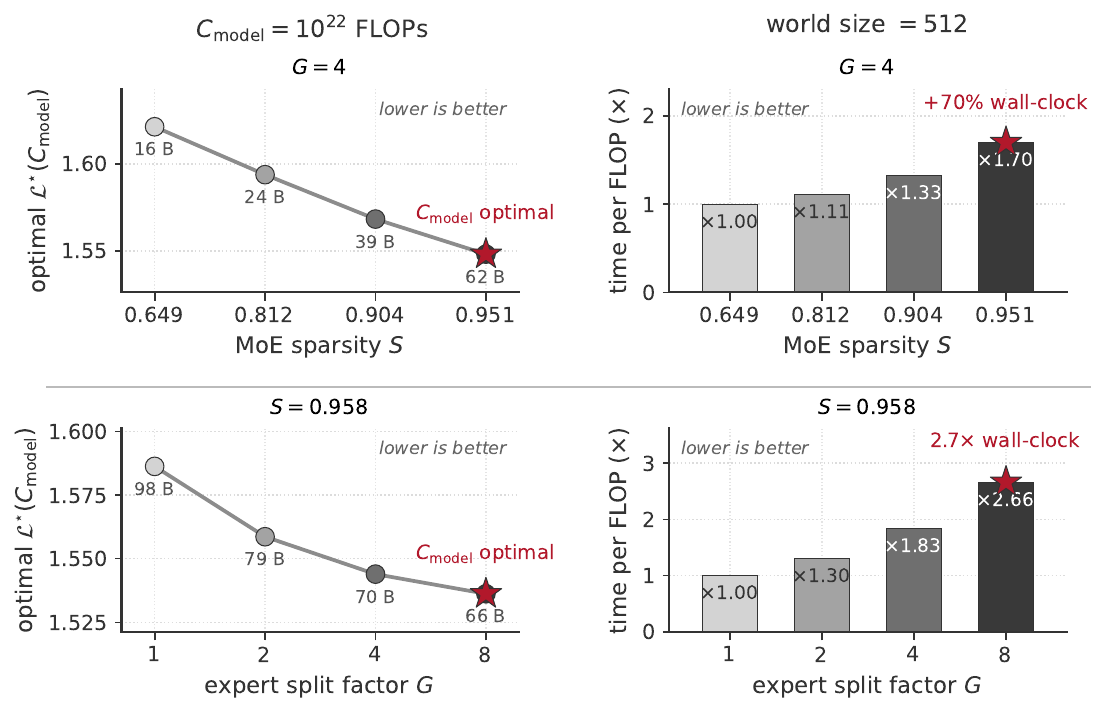}
\caption{\textbf{Opposing views of model scaling and systems cost.}
The top row varies MoE sparsity $S$ holding $G\!=\!4$. The bottom row varies the expert split factor $G$ holding $S\!=\!0.958$. The left column gives the optimal loss
under a fixed model-FLOPs budget. Each marker there is labeled with the optimal
total parameters $N_{\text{tot}}$ under that budget. The right column shows the wall-clock time each configuration needs per delivered model FLOP on $512$ devices (64 p6 nodes), relative to the training time of the lowest sparsity
configuration. The $C_\text{model}$ optimal loss configuration pays $1.70\times$ per FLOP in the top
row and $2.66\times$ in the bottom. The starred winners sit at opposite ends, so
the $C_\text{model}$ optimal loss configuration is the systems-slowest.}
\label{fig:motivation-S}
\end{figure}

Conditioning on the scaling law, the conventional scaling studies seek the \textit{optimal sparsity $S$}
that minimizes loss at a fixed budget $C_{\text{model}}$. Concretely, with
$N_{\text{act}} = (1-S)\, N_{\text{tot}}$ and the budget
$C_{\text{model}} = 6\, N_{\text{act}}\, D$ (at fixed $G$), the model FLOPs only
optimization problem reads
\begin{equation}
\min_{N_{\text{tot}}, S, D}\;
\mathcal{L}(N_{\text{tot}}, S, D, G)
\quad
\text{s.t.}
\quad
6\,(1-S)\, N_{\text{tot}}\, D \;=\; C_{\text{model}},
\quad
N_{\text{tot}}, D > 0,
\quad
S \in [0, 1).
\label{eq:compute-only-program}
\end{equation}
Solving Equation~\eqref{eq:compute-only-program} with the joint law of
Equation~\eqref{eq:joint-law} at
$C_{\text{model}} \in \{3\times10^{19},\, 6\times10^{19},\, 10^{20},\,
3\times10^{20},\, 6\times10^{20}\}$ FLOPs recovers the three trends that the
$G\!=\!4$ fit of Equation~\eqref{eq:abnar-law} produces in
Figure~\ref{fig:figure3-abnar-repro}. The important observation in the plots is that $N_{\text{tot}}^{\star}(S)$
increases monotonically with $S$, $N_{\text{act}}^{\star}(S)$
decreases, and $\mathcal{L}^{\star}(S)$ decreases monotonically at
every $C_{\text{model}}$. Because $\mathcal{L}^{\star}(S)$ decreases monotonically over the calibrated
sparsity range, the interior condition
$\partial \mathcal{L}^{\star}/\partial S = 0$ is not satisfied within this
region. The optimal sparsity under the $C_{\mathrm{model}}$ budget, therefore lies at
$S=S_{\max}$, the upper boundary of our data support. Thus, within this
range where the scaling law is calibrated, model-FLOPs only optimization
reduces to ``make $S$ as large as the supported architecture space allows.''
The left column of Figure~\ref{fig:motivation-S} shows the same monotone descent
on both knobs, in $S$ at fixed $G$ and in $G$ at fixed $S$.
We do not claim that this monotonic trend continues arbitrarily close to
$S=1$, where the asymptotic sparsity-penalty terms in the fitted law
eventually dominate.

\subsection{The Boundary Optimum Is Not Necessarily the Cluster Optimum}
\label{sec:law-to-allocation}

In the preceding section, the model-FLOPs optimum proved boundary-seeking in
sparsity, returning the largest $S$ the architecture space supports rather than
an interior choice. Additionally and more importantly, practitioners do not buy
model FLOPs, they buy GPU-hours on a fixed cluster over a fixed
wall-clock window, and what they care about at the end of training is
the training loss reached. The $C_\text{deliverable}$
the cluster can execute thus becomes the metric that matters once we
account for the systems cost of hosting a model across a fleet of GPUs
\citep{shoeybi2019megatron}, which is what we show in Figure~\ref{fig:mosaic}. In particular, sharding a model through
various parallel layouts incurs costs from expert-parallel all-to-all,
pipeline bubbles, optimizer sharding, and host-side stalls
\citep{singh2023hybrid, fernandez2024hardware}. What follows is that a model FLOPs budget leaves out the systems-side constraints that together with device peak FLOPs provide $C_\text{deliverable}$.

To make the motivation concrete, we take as the systems metric the wall-clock
time a configuration needs per delivered model FLOP, where a lower value means
more of the purchased hardware is doing useful work and the same loss is
reached in fewer GPU-hours. The top row of Figure~\ref{fig:motivation-S} shows how loss, sparsity,
and execution cost trade off on the way to the deployable optimum. Its left
panel plots the optimal loss $\mathcal{L}^{\star}$ reached at each sparsity level at
fixed $C_{\text{model}}\!=\!10^{22}$ model FLOPs, after minimizing over active
parameters. It drops monotonically as sparsity grows, so under a model FLOPs
budget the highest-sparsity configuration ranks first and the lowest-sparsity
one ranks last. The right panel prices the same grid at world-size $512$
(number of GPU devices) with the performance model of
Section~\ref{sec:performance-model}, optimized over the
parallelism layout, and the wall-clock per model FLOP rises monotonically with
sparsity to $1.70\times$ the densest configuration. This opposing ordering is precisely why the architecture-dependent deliverable model-FLOPs constraint introduced in Section~\ref{sec:mosaic} is required for sparse MoEs.

An important point to note here is that, unlike the sparsity scaling laws behind the model design report of Kimi K2 \citep{team2025kimi}, which fix the active parameters, or the recommendation of
\citet{abnar2025parameters} that at a fixed total parameter count a larger
training budget lowers the optimal sparsity, we hold neither
$N_{\text{act}}$ nor $N_{\text{tot}}$ fixed in Figure~\ref{fig:motivation-S}. We instead propose a shift toward
hardware constraints of Section~\ref{sec:mosaic} to resolve these optimality questions.

\section{Experiments and Results}
\label{sec:results}

In this section, we demonstrate how we utilize MOSAIC towards the goal of searching for optimal model configuration and the parallel execution layout jointly under a hardware budget. We utilize the aforementioned components of the scaling laws and the performance model towards this goal. We detail how this search
solves the program of Section~\ref{sec:mosiac-formal} in Appendix~\ref{app:hw-optimization} exactly through structure-exploiting enumeration. We note that generic MINLP solvers are poorly
matched to it, since the MFU and memory constraints are measured lookup
surfaces rather than symbolic functions. To
surface how a hardware-aware optimum differs from a purely
model FLOPs optimal one, we first discuss the need for a geometry scaling ladder in
Section~\ref{sec:geometry-ladder}. Sparsity $S$ and the expert split factor $G$
are derived from the discrete geometry rather than set directly, so the search
needs a closed set of realizable architectures, which the ladder supplies and the
performance model then profiles for systems cost. Our scaling law sweep runs also share the same ladder as every experiment reported here, so one family of geometries both fits the law and is searched by the framework for validation. Every experiment
also uses the joint law $\mathcal{L}(N_{\text{tot}}, S, D, G)$ of
Equation~\eqref{eq:joint-law}, not the per-$G$ fits of
Equation~\eqref{eq:abnar-law}.

After presenting the ladder, we validate the performance
model against
measured MFU in Section~\ref{sec:perf-validation}, since the whole framework rests on its predictions. The
co-design results follow in three parts, where the first two search the
full feasible grid with the framework alone and the third launches its
ranked predictions as real training runs. These comprise (1) the loss-optimal configuration and the loss-versus-model-FLOPs
spread over feasible configurations under a fixed $C_{\rm peak}$ envelope
in Section~\ref{sec:minlp-implementation}, (2) the optimal MoE-layer geometry in sparsity and expert split factor $G$ as we vary the
cluster size (nodes) and training window (days) in Section~\ref{sec:results-gpuhours}, and (3) staged real training runs that test the predicted rank ordering against measured throughput in Section~\ref{sec:results-staged}.
A fitted loss surface can be validated against the values it predicts or against
the configuration ordering it induces. This ordering is what a practitioner acts on when choosing model
configurations, which is why we end the results section with real runs
selected under a common hardware envelope rather than with extrapolations
of the loss surface alone.

All our
validation and training runs use AWS p6-b200.48xlarge
nodes\footnote{https://aws.amazon.com/ec2/instance-types/p6/}, each with
$8$ NVIDIA~B200 GPUs. We provide the details of the Megatron and
Torch-based framework we stage our runs on in
Appendix~\ref{app:training-stack}, covering which Megatron-Core MoE
features we enable, the loop-level effects that make realized MFU trail
the analytical prediction, the host-side data-pipeline costs, and the
hardware and interconnect assumptions that tie the recovered optimum to a
specific platform.

\subsection{Geometry Scaling Ladder}
\label{sec:geometry-ladder}

Since we want to show how the hardware-aware
optimum shifts as the deliverable model FLOPs change with scale, we first need a way to climb the model-scale axis. Growing a model means growing depth, width, and FFN
width together, so that axis is well defined only once we fix how those
dimensions co-scale. A geometry scaling ladder is that choice, and it reduces the
backbone to a single seed. This matters for MOSAIC, which selects a discrete geometry in a single step rather than through the traditional two-stage model-FLOPs allocation, so the geometry must be pinned down directly.
We parametrize the ladder in the following way: the layer count and head count both follow $L_{\text{layers}} = m\, q$ and
$n_{\text{head}} = m\, q$ with the multiplier $m$ fixed to 4, so the residual
width is $d = n_{\text{head}}\, d_{\text{head}}$ at $d_{\text{head}} =
128$, and the reference dense FFN width is $d_{\text{ff}} = 2.5\, d$. A
single seed $q$ thus fixes depth ($L_{\text{layers}}$), width ($d$), and FFN width
together at a constant depth-to-width ratio, so the ladder grows the
backbone along one coherent axis while the MoE axes $(E, K, G)$, and through
them the sparsity $S$, are searched on top of it. We fix the co-scaling functions of our ladder this way and
vary the geometry seed $q$, and we leave other variants of co-scaling geometry as future work.

With the ladder in place, the
framework searches over the seed $q$, the expert count $E$, the top-$K$, the
split factor $G$, and the data budget $D$, returning the optimum, and the
sparsity $S$ and the parameter counts $N_{\text{tot}}$ and $N_{\text{act}}$ are
derived from the selected geometry rather than searched directly. We show how this is realized in the following sections. We list the model scales we
sweep in Table~\ref{tab:scale-ladder} of Appendix~\ref{app:results}.

\subsection{Performance Model Validation}
\label{sec:perf-validation}
\begin{figure}[t]
  \centering
  \begin{subfigure}{0.45\textwidth}
    \includegraphics[width=\linewidth]{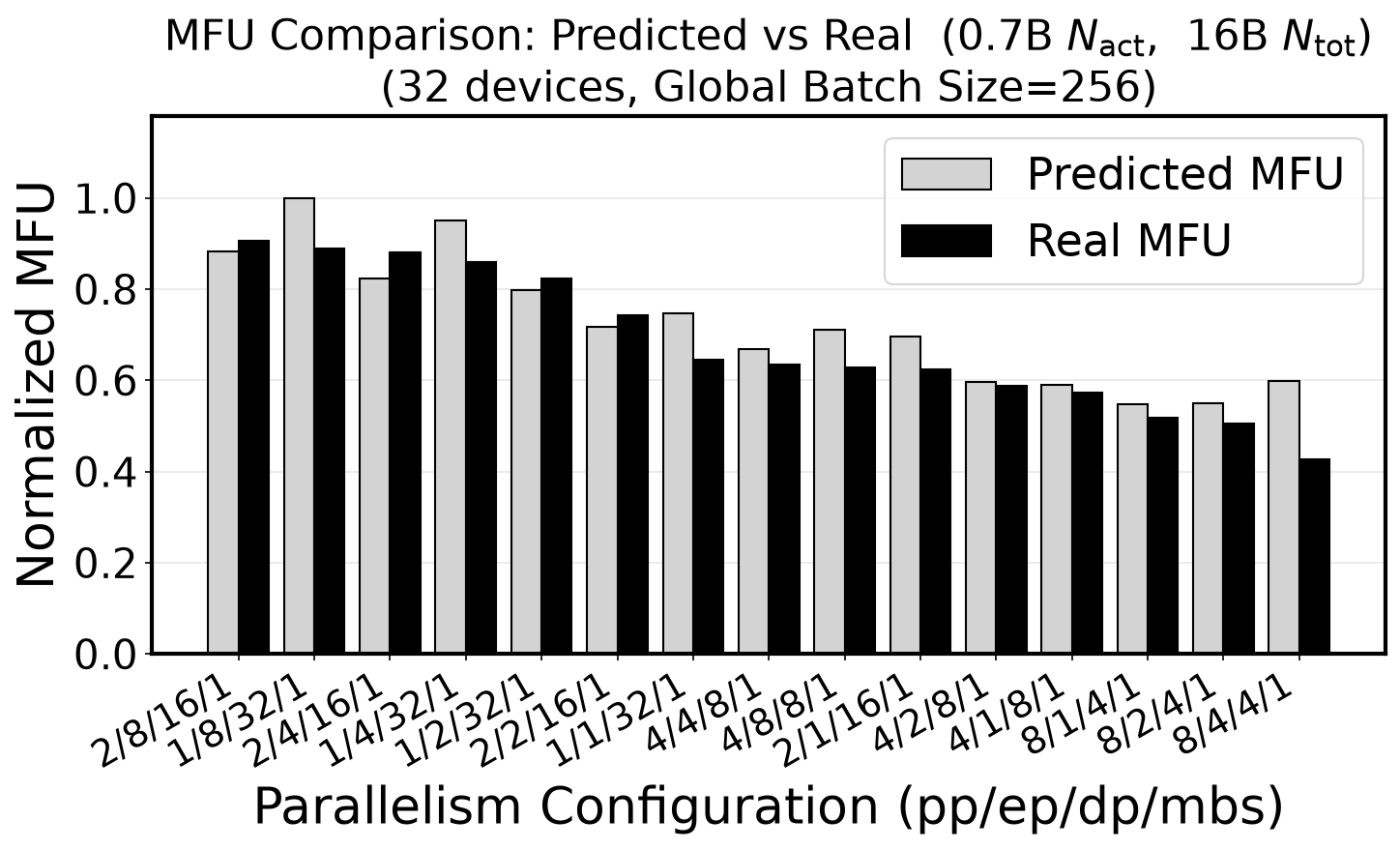}
    \label{fig:mfu-700m}
  \end{subfigure}
  \hspace{0.03\textwidth}
  \begin{subfigure}{0.45\textwidth}
    \includegraphics[width=\linewidth]{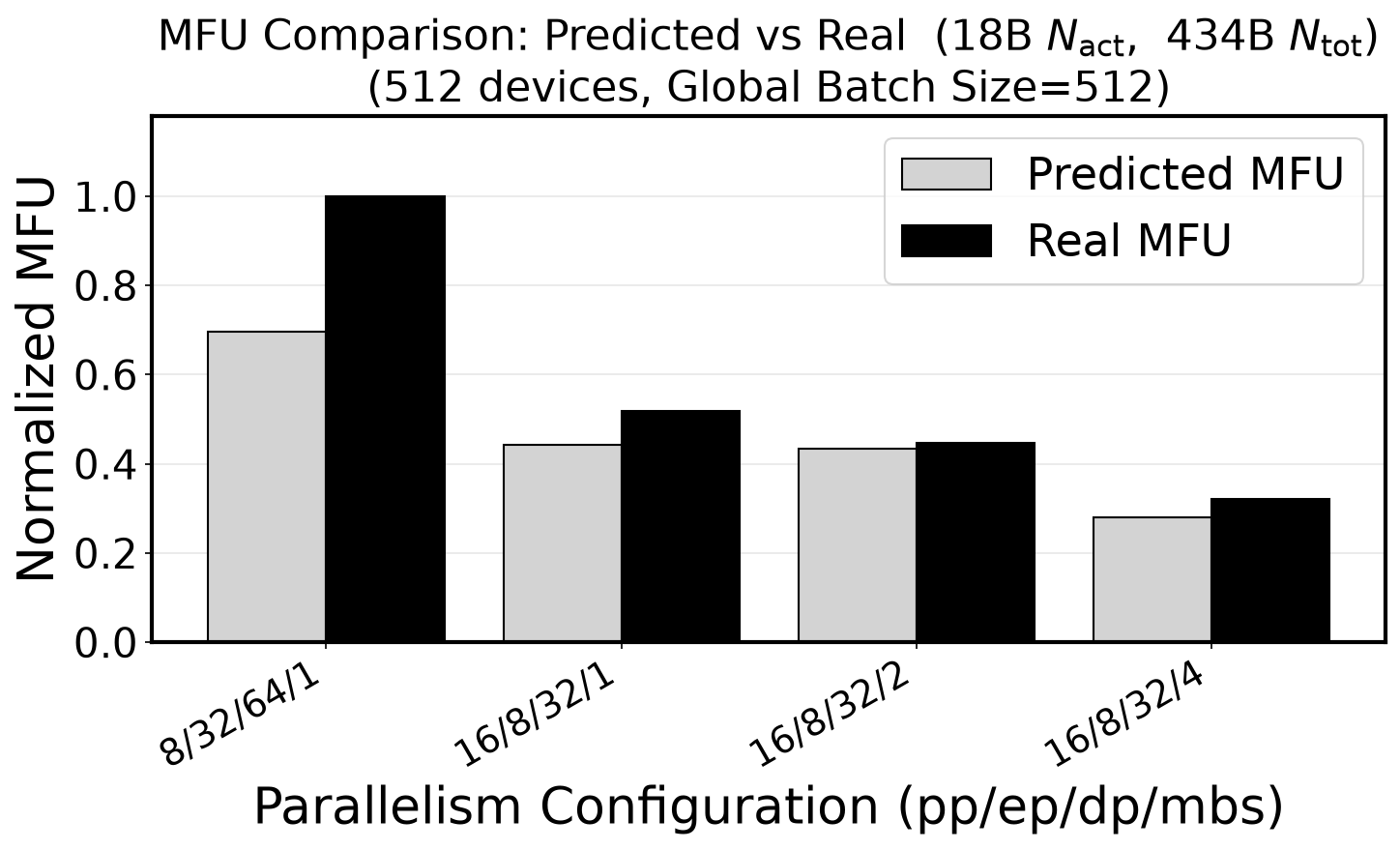}
    \label{fig:mfu-18b}
  \end{subfigure}
  \vspace{0.6em}
  \begin{subfigure}{0.45\textwidth}
    \includegraphics[width=\linewidth]{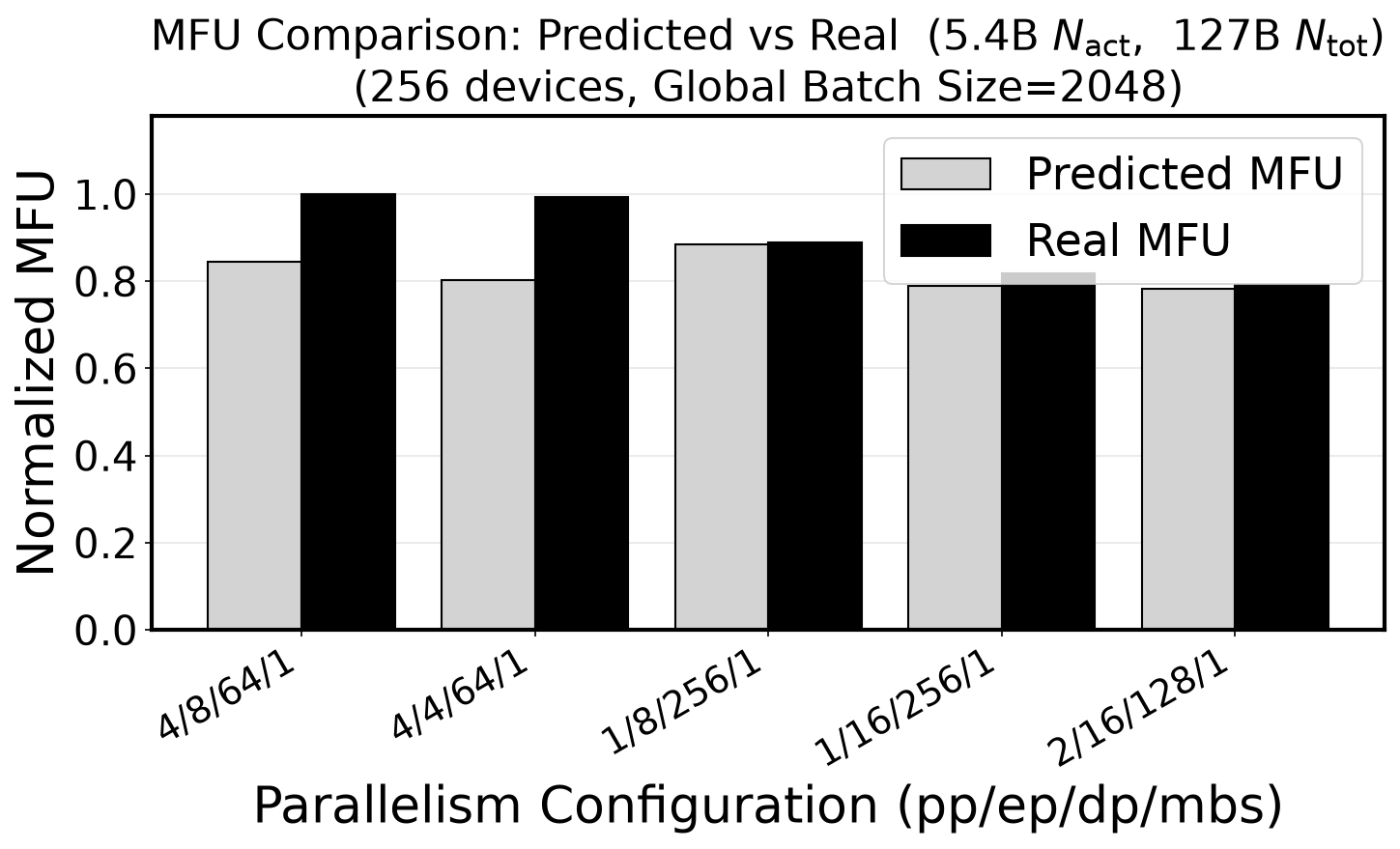}
    \label{fig:mfu-5p3b-32}
  \end{subfigure}
  \hspace{0.03\textwidth}
  \begin{subfigure}{0.45\textwidth}
    \includegraphics[width=\linewidth]{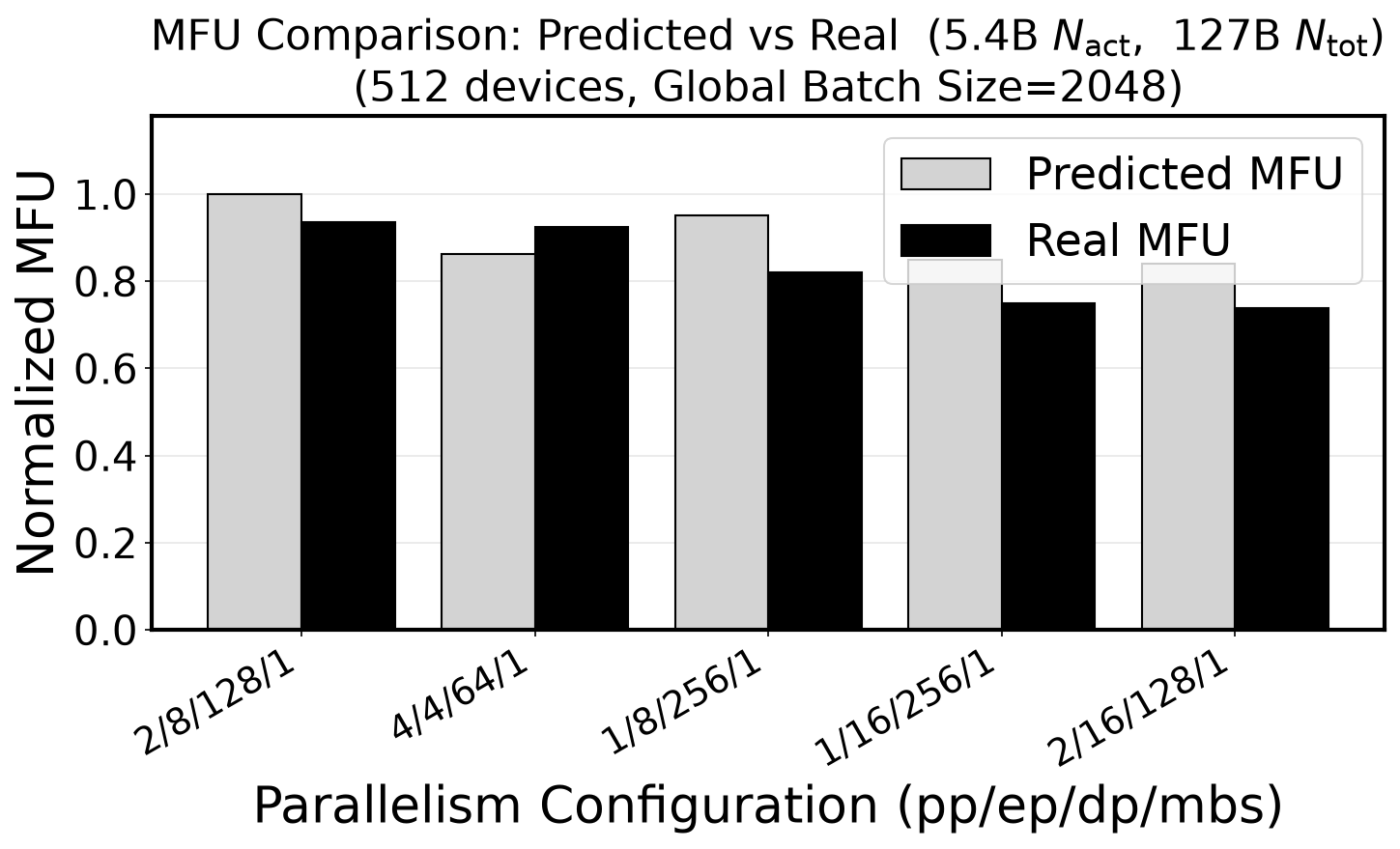}
    \label{fig:mfu-5p3b-64}
  \end{subfigure}
  \caption{Predicted versus Real MFU per parallelism configuration, each
  panel normalized to its own maximum and sorted by real MFU.}
  \label{fig:mfu}
\end{figure}

Our validation spans three sweeps on these p6 nodes, over the $q\!=\!4$, $q\!=\!8$, and $q\!=\!12$ rungs of Table~\ref{tab:scale-ladder}. The $q\!=\!4$ model ($0.7$B $N_{\text{act}}$, $16$B $N_{\text{tot}}$) runs on 4 nodes ($32$ GPUs), the $q\!=\!8$ model ($5.4$B $N_{\text{act}}$, $127$B $N_{\text{tot}}$) on 32 nodes ($256$ GPUs) and 64 nodes ($512$ GPUs), and the $q\!=\!12$ model ($18$B $N_{\text{act}}$, $434$B $N_{\text{tot}}$) on 64 nodes ($512$ GPUs).
Each data point pairs one parallelism configuration, namely expert (ep), pipeline (pp), data (dp), tensor (tp), context (cp) parallel degree and micro batch size (mbs) with its measured MFU and its predicted value. The MFU prediction results are shown in Figure~\ref{fig:mfu}. We vary the global batch size and restrict $tp=1$ and $cp=1$ for efficiency.
We release the estimator as
ScalePlan\footnote{\url{https://github.com/dmlc/ScalePlan}}, which packages the
analytical cost model, the calibration data it is fit against, and the
parallelism search that realizes the inner maximization over
$\mathcal{P}_{\text{feas}}$ of
Equation~\eqref{eq:c-flops}, so the predictions validated here and every
estimator number reported later can be regenerated from a single model and
cluster specification.

\subsubsection{MFU Prediction}
\label{sec:mfu-pred}
We find that the predicted MFU values are accurate in absolute terms, not just in ordering, across all three sweeps. The mean absolute percentage error is under $15\%$ for every model
($9.6\%$/$9.0\%$/$13.2\%$ for 700M/5.4B/18B). The estimator also captures the top region of the configuration space reliably. Ranking configurations by predicted MFU and
comparing against the measured ranking, Top-5 overlap is $100\%$ for every 5.4B
and 18B sweep and $80\%$ for the 700M sweep. Top-10 overlap is $100\%$ for all
5.4B and 18B sweeps and $90\%$ for 700M. The single 700M miss is a swap among
near-tied low-MFU configurations, so the model reliably surfaces the
high-throughput region even where it mis-orders adjacent points.

\subsubsection{Implications for Sparse-MoE Parallelism Design}
\label{sec:moe-implications}
The MFU sweeps expose a consistent structure in how a sparse MoE responds to
its parallelism axes, which we develop into
design rules in Appendix~\ref{app:moe-implications}. The dominant lever is not any single degree but the pipeline fill
ratio $r_{\text{fill}} = n_{\text{mb}}/P_{\text{PP}}$, which explains almost
all of the up to $3.1\times$ layout spread in measured MFU, with throughput
collapsing as $r_{\text{fill}}$ falls toward $1$ and saturating for
$r_{\text{fill}} \ge 4$. The expert all-to-all belongs inside the scale-up
domain, since expert parallelism helps up to the node boundary and hurts past
it, and the FFN-parallelism budget is better spent on expert than expert
tensor parallelism. In practice we fix $P_{\text{TP}}{=}1$, set
$P_{\text{EP}}$ to the scale-up domain, and choose the
$(P_{\text{PP}}, B_{\text{micro}})$ pair that keeps $r_{\text{fill}} \ge 4$,
stated as the MFU-optimal layout for 700M\textendash18B models on NVL8 under
no memory pressure rather than as a hardware-independent law.

\subsection{Optimal Allocation Under $C_{\text{peak}}$}
\label{sec:minlp-implementation}

In this section, we first show the configuration MOSAIC finds optimal
under a single fixed GPU-hour budget, and then extend it to the
loss-versus-deliverable-model-FLOPs spread over all feasible configurations.
The inputs to MOSAIC are the training window (wall-clock days),
the device type and its per-device peak FLOPs and memory, and the device
count, which together set the GPU-hour budget, alongside the global and
sequence batch.

\begin{table}[h]
\centering
\small
\setlength{\tabcolsep}{6pt}
\begin{tabular}{@{}lp{0.62\linewidth}@{}}
\toprule
\multicolumn{2}{@{}l}{\textbf{Inputs}} \\
\midrule
Hardware & 32 p6 nodes ($N_{\text{dev}} = 256$ devices) \\
Per-device peak $F_{\text{peak}}$ & $2.25$\,PFLOPs (bf16/dense) \\
Wall-clock $T_{\text{train}}$ & 20 days \\
Per-device memory cap $M_{\text{dev}}$ & $179$\,GB \\
Goodput $\eta_{\text{good}}$ & $1.0$ \\
Raw hardware ceiling $C_{\text{peak}}$ & $9.95\times10^{23}$ FLOPs \\
Global / sequence batch & $B_{\text{global}}=1024$, $T_{\text{seq}}=2048$ \\
Loss law & joint $\mathcal{L}(N_{\text{tot}}, S, D, G)$ from Eq.~\eqref{eq:joint-law} \\
\midrule
\multicolumn{2}{@{}l}{\textbf{Search space (output)}} \\
\midrule
Outer architecture grid & $q,  E,  K, G$ \\
Inner layout search & over $P_{\text{TP}}$, $P_{\text{CP}}$, $P_{\text{EP}}, P_{\text{PP}}, P_{\text{DP}}$, 
                    $B_{\text{micro}}\!\in\!\{1,2,4\}$, and
                    activation-checkpointing $\in\!\{\text{none, selective, super-selective, full}\}$ \\
\bottomrule
\end{tabular}
\caption{MOSAIC search envelope. The
outer grid over $(q, E, K, G)$ is the discrete realization of $Z$, and the
inner parallelism search realizes the embedded maximization over $\mathcal{P}_{\text{feas}}$
in Equation~\eqref{eq:c-flops}.}
\label{tab:results-envelope}
\end{table}

We target a  deployment simulation with 32 p6 nodes
($N_{\text{dev}} = 256$ devices) and
$T_{\text{train}} = 20\,\text{days}$.  These inputs are listed in
Table~\ref{tab:results-envelope}. The model FLOPs the returned optimum realizes
under this ceiling lie about $150\times$ beyond our
largest fit run, an extrapolation the deployed law supports for the
configuration ordering.  Combining the scaling ladder of Table~\ref{tab:scale-ladder} with the sparse
MoE axes, we restrict the free discrete search dimensions to
\begin{align*}
q &\in \{8, 10, 12, 14, 16, 18\}, &
K &\in \{1, 2, 4, 6, 8\}, \\
E &\in \{4, 6, 8, 16, 24, 32, 48, 96, 128, 192, 256, 384, 512\}, &
G &\in \{1, 2, 4, 8\}.
\end{align*}
We add a practical geometry guard to prune the
grid by setting $S \ge 0.7$, since the scaling law data we collect to fit
the loss law lies predominantly in the $S \ge 0.7$ region, so this keeps
the search inside the sparsity range the law is calibrated on rather than
extrapolating below. The experimental envelope for this section can be found in Table~\ref{tab:results-envelope}. We  fix the token budget $D$ per configuration by sweeping the tokens-per-parameter ratio $\mathrm{TPP} = D/N_{\text{act}}$ over the grid
$\mathrm{TPP} \in [40, 150]$. Our training data spans TPP up to $70$
(Table~\ref{tab:dataset-headline}), so the cap sits about twice beyond the
calibrated range, and we keep only $(Z, \mathrm{TPP})$ pairs whose
$C_{\text{model}}$ fits the deliverable model FLOPs $C_{\text{deliverable}}$ at that layout's MFU together with
the memory and parallelism constraints.

\subsubsection{Results}

MOSAIC returns a
single loss optimal configuration, reported in Table~\ref{tab:pareto-B}.
The constrained optimum is a $q\!=\!12, L_{\text{layers}}\!=\!48, E\!=\!96, K\!=\!2, G\!=\!4$
model with $N_{\text{act}}\!=\!14.5$\,B at sparsity $S\!=\!0.956$,
running at $12.4\%$ predicted MFU and loss $\mathcal{L}\!=\!1.3898$.
Its selected layout is $P_{\text{PP}}{=}8$, $P_{\text{EP}}{=}8$,
$P_{\text{CP}}{=}1$, $P_{\text{TP}}{=}1$ with micro-batch $B_{\text{micro}}{=}4$.
\begin{table}[h]
\centering
\small
\setlength{\tabcolsep}{5pt}
\renewcommand{\arraystretch}{0.95}
\resizebox{\textwidth}{!}{%
\begin{tabular}{@{}rrrrrrrrrrrrrrr@{}}
\toprule
\rowcolor{headergray}
$L_{\text{layers}}$ & $E$ & $K$ & $G$ & $N_{\text{act}}$ (B) & $S$ & TPP & $D$ (T) & MFU (\%) & Loss & pp & ep & mb & cp & tp \\
\midrule
48 & 96 & 2 & 4 & 14.5 & 0.956 &  98   & 1.42 & 12.38 & \textbf{1.3898} & 8 & 8 & 4 & 1 & 1 \\
\bottomrule
\end{tabular}}
\caption{The loss optimal configuration MOSAIC returns under the
envelope of Table~\ref{tab:results-envelope}. The MFU column is the predicted
peak MFU of the best feasible execution plan, not a measured value.}
\label{tab:pareto-B}
\end{table}

\begin{figure}[!t]
\centering
\includegraphics[width=0.55\linewidth]{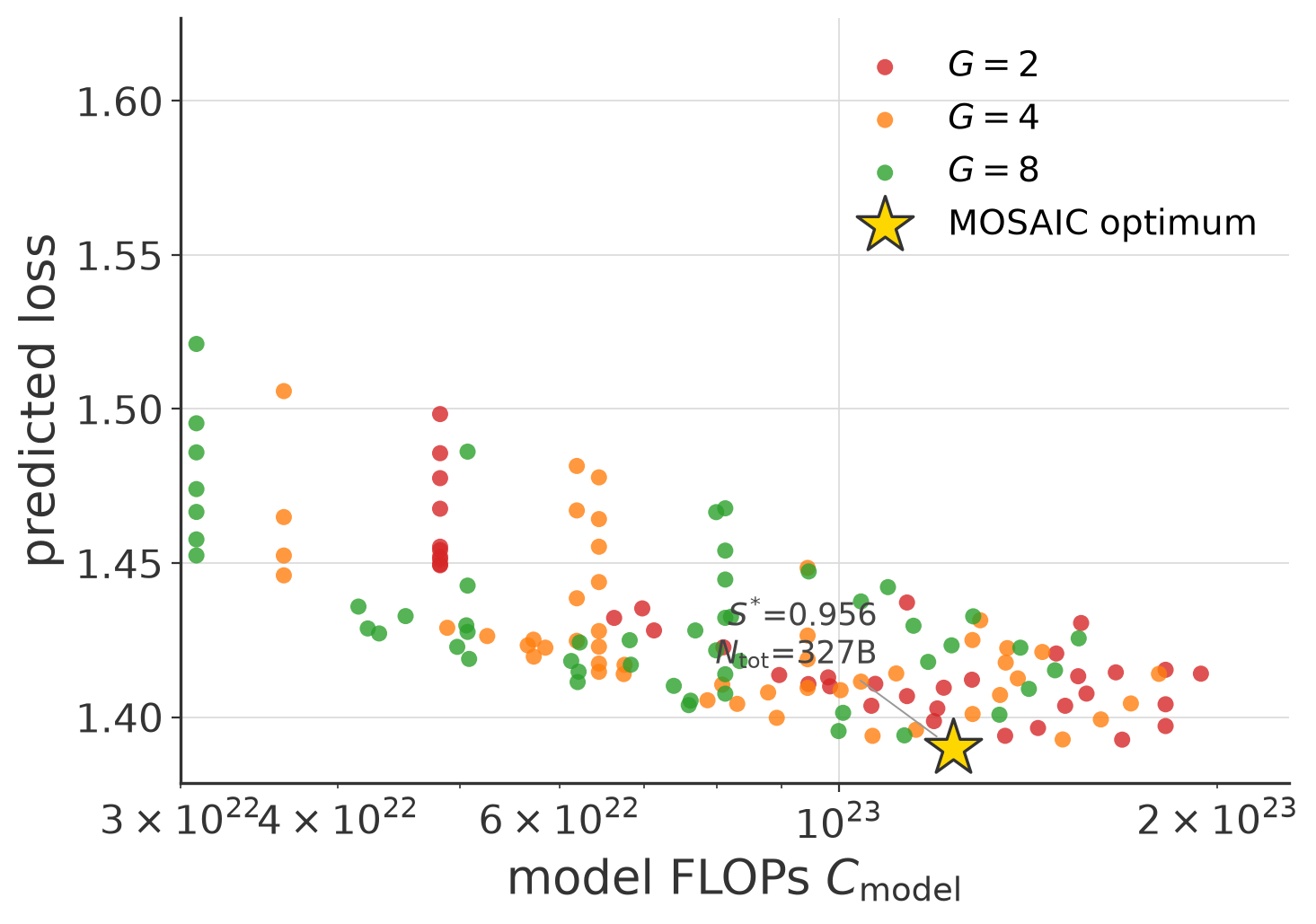}
\caption{Loss versus model FLOPs $C_{\text{model}}$ over all
feasible $(Z, D)$ configurations under the $32$-node, $20$-day envelope,
color-coded by $G$. Every candidate draws on the same raw hardware ceiling
of $9.95\times10^{23}$ hardware FLOPs. The gold star is the optimal
configuration with MOSAIC, and the candidate that realizes
the most model FLOPs is not the one that reaches the lowest loss.}
\label{fig:pareto-B}
\end{figure}

Although Table~\ref{tab:pareto-B} reports only the loss optimal configuration, we
now show all feasible $(Z, D)$ configurations
under the same envelope in Figure~\ref{fig:pareto-B}. While $D$ is selected on a capped TPP grid, the cap is non-binding for most configurations near the low-loss frontier, whose selected token counts consume nearly the full 20-day wall-clock. Near the low-loss frontier, we observe that the geometry lands farther right primarily by
sustaining a higher MFU, which converts more of the fixed hardware ceiling
into useful model FLOPs. The fixed cluster and
training window emit the same raw hardware ceiling of $9.95\times10^{23}$ FLOPs for
every candidate, so the quantity that varies is how much of that ceiling each
geometry converts into model FLOPs through its own MFU.
Which is what we show in Figure~\ref{fig:pareto-B}. It plots the loss
against the model FLOPs $C_{\text{model}}$ each candidate realizes, color-coded
by $G$, with the gold star marking the loss optimal configuration.

Because the raw hardware ceiling is the same for every candidate, a geometry
lands further right in Figure~\ref{fig:pareto-B} only by sustaining a
higher MFU, which turns the fixed $20$ days of wall-clock into more useful
model FLOPs. The expert split factor $G$ orders that spread. The
$G\!=\!8$ geometries sit at the low $C_{\text{model}}$ end at a median MFU near
$8\%$, and the coarser $G\!=\!2$ geometries reach the right end near
$13$\textendash$20\%$, so a higher $G$ delivers less useful compute out of
the same ceiling because its expert dispatch and combine communication
leaves fewer FLOPs for the model, a penalty tied to the Megatron-Core stack
our performance model mirrors (Appendix~\ref{app:training-stack}). Model FLOPs alone would read this as the
finest experts being cheapest, but under the fixed hardware ceiling, 
they are the ones the cluster feeds the least.

The consequence is that the run with the most model FLOPs is not the optimal as shown in Figure~\ref{fig:pareto-B}. The candidate that converts the shared hardware ceiling $C_\text{peak}$ into the most model FLOPs is
a denser $G\!=\!2, E\!=\!32$ geometry at $1.94\times10^{23}$ FLOPs, and it still lands
at a higher loss than the optimal configuration. The loss optimal configuration is a
$G\!=\!4, E\!=\!96, K\!=\!2$ model at $N_{\text{act}}\!=\!14.5$B and
$S\!=\!0.956$ that reaches $\mathcal{L}\!=\!1.3898$ on $1.23\times10^{23}$ model
FLOPs, roughly $36\%$ fewer than the rightmost candidate spends. Ranking these
candidates by the model FLOPs they realize therefore misorders them, and the
split factor and sparsity that win are set only once the shared deliverable
model-FLOPs budget is imposed. This is the optimum under deliverable
rather than model FLOPs.



\subsection{Optimal $S$ and $G$}
\label{sec:results-gpuhours}

The frontier in the previous section fixed a single cluster envelope. We now ask how the
optimal architecture moves as the deliverable model-FLOPs budget grows, specifically whether the
MoE configurations get sparser and finer, or denser and coarser under such a constraint. To that end, we run two budget sweeps that share a common reference point.
The first fixes the cluster at $32$ p6 nodes
($256$ GPUs) and varies the wall-clock window over
$\{10, 20, 30, 40\}$ days. The second fixes the window
at $30$ days and varies the cluster over
$\{16, 32, 64, 128\}$ p6 nodes, re-running the performance model
at each node count. In both sweeps, we run MOSAIC subject to these constraints, and for every budget we record
the architecture that minimizes predicted loss. 

\begin{figure}[!t]
\centering
\begin{subfigure}[t]{0.5\textwidth}
  \centering
  \includegraphics[width=\linewidth]{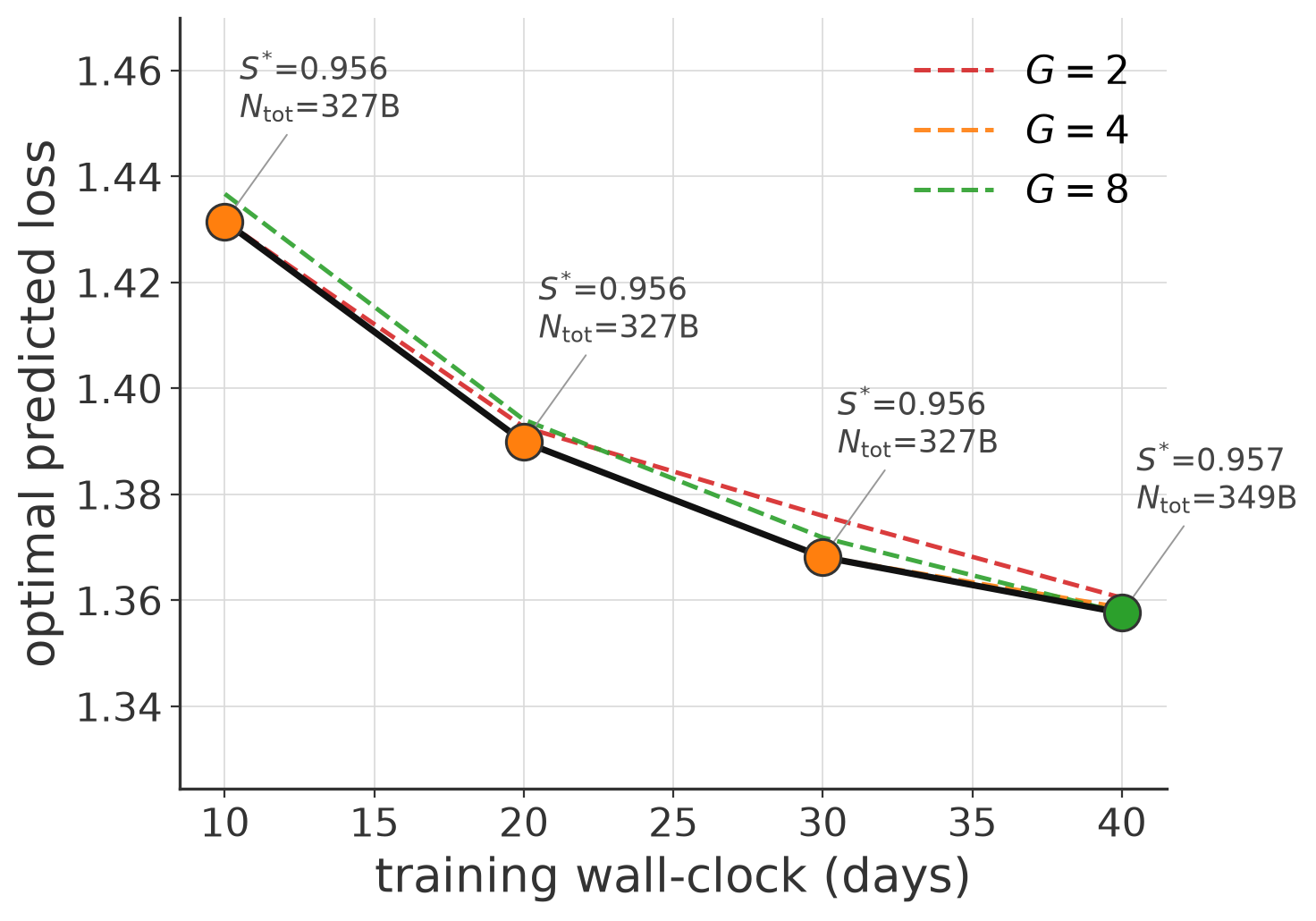}
  \subcaption{$32$ p6 nodes.}
  \label{fig:gpuhours-days}
\end{subfigure}\hfill
\begin{subfigure}[t]{0.5\textwidth}
  \centering
  \includegraphics[width=\linewidth]{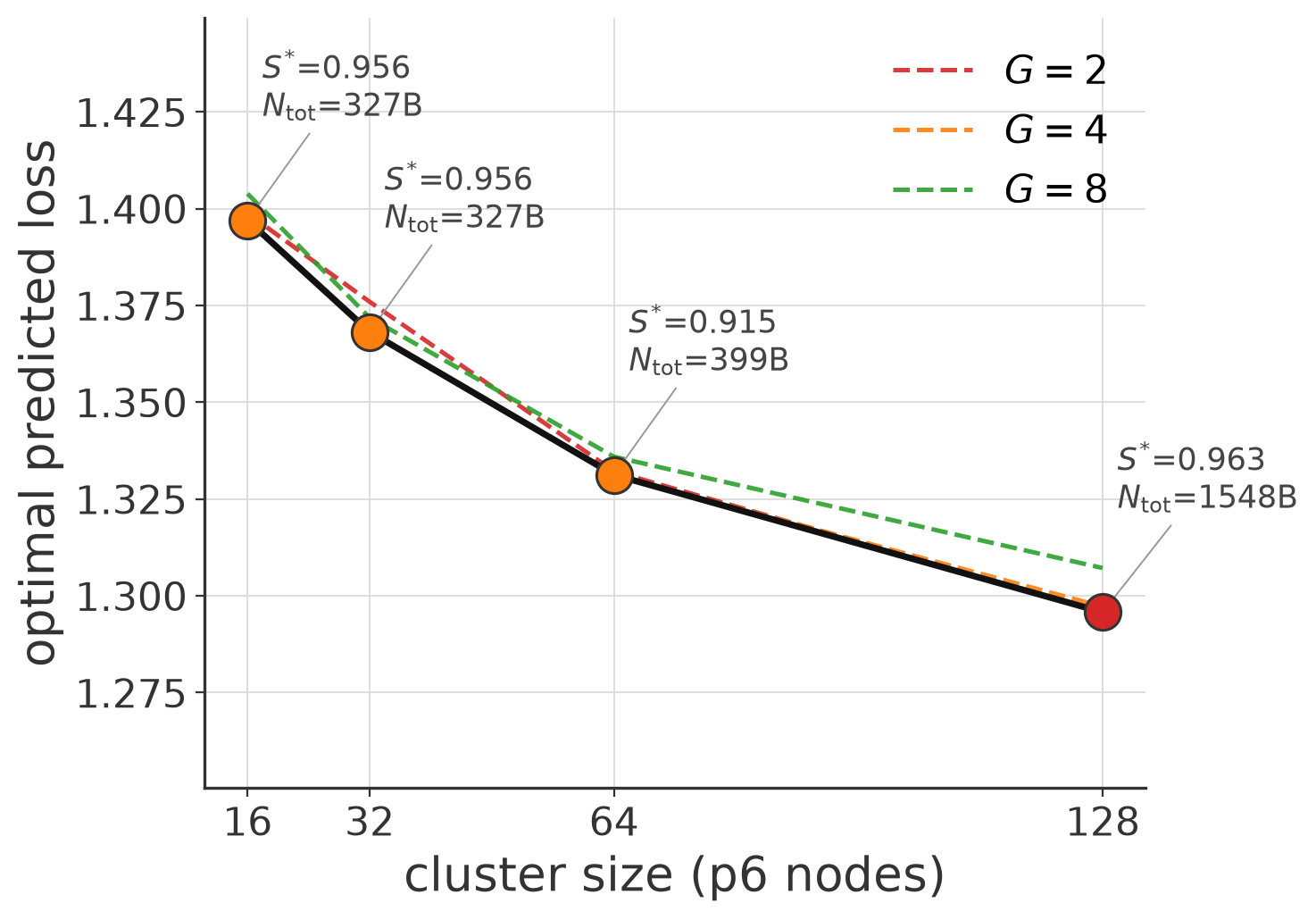}
  \subcaption{$30$ days.}
  \label{fig:gpuhours-nodes}
\end{subfigure}
\caption{Optimal predicted loss vs.\ the budget under the two sweeps.
Markers are colored by the optimal expert split factor
$G^{\star}$ and annotated with $S^{\star}$ and
$N_{\text{tot}}^{\star}$. Dashed lines denote
 best per $G$ configuration. The solid line is the overall optimum.
The optimal sparsity $S^{\star}$ does not move monotonically with the budget in
either sweep, staying on a wide plateau of roughly $0.915$ to $0.963$.}
\label{fig:gpuhours}
\end{figure}

We plot the results from running MOSAIC in Figure~\ref{fig:gpuhours}. We observe
that $G^{\star}$ is $4$ across most budgets, with $G\!=\!8$ appearing only at
the longest window and $G\!=\!2$ at the largest node budget, and the exact
$G^{\star}$ marker is often a near-tie.
Looking at the span of sparsity, we find that $S^{\star}$ sits on a wide
plateau (roughly $0.915$\textendash$0.963$) and does not sharpen with
budget, confirming that there is no monotone sparsity trend here unlike the model FLOPs optimality we find in Figure~\ref{fig:figure3-abnar-repro}. Finally, $N_{\text{act}}^{\star}$ moves slowly with budget, staying near
$14$\textendash$15$\,B parameters across the days range and growing to $34$\,B and
$58$\,B parameters at the two largest node budgets. The direction of this trend is that larger hardware budgets support larger
active parameter counts, though the near-ties at neighboring budgets mean
the exact crossover points are not
resolved at this granularity. These findings suggest that an interior optimum in sparsity appears once the systems constraints are added, where a memory floor bounds total
parameters and a GPU-hour cost charges for the MFU degradation and
all-to-all penalties of high $S$ and high $G$. We note that this avoids any explicit cap on model
dimensions, since the systems constraints alone bound the
feasible region and let an interior sparsity emerge.


\subsection{From Predicted Ranking to Measured Loss}
\label{sec:results-staged}

In this final experiment, we test whether the configuration ordering
predicted by MOSAIC transfers to real training runs. We take a small budget
envelope of eight p6 nodes ($64$ GPUs) for $10$ days, enumerate the
feasible grid, and select four configurations from the ranked
predictions. These four training configurations span the same trade-off MOSAIC
describes, and
we launch each as a real pretraining run.
Table~\ref{tab:staged-top4} lists the selected configurations with the
MFU the framework predicts for each.

\begin{table}[h]
\centering
\small
\setlength{\tabcolsep}{5pt}
\renewcommand{\arraystretch}{0.95}
\resizebox{\textwidth}{!}{%
\begin{tabular}{@{}rrrrrrrrrrrrrrr@{}}
\toprule
\rowcolor{headergray}
rank (loss) & $L_{\text{layers}}$ & $E$ & $K$ & $G$ & $N_{\text{act}}$ (B) & $N_{\text{tot}}$ (B) & $S$ & pred.\ MFU (\%) & norm.\ actual MFU & pp & ep & cp & tp & mbs \\
\midrule
ID 1 & 40 & 128 & 2 & 4 & 8.5 & 250.0 & 0.966 & 14.0 & 0.84 & 4 & 8 & 1 & 1 & 4 \\
ID 2 & 40 & 128 & 2 & 8 & 5.6 & 126.4 & 0.956  & 11.6 & 0.69 & 4 & 8 & 1 & 1 & 4 \\
ID 3 & 40 &  32 & 2 & 4 & 8.5 & 66.0 & 0.872 & 21.8 & 0.91 & 4 & 8 & 1 & 1 & 4 \\
ID 4 & 32 &  32 & 2 & 2 & 7.3 & 65.8 & 0.889 & 23.3  & 1.0 & 4 & 8 & 1 & 1 & 4 \\
\bottomrule
\end{tabular}}
\caption{The four configurations on the
$8$-node, $10$-day envelope, ranked by predicted loss. We keep the global batch size fixed to 1024 across all runs and assign each configuration’s peak learning rate using Equation~\eqref{eq:lr-compute} in Appendix~\ref{app:lr_hpo}.
The normalized actual MFU column reports each run's measured MFU divided by the
largest measured MFU among these four runs. }
\label{tab:staged-top4}
\end{table}

The first test is whether the performance model orders these
configurations the way the hardware does. Our measured MFU
reproduces this ordering exactly, where the rank correlation between predicted
and observed MFU across the four runs aligns as shown in Table~\ref{tab:staged-top4}.
Although MOSAIC relies on the absolute MFU values, we discuss the MFU ordering here to
test the selection step itself, since these four configurations were chosen by a
predicted ranking rather than by any single MFU number, and
we already established in Section~\ref{sec:mfu-pred} the absolute accuracy on which the
feasibility constraint and the token budget rest. Normalizing the predicted MFU the same way gives $0.60$, $0.50$, $0.94$, and $1.00$ against the measured $0.84$, $0.69$, $0.91$, and $1.00$, and we observe that the predicted spread is
wider than the measured one even though the ordering agrees.

\begin{figure}[t!]
\centering
\includegraphics[width=0.9\linewidth]{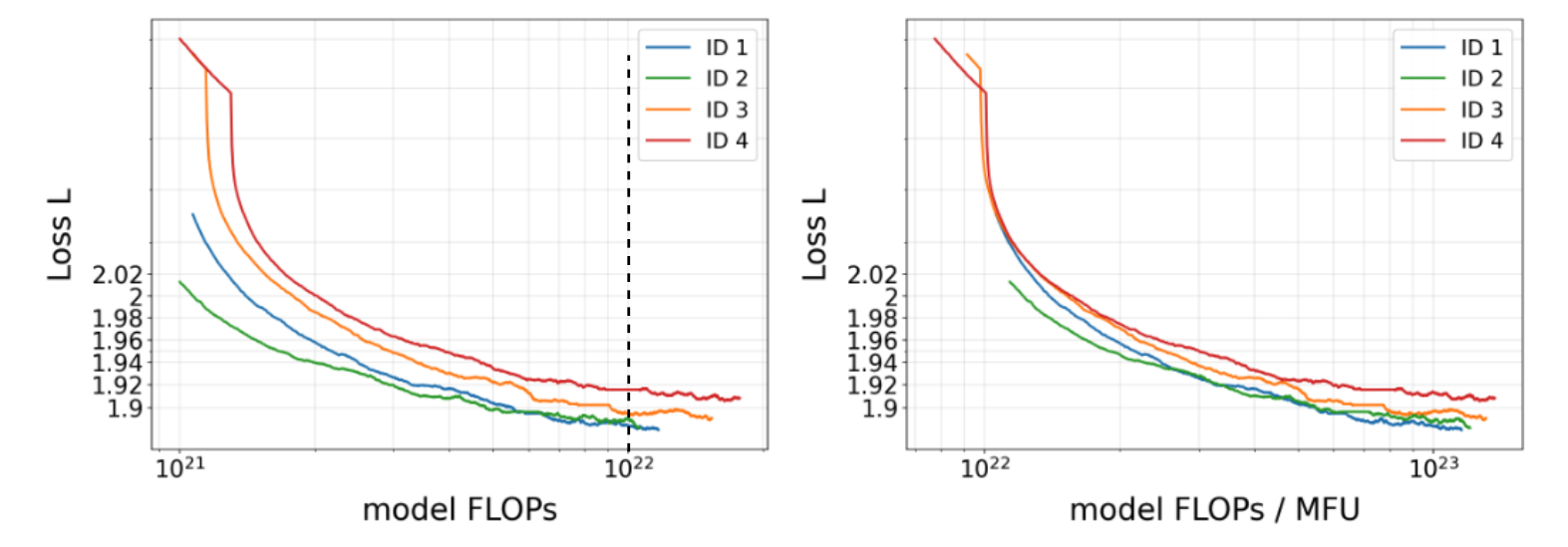}
\caption{Smoothed loss trajectories of the staged runs against model FLOPs
$C_{\text{model}}$ (left) and against the peak-equivalent
hardware compute consumed $C_{\text{model}}/\mathrm{MFU}$ (right),
with the warm-up discarded. ID~2 leads over much of the lower model-FLOPs range, while the trajectories
converge and ID~1 slightly overtakes it near the end (at $10^{22}$ FLOPs) within the width of the
smoothed curves. On the hardware-compute axis, the advantage of ID~1 over ID 2 
appears much earlier.}
\label{fig:staged-curves}
\end{figure}

We next examine the loss trajectory of the runs themselves, but where we plot against both model FLOPs axis and an axis that accounts for the efficiency in MFU. The eight-node, ten-day envelope is
only the selection envelope MOSAIC ranks against, and the staged runs train
beyond it. The hardware-compute axis of Figure~\ref{fig:staged-curves} divides
by the measured MFU of each run.

The largest model FLOPs these curves reach is roughly $15$\textendash$20\times$ the
largest budget in the scaling law fitting set of Appendix~\ref{app:dataset}, which
tops out at $8.1\times10^{20}$ FLOPs, so the runs probe the predicted ordering well
outside the regime the law was fit on.
We plot the loss of the four runs against model FLOPs and against the peak-equivalent hardware
compute consumed in Figure~\ref{fig:staged-curves}, with the warm-up discarded.
ID 3 and ID 4 isolate the expert split factor, since they sit at nearly the same
total parameters ($66.0$ against $65.8$\,B) and close sparsity ($0.872$ against
$0.889$), yet the $G\!=\!4$ run stays below the $G\!=\!2$ run on both axes.
On the per-FLOP axis, up to roughly
$10^{22}$ model FLOPs the $G\!=\!8$, low-$N_{\text{act}}$ configuration
(ID 2) holds an advantage over ID 1, since its smaller active count
($5.6$\,B against $7$\textendash$8.5$\,B) shifts its curve toward fewer model
FLOPs to reach a given loss. Past that point the two curves converge and the
margin narrows to the width of the smoothed trajectories, so we read the
advantage as a tendency over the budget range we ran rather than a separation
that holds at the end of training. Even so, the direction matches the
compute-frontier analysis, where the run slowest in wall-clock terms converts
a fixed FLOPs budget into loss most efficiently.

The observation in Figure~\ref{fig:staged-curves} changes once the same trajectories are placed on a hardware-compute
axis rather than a model-FLOPs one. We use the peak-equivalent hardware compute
consumed, $C_{\text{cost}} = C_{\text{model}}/(\mathrm{MFU}\,\eta_{\text{good}})$,
the compute proportional to the GPU-hours a run spends to deliver
$C_{\text{model}}$, and we hold $\eta_{\text{good}} = 1$ throughout, so
$C_{\text{cost}}$ reduces to $C_{\text{model}}/\mathrm{MFU}$ as plotted.
Dividing out MFU charges each run for the hardware time it actually spends,
and ID 1 sustains the higher MFU (normalized $0.84$ against ID 2's $0.69$),
so the same hardware buys ID 1 about $1.2\times$ the model computation.
Figure~\ref{fig:staged-curves} shows this
directly, where ID 2 leads on the model-FLOPs panel (left) while ID 1 leads
on the $C_{\text{cost}}$ panel (right) toward the longer horizon of the run.


Taken together, the staged runs provide two observations. First, the predicted
MFU ranking reproduces on hardware, so the selection that MOSAIC relies on
transfers from those predictions. Second, the ordering between the two leading configurations changes between
the model-FLOPs and hardware-cost views, which is the co-design effect
MOSAIC is designed to capture. We report these as an early
validation of our framework. The results above optimize one hardware envelope at a time. In Appendix~\ref{app:allocation}, we ask whether the discrete MOSAIC optima can be summarized by a Chinchilla-style optimal resource-allocation law that
maps a purchased hardware budget to the jointly optimal model and data allocation. For a fixed architecture, the reduction is exact. The hardware budget
determines its optimal token count and therefore its predicted loss.
This also gives a simple phase structure, where the preferred architecture
can change as the training budget grows. So we present the analysis as a
direction for future work rather than as a main empirical result.
\section{Related Work}
\label{sec:related-work}

\paragraph{MoE Scaling Laws.} Past work on sparse MoE architecture extends
Chinchilla-style scaling laws~\citep{hoffmann2022training} to sparse
MoEs. \citet{krajewski2024scaling} introduce granularity as an explicit
axis and \citet{abnar2025parameters} characterize the
sparsity dimension, while \citet{ludziejewski2025joint} fit a joint law
over active size, experts, and tokens and argue MoEs can be more
memory-efficient than dense models.
\citet{wan2026holistic} push toward holistic recipes that map a budget
to a full architecture, and \citet{mcleish2025gemstones} show how
architecture and hyperparameter choices change the prescriptions a
scaling law fit yields. Our scaling law work extends these previous studies but isolates the
MoE-layer geometry and is coupled to a systems cost model rather than
read under FLOPs alone. There have been few more notable studies in the arena of scaling laws for MoE models that focus on multiple dimensions \citep{tian2026towards}, however, as discussed in our desiderata, we focus on low dimensional MoE scaling laws in this work that suit the systems derivation. 

\paragraph{Systems Performance Estimation.} Measuring realized throughput of distributed training rather than
its FLOPs is central to the thesis of MOSAIC. While this usually falls under systems performance engineering in most works, we focus mainly on modeling and not on engineering the systems like writing better kernels. To that end, \citet{fernandez2024hardware} measure diminishing returns and
parallelism strategy crossovers empirically, while analytical co-design
tools such as Calculon~\citep{isaev2023calculon} and profiling driven
simulators such as vTrain~\citep{bang2024vtrain} predict training time
and memory from a model and layout description.
\citet{bian2025inference} target inference-efficient model selection,
and on the MoE side \citet{huang2023towards} and \citet{zheng2026uniep}
attack expert dispatch, all-to-all, and kernel inefficiencies in
deployment and training. These efforts model or optimize execution cost in isolation.
We instead fold an operator-level MFU and memory predictor directly into
the architecture-selection objective.

\paragraph{Joint Architecture-Systems Co-Design.} Finally, we discuss the
relevant studies that optimize model and hardware together.
\citet{anthony2024codesign} align model dimensions to GEMM and
wave-quantization boundaries whereas \citet{bian2026architecture} jointly tune
architecture for accuracy and inference efficiency. \cite{bae2025hybrid} studies hybrid architectures for compute optimality that are better on recall and 
\citet{wang2025catransformers} co-design transformers and accelerators
under a carbon objective. We also see studies like \citet{zhang2025ladder} which redefine
tensor parallelism via a Ladder Residual architectural change that
overlaps communication with computation for accelerated inference, a
complementary form of architecture-systems co-design that adapts the
model to the parallel-execution substrate rather than the GEMM tile.
We also find that classic budget studies such as
\citet{lescao2022million} ablate architecture at fixed GPU-hours, and
\citet{wu2024inference} add an inference-compute axis. We differ by
coupling a predictive MoE scaling law to a parallelism-aware throughput
model, recovering deployable MoE geometries under explicit memory and
GPU-hour constraints rather than FLOPs. 

\section{Conclusion, Limitations and Future Work}

We introduced MOSAIC, a systems-aware scaling framework that jointly selects the model architecture, training-token budget, and distributed parallel layout for a fixed cluster and training window.
Using roughly 150 from-scratch sparse MoE pretraining runs, we fit a joint scaling law over total parameters, sparsity, training tokens, and expert split factor, and show that model-FLOPs only optimization admits no interior sparsity optimum within the calibrated sparsity region.
To enable the joint systems optimization, we developed an operator-level performance model that predicts MFU, memory, and the best parallel layout across models ranging from 700 million to 18 billion active parameters, while reliably recovering the highest-throughput configurations.
We observe that coupling the scaling law and performance model changes the architecture prescription. Notably, the optimal sparsity becomes interior and hardware-dependent rather than following the monotone boundary preference implied by model FLOPs, while the expert geometry shifts with the cluster size and training horizon.
Our budget sweeps recover interior, hardware-dependent MoE configurations, and our staged pretraining runs reproduce the predicted MFU ordering on real hardware, with the loss ordering of the runs flipping between the model-FLOPs axis and the hardware-compute axis.

One limitation of MOSAIC is that the precise architectures returned are conditional on the chosen geometry ladder, hardware platform, and training recipe, and some reported optima require substantial extrapolation beyond the scaling law training regime. Moreover, uncertainty in the non-identified scaling law coefficients is not propagated through the discrete architecture search. Consequently, our evidence more strongly supports the hardware-aware ranking and co-design principle than any single geometry as a universal cluster optimum. In addition, our performance model mirrors a fairly basic Megatron-Core training stack, and the absence of a clear monotone trend in the optimal expert split factor $G$ may itself be an artifact of that stack's systems penalty. Thus, broader kernel-level changes could turn the optimal $G$ into a predictable function of the budget, which we consider both a limitation and grounds for future expansive validations. Finally, our runs limit the expert split factor to $G \le 8$, and future work could study extrapolations of the law along that dimension. Our studies are also carefully designed to bound the tokens-per-parameter ratio, so we do not study the overtrained regime where models train far beyond these ratios, and we leave that as future work. For future work, we layout the following directions that we believe are worth pursuing.
\begin{itemize}
    \item \textbf{Joint batch-size and parallelism optimization.}
    In this work, we fixed the global batch size across all our runs. Extending the decision space to include it would allow MOSAIC to jointly reason about convergence, strong scaling, and the layout changes induced by the number of microbatches in flight.

    \item \textbf{Broader and accelerator-aligned geometry families.}
    Our current scaling law sweep and MOSAIC search share a single backbone
    geometry ladder. Extending the fit across multiple depth, width and $d_{ff}$
    co-scaling families would test whether the $(N_{\mathrm{tot}},S,G)$
    representation transfers across backbone geometries and would allow MOSAIC
    to search beyond the family used to fit the law. Incorporating
    hardware-friendly tile and wave constraints into these families would further
    enable accelerator-specific architecture selection.

    \item \textbf{Communication and computation overlap.}
    Our current performance model costs compute and exposed communication largely additively, overlapping only the data-parallel collectives where scheduling windows allow. Extending the performance model to represent persistent and fused kernels that overlap collectives with computation would improve MFU prediction as modern training stacks increasingly depart from additive compute-and-communication execution.
\end{itemize}
\acks{We thank Valentyn Boreiko, Moritz Haas, Srikanth
Doss, Volkan Cevher, Cole Hawkins, Leonard Lausen, Anna Ruminshky, Hongshan Li and Subendhu Rongali for their help in setting up the
infrastructure and the training pipeline used to train the models in this paper. We also thank Justin Chiu for the initial version of the analytical model implementation used in this paper.}

\bibliography{example_paper}

\appendix
\section{Pre-training Recipe}
\label{app:training}
In this section, we cover some more details of the pretraining recipe and the architecture that went into the training runs.

\subsection{Architecture and Pretraining Details}
\label{sec:train_recipe}

\paragraph{Sparse MoE Architecture.} Our MoE block follows the DeepSeekMoE
design~\citep{dai2024deepseekmoe}, which splits the reference FFN into
fine-grained routed experts and adds a shared expert that every token passes
through, paired with the auxiliary-loss-free bias balancing of
\citet{wang2024lossfree} in the router. In our instantiation, the shared expert
has the same hidden width
$d_{\text{expert}}$ as a routed expert, so as the expert split factor $G$ increases, $d_\text{expert}$ decreases and so does the shared expert size. The attention mechanism is gated~\citep{qiu2025gated} and a
head-specific sigmoid gate is applied to the output of each
scaled-dot-product attention head before the output projection, which
the authors find consistently improves loss for both dense and MoE
models at this scale. We use pre-LayerNorm and SwiGLU activations
inside each expert and the embedding and unembedding matrices are not tied. We do not use QK layernorm for the attention layer.

\paragraph{Pretraining objective and data.} Every run is a pure
text-to-text pretraining run with the next-token-prediction objective of
GPT-2~\citep{radford2019language}. We do not use any auxiliary
denoising or instruction-tuning loss aside the sequence auxiliary loss in the MoE routing layer which we discuss in later sections. The training corpus is the
English portion of mC4~\citep{xue2020mt5}, the multilingual extension
of the Colossal Clean Crawled Corpus (C4) that is itself a cleaned
subset of Common Crawl. We tokenize with the GPT-2 BPE tokenizer, which has a vocabulary of 50,257 tokens.

\paragraph{Numerical precision.} All training runs use bf16 weights and
activations with a fp32 master copy of the optimizer state.

\paragraph{Initialization.} We use the Megatron/GPT-2 architecture in our runs
\citep{radford2019language}. Fan-in (input-side) projections, the QKV
and first FFN matrices, the router $W_r$, and the embedding and readout
matrices, are drawn flat from $\mathcal{N}(0, 0.006^{2})$. Output-side
residual projections, the attention output projection and the
expert/FFN down-projection, are depth-scaled to
$\mathcal{N}\!\bigl(0, (0.006/\sqrt{2L_{\text{layers}}})^{2}\bigr)$, so deeper models
initialize the residual-write weights smaller and keep the
residual-stream variance stable across the $2L_{\text{layers}}$ residual additions per
forward pass. For the normalization layers, their
gains are set to $1$ and biases to $0$. Initializing $W_r$ like any
other fan-in matrix leaves router logits approximately uniform across
experts at step zero. We leave $\mu$P-style initialization for
hyperparameter transfer~\citep{jiang2026hpmoe} in sparse MoE models as future
work.
\begin{figure}[h]
\centering
\begin{subfigure}[t]{0.49\textwidth}
  \centering
  \includegraphics[width=\linewidth]{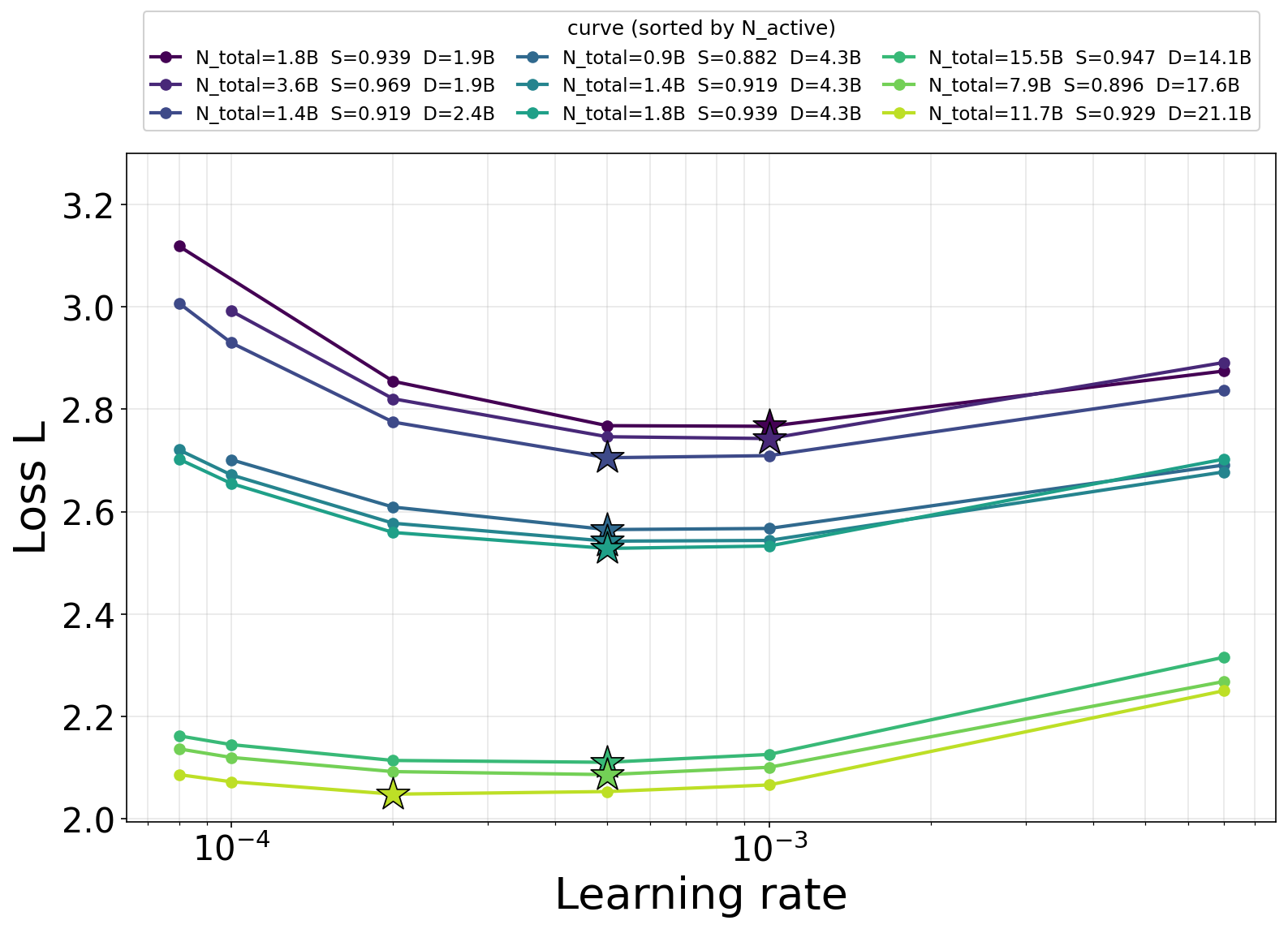}
  \subcaption{Loss vs.\ peak learning rate per configuration, with the
  per-configuration optimum (stars) read from a quadratic fit in
  $\log_{10}\mathrm{lr}$.}
  \label{fig:loss-vs-lr}
\end{subfigure}\hfill
\begin{subfigure}[t]{0.49\textwidth}
  \centering
  \includegraphics[width=\linewidth]{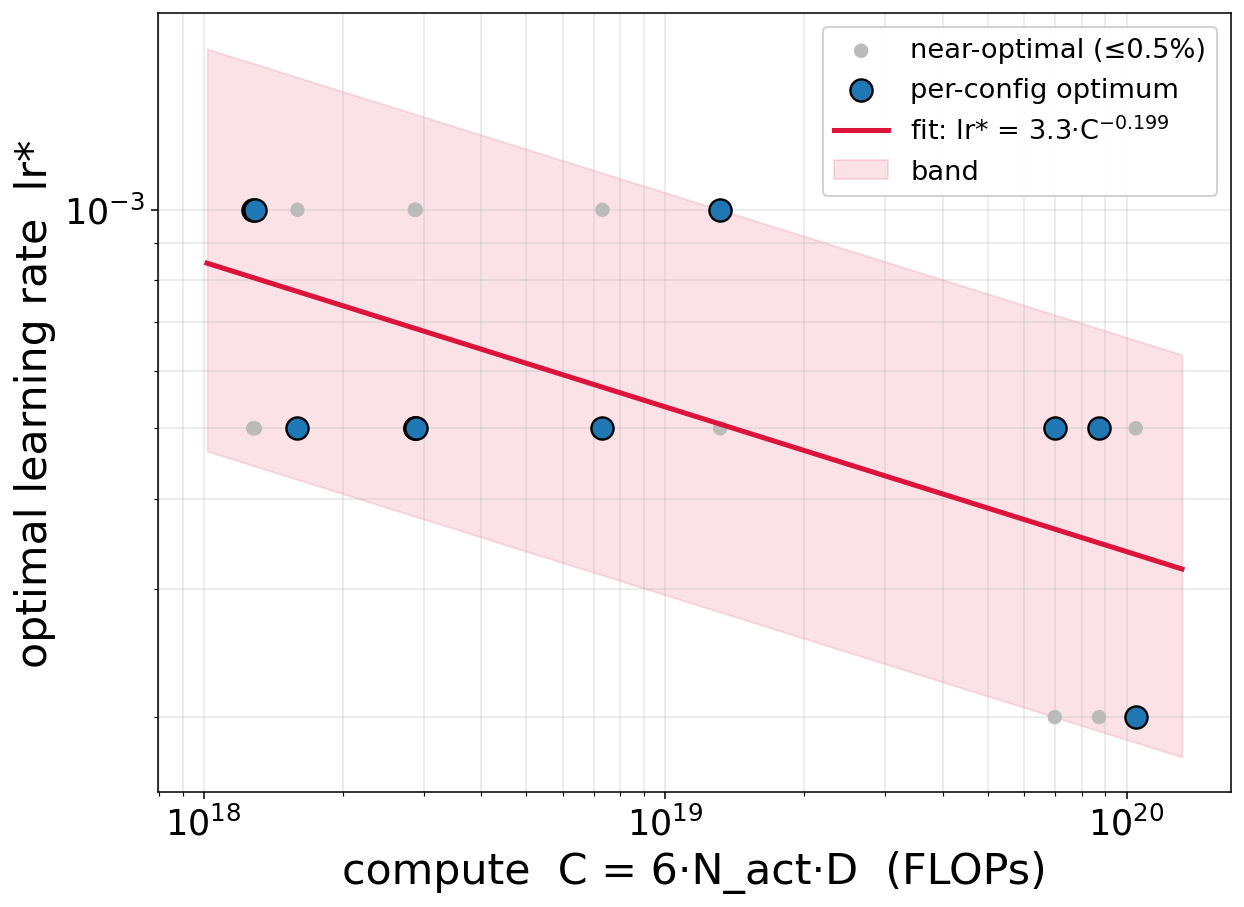}
  \subcaption{Optimal learning rate vs.\ model FLOPs
  $C_{\text{model}} = 6\,N_{\text{act}}\,D$, solid line is the fit
  $\mathrm{lr}^{\star}(C_{\text{model}}) = 3.30\,C_{\text{model}}^{-0.199}$, shaded region is the
  near-optimal band.}
  \label{fig:lr-compute}
\end{subfigure}
\caption{Setting the peak learning rate. Left: loss-vs-learning-rate
curves across MoE configurations, each with a quadratic-fit optimum.
Right: those optima collapse onto a single power law in the model
FLOPs $C_{\text{model}}$, with one law predicting the optimal learning rate
across the full range of total parameters and sparsity.}
\label{fig:lr}
\end{figure}
\subsection{Optimizer and Learning-rate Settings}
\label{app:lr_hpo}

We use AdamW at weight decay $0.1$ with the
warmup-stable-decay (WSD) schedule of~\citet{hagele2024scaling}, namely a
fixed warmup of approximately $100$-million tokens regardless of run
length, a stable phase at peak learning rate, and a linear decay to
zero over the final $20\%$ training steps. The AdamW moments are the standard defaults
$\beta_1 = 0.9$ and $\beta_2 = 0.999$ with $\varepsilon = 10^{-14}$, and we do
not clip gradients. We do not conduct an independent learning-rate sweep for every scaling law run. Instead, each run’s peak learning rate is assigned from a single compute-conditioned rule fitted on a separate pilot sweep. To keep learning rate from confounding the scaling law runs, we set the
peak learning rate from a single power law in the model FLOPs
$C_{\text{model}} = 6\,N_{\text{act}}\,D$. We sweep a small grid of peak
learning rates across MoE configurations spanning a wide range of total
parameters, sparsity, and token budgets. Each configuration traces a
loss-vs-learning-rate curve whose optimum we read from a quadratic fit
in $\log_{10}\mathrm{lr}$ (Figure~\ref{fig:lr}, left). For each
configuration we keep the near-optimal band, every swept rate whose loss
lands within a small tolerance of that configuration's best, following
the band-fit practice of prior learning-rate
studies~\citep{tian2026towards}. Fitting
$\log_{10}\mathrm{lr}^{\star}$ against $\log_{10} C_{\text{model}}$ over these
near-optimal points gives
\begin{equation}
\mathrm{lr}^{\star}(C_{\text{model}}) \;=\; 3.30\;C_{\text{model}}^{-0.199},
\qquad C_{\text{model}} = 6\,N_{\text{act}}\,D \,
\label{eq:lr-compute}
\end{equation}
with a tight parallel band around the central line
(Figure~\ref{fig:lr-compute}) where any rate inside the band is well tuned,
and the band width reflects the resolution of the swept grid.

We use compute alone as the axis for learning rate predictions and we leave more appropriate sparsity specific learning rate predictions (if there are any) for future work. AWithin our pilot sweep, after conditioning on $C_{\mathrm{model}}$, we detect
no systematic residual dependence of the near-optimal learning rate on
$N_{\mathrm{tot}}$, sparsity, or the $N_{\mathrm{act}}/D$ split. We therefore use the compute-conditioned rule rather than fitting a higher-dimensional learning-rate surface, which would require substantially
more dedicated hyperparameter sweeps and could itself confound the scaling law fit.  The decay exponent also agrees with independent
MoE learning-rate fits, where \citet{tian2026towards} report a slope of
$-0.153$ over a comparable compute range, close to our value. In
practice, this gives a single rule, where for any candidate configuration
we compute $C_{\text{model}} = 6\,N_{\text{act}}\,D$ and read the peak learning rate off
Equation~\eqref{eq:lr-compute}.

\paragraph{Out-of-scope HPO axes.} Optimizer hyper-parameters
(weight decay $0.1$, warmup, scheduler shape), sequence length
($T_{\text{seq}} = 2048$), drop-token capacity factor (default infinite capacity), are
each held to a single value across all runs. The law is therefore
conditional on this recipe. 

\subsection{MoE Routing Details}
\label{app:routing-details}

In this section, we describe the choices behind the routing algorithm for our scaling recipe presented in Algorithm~\ref{alg:routing}.

\paragraph{Sigmoid scoring with per-expert bias.}
Following \citet{wang2024lossfree}, we instantiate the score vector
$\boldsymbol s(x)$ of Equation~\eqref{eq:moe-output} as a clean sigmoid score $s'_i(x)$ plus an
additive per-expert bias $\phi \in \mathbb{R}^{E}$, so that
\begin{equation}
s_i(x) \;=\; s'_i(x) + \phi_i \;=\; \sigma\!\bigl(\theta_i(x)\bigr) + \phi_i,
\qquad
\theta_i(x) \;=\; \tilde{w}_{r,i}^{\!\top}\, x
\label{eq:sigmoid-score}
\end{equation}
where $\tilde{w}_{r,i}$ denotes the row-normalized
$i$-th row of $W_r$. The sigmoid form decouples per-expert logits, avoiding
the shared-softmax competition that drives unbounded logit growth and forces
ad-hoc remedies such as router $z$-loss or logit capping. The bias $\phi$ is not a
learned parameter and it is updated by a controller using
only routing statistics, and it influences which experts are
selected.

\paragraph{LFLB bias controller.}
Following Algorithm~\ref{alg:routing}, let $n_i$ denote the number of tokens routed to expert $i$ in a batch and let
$\bar n = \frac{1}{E}\sum_{i=1}^{E} n_i$. Following \citet{wang2024lossfree, team2025kimi}, we
update $\phi$ in the direction of imbalance, but adopt an RMS-normalized
violation rather than a raw sign so that the magnitude of the correction is
proportional to the shape of the imbalance pattern and not the absolute
batch size, as follows
\begin{equation}
u_i \;=\; \frac{\bar n - n_i}{\sqrt{E^{-1}\sum_j (\bar n - n_j)^2 + \varepsilon}},
\qquad
\phi_i \;\leftarrow\; \phi_i + \mu_b\, u_i
\label{eq:bias-update}
\end{equation}
We additionally
center the bias after each update,
$\phi_i \leftarrow \phi_i - \frac{1}{E}\sum_j \phi_j$, since only relative bias differences
affect routing.

\paragraph{Probabilistic top-$K$ via Gumbel proposal--correction.}
We replace deterministic selection with a Gumbel-noise
proposal--correction sampler operating at a fixed temperature
$\tau$ :
\begin{align}
\text{(Gumbel proposal)}\quad
\mathcal{C}_t &= \operatorname{Top}\!M_i\bigl(\log s'_{t,i} + \tau\, g_{t,i}\bigr),
& g_{t,i} &\sim \mathrm{Gumbel}(0,1), \label{eq:gumbel-prop}\\
\text{(Gumbel correction)}\quad
\mathcal{T}_{K,t} &= \operatorname{Top}\!K_{i \in \mathcal{C}_t}\bigl(s_{t,i}\bigr),
& M &= \min(\rho K, E). \label{eq:gumbel-corr}
\end{align}
The construction rests on the Gumbel-max theorem. For $K\!=\!1$ and
$g_i \sim \mathrm{Gumbel}(0,1)$,
\begin{equation}
\operatorname*{arg\,max}_i\bigl(\log p_i + \tau\, g_i\bigr)
\;\sim\;
\mathrm{Categorical}\!\left(\tfrac{p_i^{1/\tau}}{\sum_j p_j^{1/\tau}}\right)
\label{eq:gumbel-max}
\end{equation}
and its top-$K$ extension gives exact sampling without
replacement from the same distribution~\citep{yellott1977relationship,kool2019stochastic}.
Injecting the noise in log-score space, as in Equation~\eqref{eq:gumbel-prop},
therefore draws the proposal set exactly from $P(i) \propto s'^{1/\tau}_{t,i}$
on the clean scores, and the bias $\phi$ is then applied in the correction
stage of Equation~\eqref{eq:gumbel-corr} to preserve those sampling semantics.
This tempered sampling interpretation holds for the proposal before centering
and correction, and the implemented algorithm modifies that distribution. The
per-expert centering of
Algorithm~\ref{alg:routing} makes the $g_{t,i}$ dependent across the batch and the
bias-aware correction reweights the survivors, so we treat both as heuristic
variance-control and load-balancing steps rather than as sampling from the
tempered clean-score distribution. The proposal size is $M = \min(\rho K, E)$
with $K \le E$, since $\rho K$ exceeds $E$ at the small expert counts in our
grid. The temperature $\tau$ sets how peaked the proposal distribution is, and
as $\tau\!\to\!0$ the algorithm approaches a deterministic clean-score top-$M$
proposal followed by the bias-aware top-$K$ correction within that proposal.
This limit need not coincide with the global deterministic top-$K$ on
$s_{t,i}$, since the bias can favor an expert outside the clean top-$M$. For
the uncentered, uncorrected Gumbel-max baseline with two experts and $K\!=\!1$,
the probability of selecting $b$ over $a$ is
$\sigma\!\bigl(\tau^{-1}\log(s'_{b}/s'_{a})\bigr)$, so near-tied experts
receive meaningful exploration while confident decisions are exponentially
protected. In Algorithm~\ref{alg:routing} itself, at two experts $\rho = 12$
puts both in the proposal and the correction then selects deterministically
under $s_{t,i}$. 

As a final, empirically important step, we
center the noise per expert across the current batch,
$\tilde g_{t,i} = g_{t,i} - \frac{1}{T}\sum_t g_{t,i}$, so that finite-sample
noise means do not waste the bias controller's correction budget chasing
purely stochastic imbalances. We set
$\tau = 0.01$ and $\rho = 12$, with token-level granularity across all our runs, and
we use $\mu_b = 10^{-3}$ and $\varepsilon = 10^{-4}$ in the bias update of
Equation~\eqref{eq:bias-update}. We find that this method of sampling with low $\tau$ allows us to overcome issues of load balancing and experts starved of tokens in a single batch later in long horizon runs especially at large active parameters of the scale of 18 billion parameters $N_\text{act}$ and more. Compared to adding more dense layers at the beginning as done in \citet{singh2026arcee} which reduces the model capacity at matched $N_{\text{act}}$, we find that our Gumbel sampling method along with the proposed routing algorithm helps in more efficient mitigation of instabilities due to spikes in starved experts.

\paragraph{Router-weight row normalization.}
A subtle source of routing pathology is heterogeneity in the row norms of
$W_r$. Experts whose corresponding rows have grown a larger norm are favored
for reasons unrelated to specialization. We optionally row-normalize the
router weights at every forward pass $\tilde w_{r,i} \;=\; c_r \cdot \frac{w_{r,i}}{\|w_{r,i}\|_2}$, where the target norm $c_r$ is a fixed constant, in our case set to 1.0. It allows to recover routing decisions that depend on directional alignment
between $w_{r,i}$ and $x$ rather than on accidental norm drift. This is
the $\textsc{RowNorm}(W_r, c_r)$ step in Algorithm~\ref{alg:routing}, which
produces the normalized router $\tilde W_r$ used to compute the
pre-sigmoid logits $\Theta$.

Additionally, the bias update of Eq.~\eqref{eq:bias-update} is invoked once per optimizer step on
the per-expert counts $n$ returned by the algorithm. We emphasize that load
balancing is carried by the bias controller and
the noise distribution, where no batch-level load-balancing loss term is
added to the optimization objective, in contrast to the auxiliary
LBL of \citet{shazeer2017outrageously,fedus2022switch}. The one auxiliary term we
retain is the sequence-level loss described below.

\paragraph{Router regularization.} In addition to the LFLB bias
controller of Section~\ref{sec:routing}, we apply the router $z$-loss
of~\citet{zoph2022st} to suppress unbounded growth in the
pre-sigmoid router logits $\Theta_{t,i} = \theta_i(x_t)$ of
Eq.~\eqref{eq:sigmoid-score}:
\begin{equation}
\mathcal{L}_{z}
\;=\;
\frac{1}{T}\sum_{t=1}^{T}
\Bigl(\log \sum_{i=1}^{E} e^{\Theta_{t,i}}\Bigr)^{2}
\label{eq:zloss}
\end{equation}
added to the cross-entropy loss with a small fixed coefficient. Because
selection in our routing variant is taken on $\sigma(\Theta) + \phi$ rather
than on $\Theta$ directly, $\mathcal{L}_{z}$ acts purely as a numerical
guard on the logits and does not interfere with the bias controller. We set the coefficient of $z$-loss to 1e-4 in all our runs.

\paragraph{Sequence Auxiliary Loss.} Unlike the load balancing batch auxiliary loss used in the softmax routing variants, we follow the sequence auxiliary loss as used in \cite{deepseekai2024v3}. We set its coefficient to 1e-3 in all our runs.

\section{Scaling Analysis}
\label{app:scaling-law-coeffs}
\begin{figure}[t!]
\centering
\includegraphics[width=\linewidth]{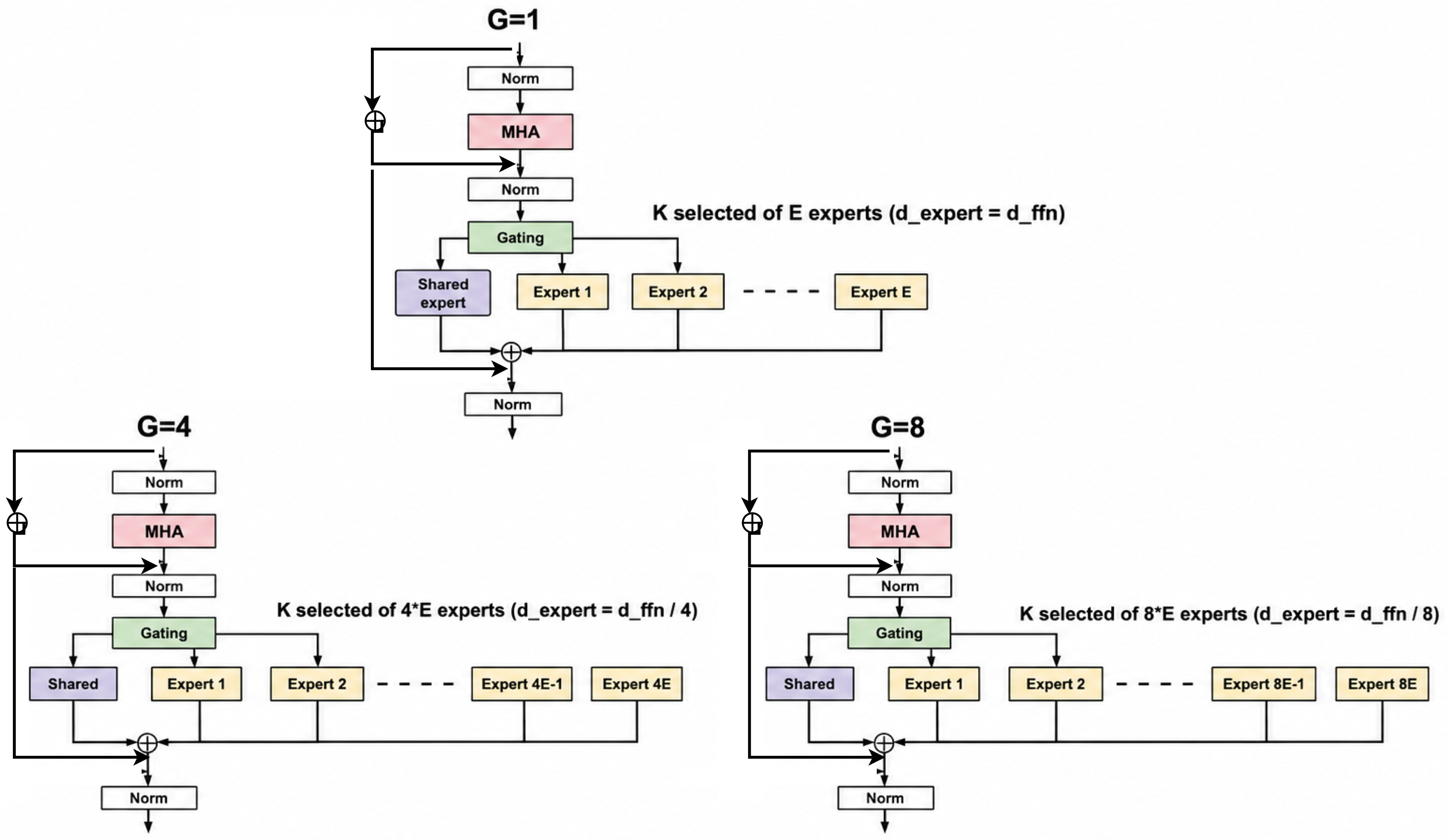}
\caption{Example of an MoE layer in a transformer block, at three expert split factors
  $G\!\in\!\{1,4,8\}$. In every panel, the routed experts sit beside a
  shared expert. $d_{\text{ff}}$ is the hidden width of a single reference
  dense feed-forward expert ($G\!=\!1$), and the total routed model width $E\,d_{\text{ff}}$ is \textbf{held fixed} across the
  panels. Raising $G$ splits each reference FFN into $G$ narrower experts,
  so the routed-expert count grows ($E\!\to\!G\,E$) while each expert's
  hidden width shrinks proportionally ($d_{\text{expert}} = d_{\text{ff}}/G$). Relative to $G\!=\!1$, a factor-$G$ configuration has $G$ times as many
  experts, each $G$ times narrower. The top-$K$ selection is kept the same
  across the three panels, so the active parameters $N_{\text{act}}$ shrink
  with higher $G$ rather than staying fixed. Note that this is different from granularity defined in \cite{krajewski2024scaling}.}
\label{fig:g-geometry}
\end{figure}

This appendix collects the per-$G$ evidence behind the scaling laws of
Section~\ref{sec:scaling-law-design}, starting with
Figure~\ref{fig:g-geometry}, which shows how the expert split factor $G$ of
Section~\ref{sec:preliminaries} reshapes an MoE layer.

\begin{figure}[t!]
\centering
\begin{subfigure}[t]{0.30\textwidth}
  \centering
  \includegraphics[width=\linewidth]{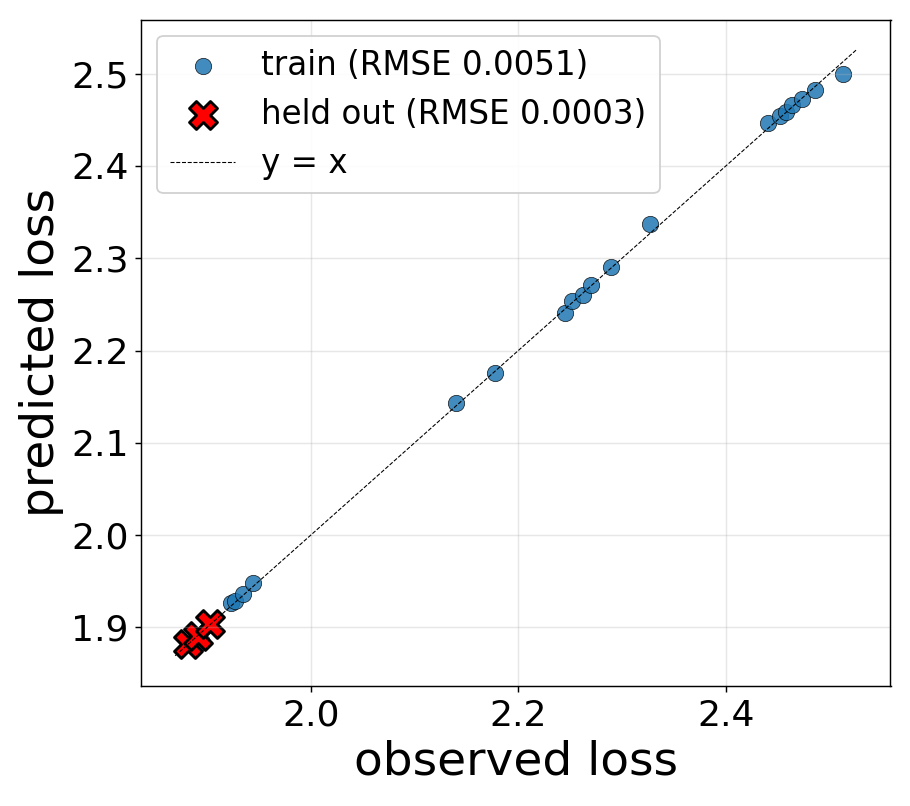}
  \subcaption{$G=1$.}
  \label{fig:parity-G1-app}
\end{subfigure}\hspace{0.03\textwidth}
\begin{subfigure}[t]{0.30\textwidth}
  \centering
  \includegraphics[width=\linewidth]{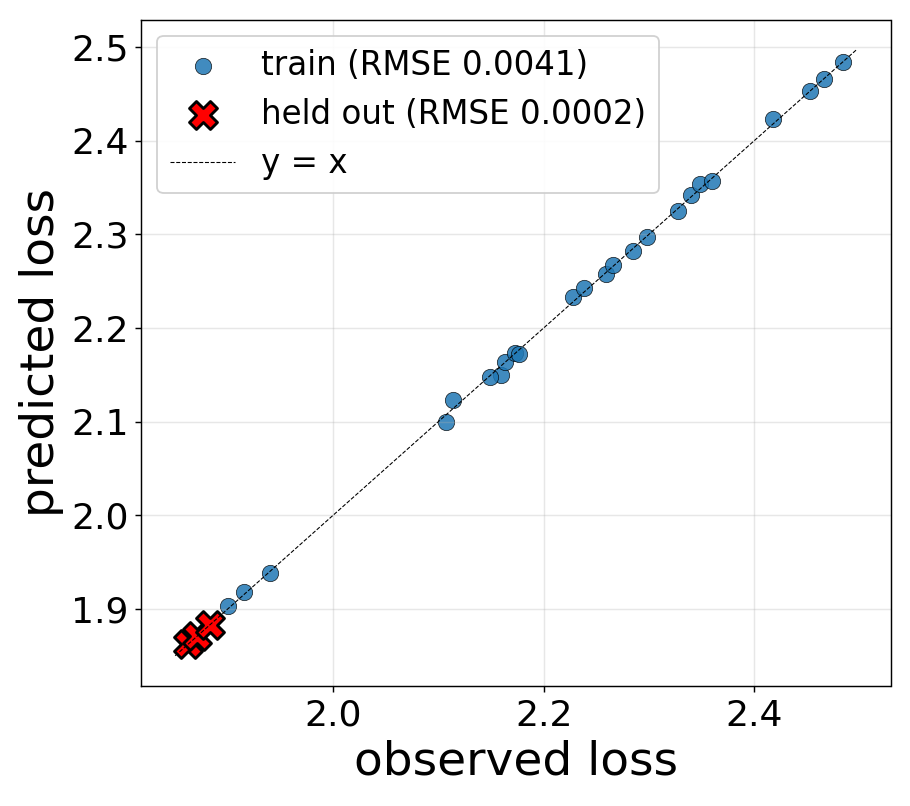}
  \subcaption{$G=2$.}
  \label{fig:parity-G2-app}
\end{subfigure}

\vspace{0.6em}

\begin{subfigure}[t]{0.30\textwidth}
  \centering
  \includegraphics[width=\linewidth]{figs/parity_G4_main.png}
  \subcaption{$G=4$.}
  \label{fig:parity-G4-app}
\end{subfigure}\hspace{0.03\textwidth}
\begin{subfigure}[t]{0.30\textwidth}
  \centering
  \includegraphics[width=\linewidth]{figs/parity_G8.png}
  \subcaption{$G=8$.}
  \label{fig:parity-G8-app}
\end{subfigure}
\caption{Predicted-vs-observed loss for the
$\mathcal{L}(N_{\text{tot}}, S, D)$ fit with Equation~\eqref{eq:abnar-law},
refit independently on each expert split factor stratum
$G\!\in\!\{1, 2, 4, 8\}$. The $G\!=\!4$ and $G\!=\!8$ panels are the
ones reproduced in Figure~\ref{fig:per-G-parity}.}
\label{fig:per-G-parity-appendix}
\end{figure}

\subsection{Per-$G$ Scaling Law Fits}
\label{sec:sl_sparsity}

This appendix section gives the per-$G$ details behind
Section~\ref{sec:law-with-sparsity}. We fit $\mathcal{L}(N_{\text{tot}}, S, D)$ in Equation~\eqref{eq:abnar-law} on each $G$ separately, holding out the
largest $10\%$ by active parameters. The fit is noise-limited in all the per $G$ optimization of the functional form and the
held-out large models match it, with no residual structure on $E$, $K$,
$N_{\text{tot}}$, $S$, $D$, or $L_{\text{layers}}$ (all Pearson correlations below
$0.13$ in magnitude).
Refitting the same form independently per expert split factor
as shown in Figure~\ref{fig:per-G-parity-appendix}, ($G\!\in\!\{1, 2, 4, 8\}$) fits
each $G$ well in isolation, but the fitted coefficients drift
substantially across $G$, so no single set of $(N_{\text{tot}}, S, D)$
coefficients describes all expert split factors at once. This is the
signal that $G$ must enter the law itself, which we do in
Section~\ref{sec:joint-NtSDG-law}.



\subsection{Joint-Law Uncertainty and Identifiability}
\label{sec:sl_joint_uncertainty}

This appendix section gives the evidence behind the $G$-augmented joint law
of Section~\ref{sec:joint-NtSDG-law}, namely that a larger expert split
factor $G$ lowers loss at a fixed parameter and data budget. When we fit the
joint law, this preference shows up as the split-factor exponent
$\eta\!\approx\!+0.95$, and the question is whether that value is a real
effect or an artifact of the particular runs we happened to collect.
To quantify uncertainty in $\eta$, we perform $500$ warm-started
wild-bootstrap refits using Rademacher residual perturbations and report a
$95\%$ BCa interval. The resulting interval is $[+0.91, +0.95]$, which
remains well above zero.


\paragraph{Which coefficients are identified.} The bootstrap above is
warm-started at the fitted point, so it describes only the local basin
around it. A separate scan of $400$ fresh restarts shows the loss surface
has a wide, near-flat identifiability ridge along which the prefactors and
the remaining exponents move at near-constant objective. Only $\eta$ and
$\beta$ are separately identified, the former with the interval reported
above and the latter with a BCa interval of $[+0.206, +0.208]$. The other
coefficients are best read as coordinates along this ridge, where their
tight local intervals in Table~\ref{tab:joint-law-coeffs} understate the
true cross-basin spread, so downstream sensitivity studies should perturb
them jointly along the ridge rather than one at a time.

\paragraph{Identifiability caveats.} Because the fit sits close to the
run-to-run scatter, the loss surface is flat along a ridge in coefficient space. The
capacity prefactor and exponent $(a, \alpha)$, the joint-term
parameters $(j, \gamma, \delta)$, and the stand-alone sparsity floor
$(c, \lambda)$ trade off against one another at near-constant objective.
This is why we report only $\eta$ and $\beta$ as identified and treat
the rest as ridge coordinates: the law's loss value and its per-$G$
calibration are robust, but no single split of the prefactors and
exponents is privileged, and pinning one for publication requires
fixing a prefactor (for example $a$) and refitting. The
stand-alone sparsity-floor term $c/(1-S)^{\lambda}$ is also numerically
small at our $S\!\leq\!0.981$ because most of the sparsity penalty is
carried by the joint
$j/((1-S)^{\delta}\,N_{\text{tot}}^{\gamma}\,G^{\eta})$ term. Finally,
the asymptotic blow-up $\mathcal{L}\!\to\!\infty$ as $S\!\to\!1$ is
imposed by the form rather than measured. Pinning $(c, \lambda, \delta)$
empirically would require runs at $S$ near $0.999$, which our grid does
not contain.

%
%
%
%
%
%
%
\section{Scaling Sweep Training Runs}
\label{app:dataset}

\begin{table}[!t]
\centering
\small
\setlength{\tabcolsep}{5pt}
\begin{tabular}{@{}lrrr@{}}
\toprule
Quantity & Min & Median & Max \\
\midrule
Total parameters $N_{\text{tot}}$ (B)        & 0.22  & 5.34  & 79.24 \\
Active parameters $N_{\text{act}}$ (B)       & 0.104 & 0.339 & 2.72  \\
Sparsity $S$                                 & 0.499 & 0.944 & 0.981 \\
Expert split factor $G$                      & 1     & -     & 8     \\
Expert count $E$                             & 8     & -   & 512   \\
Top-$K$ activation                           & 1     & -     & 8     \\
$L_{\text{layers}}$                                         & 8     & -    & 24    \\
Residual width $d$                           & 1024  & -  & 3072  \\
Training tokens $D$ (B)                      & 2.88  & 5.76  & 56.38 \\
$\mathrm{TPP} = D/N_{\text{act}}$            & 10.8  & -  & 70  \\
Compute (non-embedding FLOPs)                & $1.9\times10^{18}$ & - & $8.1\times10^{20}$ \\
\bottomrule
\end{tabular}
\caption{Coverage of the training run dataset. These statistics cover only the
runs used to fit and validate the scaling law. The staged MOSAIC validation
runs of Section~\ref{sec:results-staged}, trained for over 200B tokens each,
are not included.}
\label{tab:dataset-headline}
\end{table}

\begin{table}[h]
\centering
\small
\setlength{\tabcolsep}{4pt}
\begin{tabular}{@{}rrrrrll@{}}
\toprule
$G$ & $N_{\text{tot}}$ (B) & $N_{\text{act}}$ (B) & $S$ & $D$ (B) & $L_{\text{layers}}$ & $E$, $K$ \\
\midrule
1 & 0.54\textendash30.91 & 0.152\textendash1.179 & 0.72\textendash0.98 & 3.84\textendash28.19 & \{8,12,16\} & $E\!\in\!\{8\text{\textendash}128\}$, $K\!\in\!\{1,2\}$ \\
2 & 0.95\textendash12.65 & 0.124\textendash0.387 & 0.87\textendash0.97 & 3.84\textendash5.76  & \{8,12,16\}    & $E\!\in\!\{32\text{\textendash}128\}$, $K\!\in\!\{1,2\}$ \\
4 & 0.22\textendash79.24 & 0.110\textendash2.723 & 0.50\textendash0.98 & 2.88\textendash56.38 & \{8,12,16,24\} & $E\!\in\!\{12\text{\textendash}384\}$, $K\!\in\!\{2, 4,6,8\}$ \\
8 & 0.71\textendash53.0 & 0.104\textendash2.52 & 0.85\textendash0.98 & 3.84\textendash50.00 & \{8,12,16,24\} & $E\!\in\!\{96\text{\textendash}512\}$, $K\!\in\!\{2,4,8\}$ \\
\bottomrule
\end{tabular}
\caption{Ranges of total and active parameters, sparsity, training tokens,
depth, expert count, and top-$K$ spanned by the scaling runs at each expert
split factor $G$, where B denotes billions.}
\label{tab:dataset-perG}
\end{table}

\subsection{Coverage of the Scaling-Law Dataset}

This appendix documents the slice of the
$(N_{\text{tot}}, N_{\text{act}}, S, G, D)$
space that the joint law of Section~\ref{sec:joint-NtSDG-law} is fit on.
The dataset comprises $n_{\text{runs}} = 150$ from-scratch sparse MoE pretraining runs. Our runs span
$N_{\text{tot}} \in [0.22, 79.24]\,$B parameters,
$N_{\text{act}} \in [0.10, 2.72]\,$B parameters,
$S \in [0.499, 0.981]$, $G \in \{1, 2, 4, 6, 8\}$,
$E \in \{8, 12, 16, 24, 32, 48, 64, 96, 128, 192, 256, 384, 512\}$, $K \in \{1, 2, 4, 6, 8\}$, training tokens
$D \in [2.88, 56.38]\,$B
($\mathrm{TPP} = D/N_{\text{act}} \in [10.8, 70]$),
across four architecture families \\
$(L_{\text{layers}}, d) \in \{(8,1024), (12,1536), (16,2048), (24,3072)\}$. Total
compute for our scaling law fits spans $1.9\times10^{18}$\textendash$8.1\times10^{20}$
FLOPs. We display some of the statistics of our training runs in Table~\ref{tab:dataset-headline}.

\begin{table}[h]
\centering
\small
\setlength{\tabcolsep}{6pt}
\begin{tabular}{@{}rrrrrl@{}}
\toprule
$L_{\text{layers}}$ & $d$ & $n_{\text{runs}}$ & $D$ (B) & $N_{\text{tot}}$ (B) & $G$ values \\
\midrule
 8 & 1024 & 52 & 2.88\textendash7.68   & 0.22\textendash7.14   & $\{1,2,4,6,8\}$ \\
12 & 1536 & 56 & 5.76\textendash11.52  & 0.73\textendash25.20  & $\{1,2,4,6,8\}$ \\
16 & 2048 & 39 & 28.19\textendash56.38 & 4.13\textendash45.66  & $\{1,2,4,8\}$ \\
24 & 3072 &  4 & 50.00                 & 40.16\textendash79.24 & $\{4,8\}$ \\
\bottomrule
\end{tabular}
\caption{Per-family coverage. $D$ is chained to $(L_{\text{layers}}, d)$ by an outer
Chinchilla-style budgeting decision.}
\label{tab:dataset-perfamily}
\end{table}

\subsection{Training Dynamics of the MoE Models}
\label{app:train_dynamics}

\begin{figure}[p]
\centering
\begin{subfigure}[t]{0.44\textwidth}
  \centering
  \includegraphics[height=0.42\textheight]{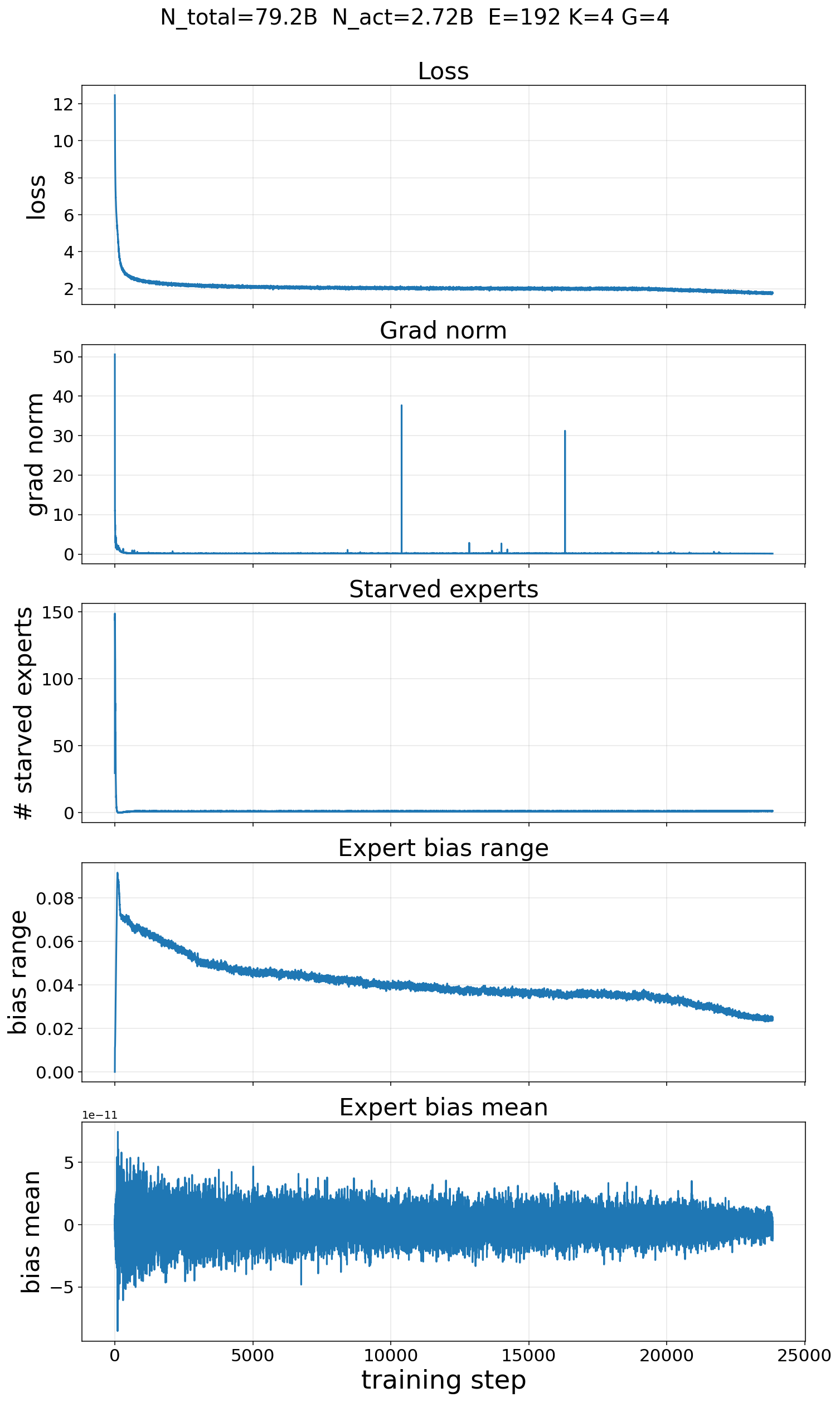}
\end{subfigure}\hfill
\begin{subfigure}[t]{0.44\textwidth}
  \centering
  \includegraphics[height=0.42\textheight]{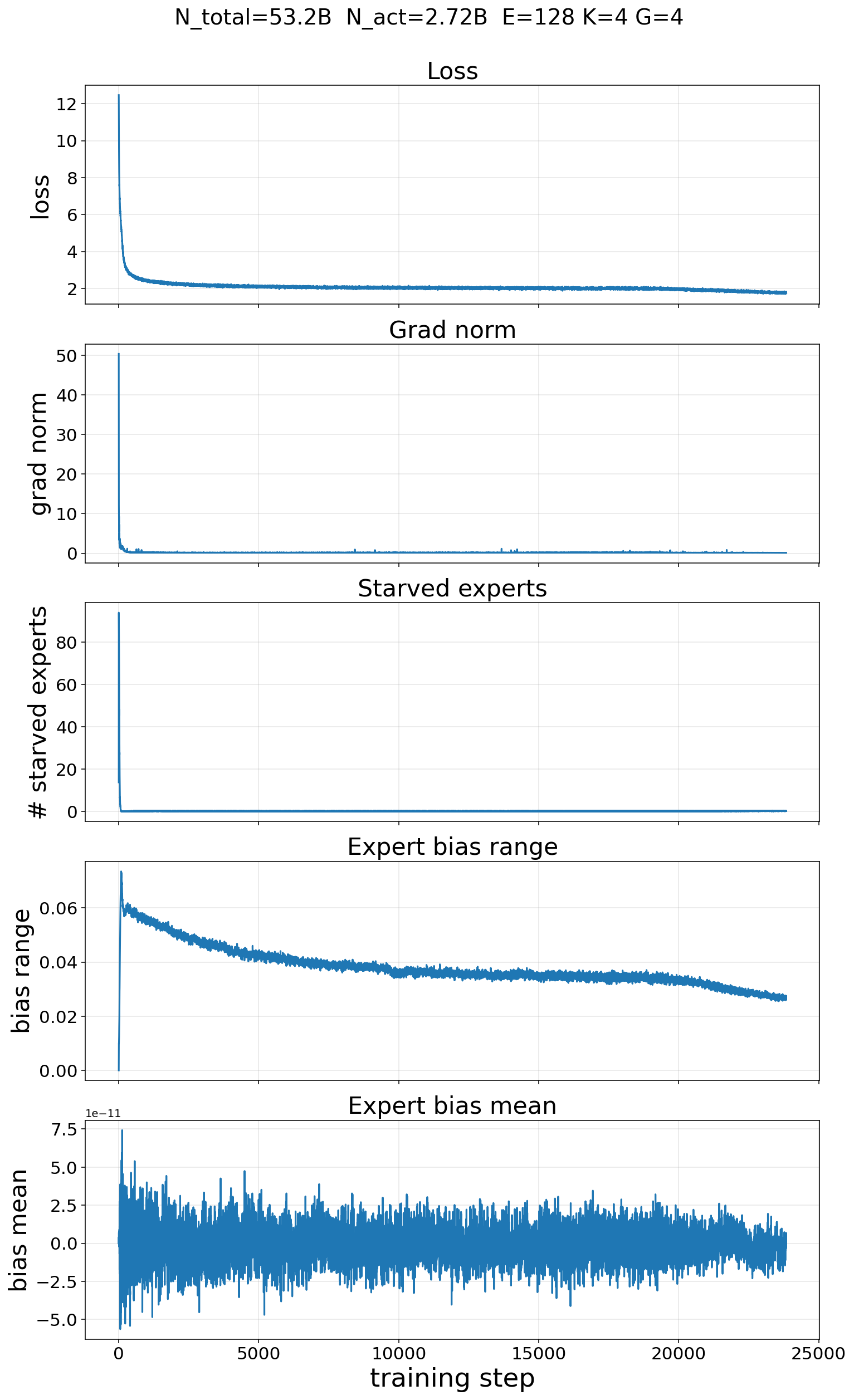}
\end{subfigure}

\vspace{0.6em}

\begin{subfigure}[t]{0.44\textwidth}
  \centering
  \includegraphics[height=0.42\textheight]{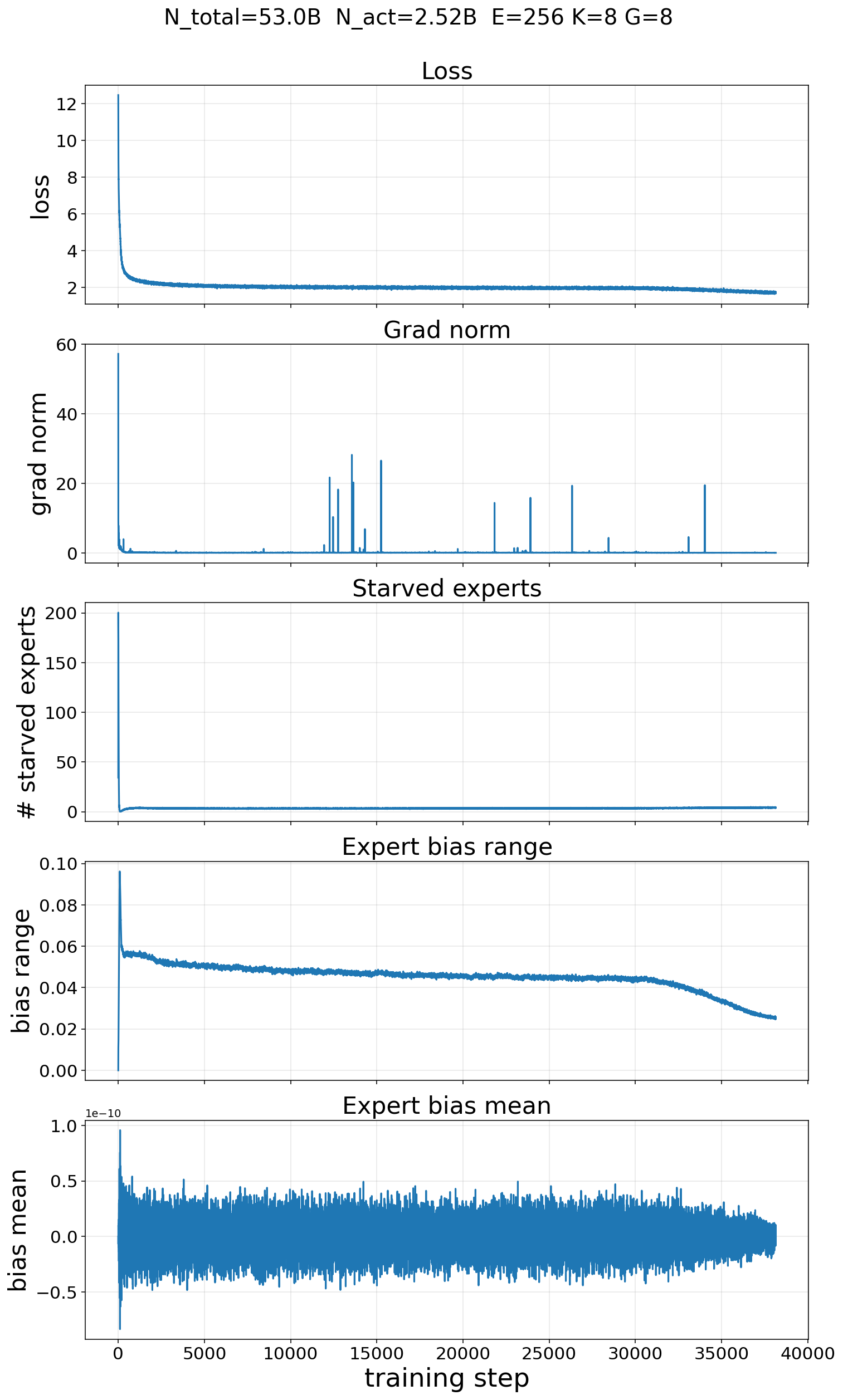}
\end{subfigure}\hfill
\begin{subfigure}[t]{0.44\textwidth}
  \centering
  \includegraphics[height=0.42\textheight]{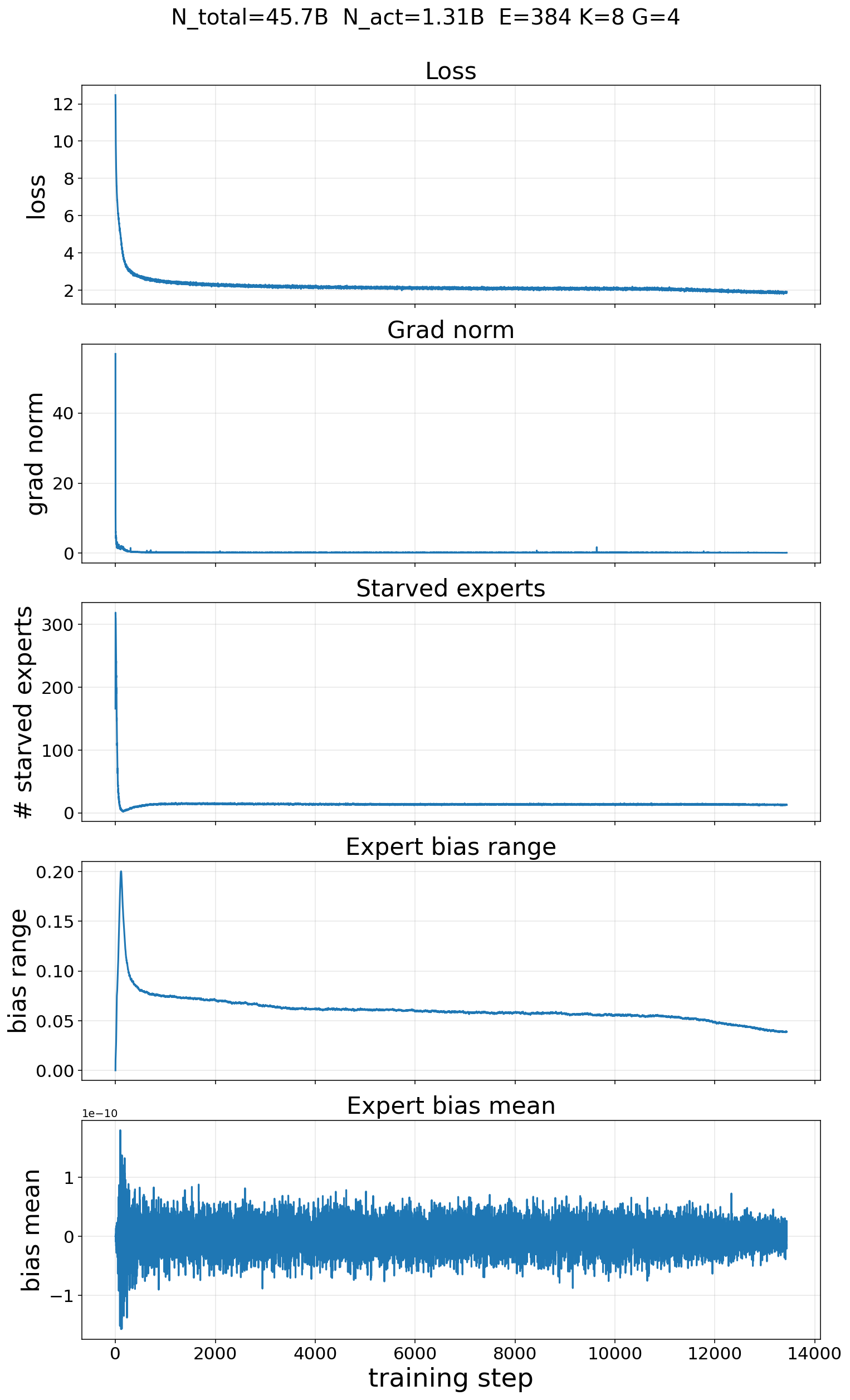}
\end{subfigure}
\caption{Training dynamics of the four largest MoE models in our scaling run sweep
(annotated with $N_{\text{tot}}$, $N_{\text{act}}$, $E$, $K$, $G$). Each
panel stacks, over training steps, the loss, gradient norm, starved-expert count, and per-expert bias range and mean. Loss descends
smoothly, starved experts vanish early, and the bias range contracts,
showing stable and balanced routing at scale.}
\label{fig:train-dynamics}
\end{figure}

Figure~\ref{fig:train-dynamics} shows the training dynamics of the four
largest MoE models in our scaling runs. For each run we plot, over
training steps, the loss, the gradient norm, the number of starved
(under-utilized) experts, and the range and mean of the per-expert
routing bias. We call an expert ``\textit{starved}" when it receives fewer than $5\%$ of the
tokens in a batch that a uniform assignment over the $E$ experts would send it, so
the reported count ``starved experts" is the number of such experts in an MoE layer averaged over all layers. Across all four,
the loss descends smoothly, the starved-
expert count collapses to near zero early in training and stays there,
and the expert-bias range contracts steadily, indicating that the
load-balancing controller of Section~\ref{sec:routing} keeps routing
stable and balanced throughout training even at the largest scales.

\section{MOSAIC Optimization}
\label{app:hw-optimization}


\subsection{Sparse-MoE Execution-Plan Realization}
\label{app:plan-realization}

For the Megatron-Core stack used in our experiments, the dense layers factor
the world size as
$N_{\text{dev}} = P_{\text{TP}}\,P_{\text{CP}}\,P_{\text{PP}}\,P_{\text{DP}}$,
the factorized device mesh of Megatron-LM \citep{shoeybi2019megatron} and
PyTorch \citep{liang2024torchtitan}, with the degrees running over tensor,
context, pipeline, and data parallel groups. Within an MoE layer, the dense
data-parallel domain is repartitioned into expert parallelism and expert data
parallelism,
\begin{equation}
P_{\text{DP}}
\;=\;
P_{\text{EP}}\,P_{\text{EDP}},
\qquad
P_{\text{EDP}}
\;=\;
\frac{P_{\text{DP}}}{P_{\text{EP}}}
\label{eq:mesh-moe}
\end{equation}
which requires that $P_{\text{EP}}$ divides $P_{\text{DP}}$. Equivalently, the
MoE-layer mesh factors as
$N_{\text{dev}} = P_{\text{TP}}\,P_{\text{CP}}\,P_{\text{PP}}\,
P_{\text{EP}}\,P_{\text{EDP}}$. Thus $P_{\text{EP}}$ is an input degree of the
plan that participates in the MoE-layer device factorization, but it does not
multiply the dense-layer $P_{\text{DP}}$ axis independently, and the expert
data parallelism $P_{\text{EDP}}$ is determined rather than chosen. These
constraints, together with the discrete micro-batch and
activation-checkpointing choices, define the realizable plan set
$\mathcal{P}_{\text{disc}}(N_{\text{dev}})$ of Section~\ref{sec:mosiac-formal}.

\subsection{Structural Properties}
\label{sec:minlp-structure}

Three properties of Equation~\eqref{eq:hw-minlp} are directly useful for interpreting the result. 

\paragraph{Property 1 (objective convexity).} In the continuous variables
$(\log N_{\text{tot}}, \log D, \log(1-S), \log G) \in \mathbb{R}^{4}$,
the loss
$\mathcal{L}(N_{\text{tot}}, S, D, G)$ of Equation~\eqref{eq:joint-law} is convex, since
$\mathcal{L}$ is a posynomial in
$(N_{\text{tot}}, D, 1{-}S, G)$~\citep{boyd2007tutorial} on the positive
orthant and posynomials are log-convex in these coordinates.

\paragraph{Property 2 (constraint-driven non-convexity).} The non-convexity
of Equation~\eqref{eq:hw-minlp} resides in the constraints, not in the objective.

\paragraph{Property 3 (decoupling of $D$ from layout).} Tokens $D$ enter
only the objective and the FLOPs budget, never the memory cap or the
layout product. Consequently, for any $Z$ and any
$P_{\ell} \in \mathcal{P}_{\text{feas}}(Z, B)$, the largest
admissible $D$ is the closed-form
\begin{equation}
D^{\star}(Z, P_{\ell})
\;=\;
\frac{F_{\text{peak}}\, N_{\text{dev}}\, T_{\text{train}}\,
      \mathrm{MFU}(Z, P_{\ell}, B)\,
      \eta_{\text{good}}(N_{\text{dev}})}
     {6\, N_{\text{act}}(N_{\text{tot}}, S)}
\label{eq:Dstar}
\end{equation}
and since $\mathcal{L}$ is monotone-decreasing in $D$, any optimum
saturates Equation~\eqref{eq:c-flops} at $D = D^{\star}(Z, P_{\ell})$. This eliminates one
continuous decision variable analytically and reduces the inner problem to
a search over $(Z, P_{\ell})$ alone.

\paragraph{Algorithmic consequence.} Properties~1\textendash3 are what let the
bilevel program of Equation~\eqref{eq:hw-minlp} be solved exactly by the
single-stage search of Section~\ref{sec:results}, with no generic MINLP
machinery. The decision space is finitely realizable, since a deployable
architecture ranges over the discrete ladder geometries
$(L_{\text{layers}}, d, E, K, G)$ and an execution plan over small grids of
parallelism degrees, micro-batch sizes, and checkpointing modes, and a
fractional expert count or parallel degree is not something one can launch.
Tokens then drop out analytically. Loss decreases monotonically in $D$, so an
optimal run never leaves wall-clock budget unused and
Equation~\eqref{eq:Dstar} fixes the saturating $D^{\star}$ in closed form,
removing the only continuous decision variable. The deployed search of
Section~\ref{sec:results} realizes this on the finite grid
$\mathrm{TPP} \in [40, 150]$, taking for each candidate the largest ratio the
budget supports, so $D^{\star}$ is reached up to grid resolution and the cap
binds only for the smallest models. What remains splits cleanly,
because at fixed $Z$ the best execution plan is simply the feasible plan of
maximum MFU within Equation~\eqref{eq:p-feas}, so the scaling
law never participates in the layout search.

MOSAIC therefore reduces to enumerating the ladder geometries, selecting the
max-MFU feasible plan for each, fixing $D$ from Equation~\eqref{eq:Dstar}, and scoring
the loss with Equation~\eqref{eq:joint-law}, and the configuration it returns is the
exact optimum of Equation~\eqref{eq:hw-minlp} over the realizable grid rather
than a heuristic solution. A generic MINLP solver would in any case have
little to work with here, since $\mathrm{MFU}$ and $\mathrm{Mem}$ come out of
the calibrated performance model of Section~\ref{sec:performance-model} as
measured kernel and collective lookup tables rather than symbolic expressions
that relaxation or branch-and-bound machinery can bound. Searching the
discrete geometries directly, rather than relaxing to a continuum and
projecting back, is what returns the loss-versus-wall-clock frontier that
practitioners want.

\subsection{Optimal Resource Allocation}
\label{app:allocation}
MOSAIC solves the allocation problem numerically by enumerating realizable
architectures and searching over their feasible execution plans. This section asks
what an additional closed-form prescription in the style of
\citet{hoffmann2022training} would look like, and what assumptions would be required
to state one. Nothing below replaces the discrete search of
Section~\ref{sec:results}.

\subsubsection{Reduction to a $C_\text{peak}$ Law}
\label{app:alloc-reduction}
The exogenous hardware-time budget is the raw ceiling
$C_{\text{peak}} = F_{\text{peak}}N_{\text{dev}}T_{\text{train}}$ of
Section~\ref{sec:why-flops-is-not-time} together with the goodput fraction
$\eta_{\text{good}}(N_{\text{dev}})$. Both are fixed before an architecture is
selected, so $C_{\text{peak}}\eta_{\text{good}}$ is the counterpart of the compute
budget in a Chinchilla-style prescription, whereas the deliverable budget $C_\text{deliverable}$ is not,
since it also carries the best attainable
$\operatorname{MFU}^{\star}(Z) = \max_{P_{\ell}\in\mathcal{P}_{\text{feas}}(Z, B)}
\operatorname{MFU}(Z,P_{\ell},B)$ over the feasible plan set
$\mathcal{P}_{\text{feas}}(Z, B)$ of Section~\ref{sec:mosiac-formal}. Loss decreases monotonically in the token count, so
saturating Equation~\eqref{eq:c-flops} recovers Property~3,
\begin{equation}
D^{\star}(Z,C_{\text{peak}})
= \frac{C_{\text{peak}}\,\operatorname{MFU}^{\star}(Z)\,\eta_{\text{good}}}
       {6N_{\text{act}}(Z)}.
\label{eq:alloc-D}
\end{equation}
Since $\operatorname{MFU}^{\star}(Z)$ does not depend on the training-window
length, substituting Equation~\eqref{eq:alloc-D} into
Equation~\eqref{eq:joint-law} collapses the law to
\begin{equation}
\mathcal{L}^{\star}(Z,C_{\text{peak}})
= \mathcal{L}_{\infty}(Z) + \frac{\Psi(Z)}{C_{\text{peak}}^{\beta}},
\qquad
\Psi(Z) = b\left(\frac{6N_{\text{act}}(Z)}
{\operatorname{MFU}^{\star}(Z)\,\eta_{\text{good}}}\right)^{\!\beta}
\label{eq:alloc-reduction}
\end{equation}
where $\mathcal{L}_{\infty}(Z)$ collects the three terms of
Equation~\eqref{eq:joint-law} that do not involve $D$, together with the
irreducible constant $e$. We note that $D^{\star} \propto C_{\text{peak}}$ at
any fixed architecture, provided no external data or tokens-per-parameter cap
binds, though the globally optimal token count doesn't need to be linear in
$C_{\text{peak}}$ because the selected architecture can change the constant
$\operatorname{MFU}^{\star}(Z)/N_{\text{act}}(Z)$.


\subsubsection{Conditions for an allocation law}
The reduction above is exact for each discrete candidate, but a continuous
optimal solution needs more assumptions, and what follows is a local,
regime-specific approximation rather than a universal law. We write
$r \equiv 1-S$ and $N \equiv N_{\text{tot}}$, and we assume (1) an interior
memory-slack region where any memory or geometry boundary constraint is non-existent, (2) the same
execution layout stays optimal locally, and (3) 
$\operatorname{MFU}^{\star}(N,r,G)$ is differentiable. We define
the local MFU log-slopes
$\omega_N = \partial\log\operatorname{MFU}^{\star}/\partial\log N$,
$\omega_S = \partial\log\operatorname{MFU}^{\star}/\partial\log r$, and
$\omega_G = -\,\partial\log\operatorname{MFU}^{\star}/\partial\log G$, so that
$\omega_G > 0$ means MFU falls as experts become finer. After writing the four terms of
Equation~\eqref{eq:joint-law} as $\mathcal{L}_{\text{cap}} = aN^{-\alpha}$,
$\mathcal{L}_{\text{data}} = bD^{-\beta}$, $\mathcal{L}_{\text{sp}} = cr^{-\lambda}$
and $\mathcal{L}_{\text{int}} = jN^{-\gamma}r^{-\delta}G^{-\eta}$, and eliminating
$D$ through Equation~\eqref{eq:alloc-D}, stationarity in
$(\log N, \log r, \log G)$, we obtain
\begin{align}
\alpha\,\mathcal{L}_{\text{cap}}
&= \beta\,\mathcal{L}_{\text{data}}\bigl(1-\omega_N-\gamma\omega_G/\eta\bigr),
\notag\\
\lambda\,\mathcal{L}_{\text{sp}}
&= \beta\,\mathcal{L}_{\text{data}}\bigl(1-\omega_S-\delta\omega_G/\eta\bigr),
\qquad
\eta\,\mathcal{L}_{\text{int}}
= \beta\,\omega_G\,\mathcal{L}_{\text{data}}.
\label{eq:alloc-foc}
\end{align}
Since all four loss components and their exponents are positive, these
stationarity equations require necessary sign conditions for an interior
solution. In particular, the $G$ equation requires $\omega_G>0$, while
the $N$ and $r$ equations require
$1-\omega_N-\gamma\omega_G/\eta>0$ and
$1-\omega_S-\delta\omega_G/\eta>0$, respectively.
Without a systems penalty on finer experts, $\omega_G=0$, the third
stationarity condition cannot hold. The objective is then strictly decreasing
in $G$ within this regime, so no interior $G$ optimum exists and the solution
lies at the largest admissible expert split factor. Where memory binds, a geometry
boundary is active, or the optimal layout changes, the corresponding KKT or
nonsmooth terms must be included instead.

Note that Equation~\eqref{eq:alloc-foc} alone does not give a power law, since the
log-slopes may drift as the architecture moves. The key additional assumption
is that they stay approximately constant within one systems regime, so that
$\operatorname{MFU}^{\star} \simeq m_0 (N/N_0)^{\omega_N} (r/r_0)^{\omega_S}
(G/G_0)^{-\omega_G}$. Under this approximation the four terms stay in fixed
ratios and the optimum traces the regime-local path
\begin{equation}
\boxed{\;
r^{\star} \propto \left(N^{\star}\right)^{\alpha/\lambda},
\quad
D^{\star} \propto \left(N^{\star}\right)^{\alpha/\beta},
\quad
G^{\star} \propto \left(N^{\star}\right)^{(\alpha-\gamma-\alpha\delta/\lambda)/\eta}
\;}
\label{eq:alloc-path}
\end{equation}
Writing $p = \alpha/\lambda$, $\varrho = \alpha/\beta$ and
$g = (\alpha-\gamma-\alpha\delta/\lambda)/\eta$ for these exponents, and
$\nu = [(1-\omega_N) + (1-\omega_S)p + \varrho + \omega_G g]^{-1}$, the same
path in terms of the purchased budget is
\begin{equation}
\boxed{\;
N^{\star} \propto C_{\text{peak}}^{\nu},
\quad
1-S^{\star} \propto C_{\text{peak}}^{p\nu},
\quad
D^{\star} \propto C_{\text{peak}}^{\varrho\nu},
\quad
G^{\star} \propto C_{\text{peak}}^{g\nu}.
\;}
\label{eq:alloc-budget-path}
\end{equation}
These relations hold conditional on fixed hardware, device count, batch envelope,
goodput model, and systems regime. Layout transitions, memory boundaries, and
topology changes alter the log-slopes and therefore produce different allocation
phases.

\section{Performance Model Input/Outputs}
\label{app:performance-model-input}
The listings below describe ScalePlan (\url{https://github.com/dmlc/ScalePlan}), the performance model released with this paper.
Beyond scoring one plan, it enumerates the execution plans $P_{\ell} = (P_{\text{TP}},
P_{\text{EP}}, P_{\text{CP}}, P_{\text{PP}}, P_{\text{DP}}, B_{\text{micro}},
A_{\text{ckpt}})$ for a
given device count, discards those that breach the memory cap, and returns the survivors ranked
by predicted MFU. That search is what supplies the inner maximization over $\mathcal{P}_{\text{feas}}$ of
Equation~\eqref{eq:c-flops}, so MOSAIC calls it once per candidate architecture
rather than assuming a layout.
The field names below are those of the tool, and the trailing comment on each line
gives the corresponding paper symbol where one exists.
\begin{lstlisting}[style=yamlstyle,caption={Performance model input specification.},captionpos=b,label={lst:perf-model-input}]
model:
  n_layers: 48          # L_layers, transformer blocks
  hidden_sz: 6144       # d, residual-stream width
  inter_sz: 15360       # d_ff, dense FFN width (2.5 d)
  n_q_heads: 48         # n_head, query heads
  n_kv_heads: 8         # key/value heads (grouped-query attention)
  head_dim: 128         # d_head, so d = n_head * d_head
  vocab_sz: 257536      # vocabulary size
  precision: "bf16"
  sdpa_precision: "bf16"
  glu: true
  rotary_embeds: true
  dropout: false
  tie_embeddings: false

  moe:
    n_experts: 128      # E, routed experts per MoE block
    experts_per_token: 3  # K, top-K active experts
    capacity_factor: 1.0
    expert_inter_sz: 3840 # d_expert, so G = d_ff / d_expert = 4
    moe_frequency: 1
    expert_tp_degree: 1 # P_ETP, expert tensor parallel degree

search:
  num_devices: 512      # N_dev

performance:
  activation_checkpointing_type: full

optimizer:
  optimizer_type: muon

data:
  gbs: 8192             # B_global, sequences per optimizer step
  seqlen: 8192          # T_seq, sequence length
  microbatch_sz: 2      # B_micro

hardware:
  node_type: p6-b200.48xlarge  # 8 GPUs per node, so N_gpu/node = 8
\end{lstlisting}
Output example, where the layout degrees are
$(P_{\text{TP}}, P_{\text{EP}}, P_{\text{PP}}, P_{\text{CP}}, P_{\text{DP}})$ of
$P_{\ell}$ and the reported MFU and memory are the
$\mathrm{MFU}(Z, P_{\ell}, B)$ and $\mathrm{Mem}(Z, P_{\ell}, B)$ predictions of
Section~\ref{sec:performance-model}.

\begin{lstlisting}[style=yamlstyle,caption={Performance model output.},captionpos=b,label={lst:perf-model-output}]
Rank 1:
  Configuration: tp=1, ep=8, pp=16, cp=1, dp=32
  Analytical MFU: 13.69%
  Memory per device: 21.30 GB
  Iteration time: 44.892 s
\end{lstlisting}
\section{Results}
\label{app:results}

\subsection{Geometry Scale Ladders}

Table~\ref{tab:scale-ladder} lists the rungs of the geometry scaling ladder of
Section~\ref{sec:geometry-ladder} that our search enumerates. Each row is one
seed $q$, giving the backbone dimensions the seed fixes and, at a representative
MoE configuration, the parameter counts and sparsity that backbone realizes.

\begin{table}[h]
\centering
\small
\setlength{\tabcolsep}{6pt}
\begin{tabular}{@{}rrrrrrrr@{}}
\toprule
$q$ & $L_{\text{layers}}$ & $n_{\text{head}}$ & $d$ & $d_{\text{ff}}$ & $N_{\text{tot}}$ (B)$^{\ast}$ & $N_{\text{act}}$ (B)$^{\ast}$ & $S$ \\
\midrule
 4 & 16 & 16 & 2048 &  5120 &  15.5 &  0.7 & 0.954 \\
 6 & 24 & 24 & 3072 &  7680 &  53.2 &  2.3 & 0.957 \\
 8 & 32 & 32 & 4096 & 10240 & 127.3 &  5.4 & 0.958 \\
10 & 40 & 40 & 5120 & 12800 & 250.0 & 10.4 & 0.958 \\
12 & 48 & 48 & 6144 & 15360 & 433.7 & 17.9 & 0.959 \\
14 & 56 & 56 & 7168 & 17920 & 690.5 & 28.2 & 0.959 \\
16 & 64 & 64 & 8192 & 20480 & 1032.8 & 41.9 & 0.959 \\
18 & 72 & 72 & 9216 & 23040 & 1472.8 & 59.5 & 0.960 \\
\bottomrule
\end{tabular}
\caption{Seeds of the model-scale ladder used in our experiments. The geometry seed $q$ sets the backbone
depth $L_{\text{layers}} = 4q$, head count $n_{\text{head}} = 4q$, residual width
$d = 128\,n_{\text{head}}$, and reference dense FFN width
$d_{\text{ff}} = 2.5\,d$. The last three columns give the total and
active parameter counts and the induced sparsity $S$
at an example MoE configuration $E\!=\!128$, $K\!=\!3$, $G\!=\!4$. The
MoE axes $(E, K, G)$ are searched at each rung.
$^{\ast}$Non-embedding parameter counts.}
\label{tab:scale-ladder}
\end{table}

\subsection{Implications for Sparse-MoE Parallelism Design}
\label{app:moe-implications}

This appendix develops the design rules summarized in
Section~\ref{sec:moe-implications}. Reading the measured best MFU achieved by
different parallelism configurations in the validation sweeps yields two
design rules and one caveat that bounds them.
\begin{itemize}
  \item \textbf{The dominant lever is not any single degree but the pipeline
  fill ratio $r_{\text{fill}} = n_{\text{mb}} / P_{\text{PP}}$, the microbatches in flight
  per pipeline stage.}
  Within a sweep, layout choice moves measured MFU by up to $3.1\times$ (18B)
  and $2.1\times$ (700M), and almost all of that spread tracks $r_{\text{fill}}$, since
  1F1B leaves each stage idle for a bubble fraction of
  $(P_{\text{PP}}{-}1)/(P_{\text{VPP}}\,n_{\text{mb}})$, so throughput collapses as
  $r_{\text{fill}}$ falls toward $1$ and
  saturates once $r_{\text{fill}}$ is large. The 18B pipeline sweep makes this explicit,
  where at $P_{\text{PP}}{=}16$, MFU climbs as $r_{\text{fill}}$ goes
  $0.25 \to 0.5 \to 1$,
  then plateaus at $\sim\!10$\textendash$12\%$ for $r_{\text{fill}} \ge 4$. This reframes the two
  parallelism knobs that people tune independently, since deeper pipelines and larger
  microbatches both \emph{shrink} $r_{\text{fill}}$ and are only free once the fill ratio is
  already comfortable. Concretely, every extra pipeline stage or microbatch that
  drops $r_{\text{fill}}$ below $\sim\!4$ costs measurable MFU, whereas the same change
  at $r_{\text{fill}} \ge 4$ is nearly free.
  \item \textbf{Keep the expert all-to-all inside the scale-up (NVLink) domain,
  and, once the fill ratio is satisfied, pipeline depth is what should absorb
  the rest of the budget.} Expert parallelism helps up to the node boundary and
  hurts past it, where MFU rises monotonically with $P_{\text{EP}}$ in the 700M sweep
  (for $P_{\text{EP}} \in \{1,2,4,8\}$ at fixed depth) and
  drops from $P_{\text{EP}}{=}8$ to $P_{\text{EP}}{=}16$ in both 5.4B sweeps as the group
  leaves the 8-GPU node.
  Once $P_{\text{EP}}$ is pinned to the domain,
  moderate pipeline depth is the cheapest remaining axis provided it
  does not push $r_{\text{fill}}$ below the plateau, and the 5.4B optima use
  $P_{\text{PP}}{\in}\{2,4\}$
  because at these depths $r_{\text{fill}}$ stays $\ge 8$. This is the
  actionable ordering the model exposes, where we set $P_{\text{EP}}$ to the scale-up
  domain, then add pipeline depth only while $r_{\text{fill}}$ remains large.
  \item \textbf{Spend the FFN-parallelism budget on expert parallelism rather
  than expert tensor parallelism.} For the expert FFN this is the expected ordering, since
  expert parallelism
  partitions the experts with one all-to-all per layer, whereas tensor
  parallelism adds two all-reduces per layer on the critical path.
\end{itemize}
Scoped this way, the rules collapse the viable search region sharply and in a
model-checkable order, where we fix $P_{\text{TP}}{=}1$, set $P_{\text{EP}}$ to the
scale-up domain, then choose the
$(P_{\text{PP}}, B_{\text{micro}})$ pair that keeps the pipeline fill ratio
$r_{\text{fill}} \ge 4$ at the target
global batch. This retains the measured optimum in every sweep that is not
memory-bound while discarding the configurations the analytical model also ranks
worst, which lowers the cost of the hardware-confirmation step. Long-context,
memory-bound, or wider-NVLink regimes shift the thresholds along the
principles above.

\subsection{The Training Stack Behind the Staged Runs}
\label{app:training-stack}

The staged runs of Section~\ref{sec:results-staged} are launched on the
same discrete grid that Stage-2 enumerates, but their realized throughput
is set by a production training stack whose overheads the analytical
performance model only approximates. Each run is driven by a PyTorch
Lightning, NeMo, and Megatron-Core pipeline with one process per GPU and
eight GPUs per p6-b200 node. We use the Megatron-Core MoE training
stack\footnote{https://github.com/NVIDIA/Megatron-LM/tree/main/megatron/core/transformer/moe}
as is, both for the performance-model validation and for the staged
runs, so we measure a standard stack and not a custom one. From that stack we enable the
grouped-GEMM expert path (so the per-expert matrix multiplications for a
layer are issued as one batched kernel), cross-entropy loss fusion, the
all-to-all token dispatcher for expert parallelism, and shared-expert
overlap (which hides the shared-expert compute behind the routed
all-to-all). Fine-grained activation recomputation is treated as a sweep
axis inside the MFU search rather than fixed, since it trades memory for
recompute cost and thereby moves the feasible-layout boundary. We
deliberately leave the rest of the stack's kernels and fusions off, and
in particular we do not use parallel folding or DeepEP, and we leave many of the newer optimizations in the stack as future work. An important point is that our performance model is itself an evolving artifact that mirrors one systems stack at a given point in time, so as we turn on new kernels or change the stack, the model is simply re-calibrated or replaced by an updated one rather than being a fixed assumption of the framework.

Several loop-level effects open a gap between this realized MFU and the
value the analytical model predicts, and they are the reason the analytical model at times is systematically optimistic (Section~\ref{sec:results-staged}).
Pipeline parallelism over large number of micro-batches leaves an
idle bubble that a steady-state FLOPs
count does not see. Full activation recomputation, used to fit the model
in memory, adds an extra forward pass per layer inside the backward,
raising step time without adding any of the idealized compute that MFU credits.
The expert all-to-all dispatch is exposed collective time whose cost
depends on whether an expert-parallel group stays inside a node or
crosses the fabric, and the fp32 router is a precision island inside an
otherwise bf16 model. Together with asynchronous S3 checkpointing,
periodic garbage collection, and launch jitter, these are wall-clock the
device-side model does not fully charge, so it over-predicts MFU while
preserving the ranking the selection relies on.

A second class of overhead lives entirely on the host. Our tokenizer is a
fast Hugging Face tokenizer, downloaded once per node from S3 into shared memory
and patched to be thread-safe because the loader calls it from worker
threads. The collator pads every micro-batch to a multiple of eight and
then to the full $2048$-token context, since pipeline parallelism
requires uniform shapes across stages. The streaming dataloader shards
only along the data-parallel axis, replicates identical batches across the
tensor-, pipeline-, and expert-parallel ranks of a group. Because we run
a single loader worker per rank with a prefetch depth of two, host-side
tokenization, collation, and the host-to-device copy can stall the
accelerators. Such input-pipeline stalls lower realized MFU and are
invisible to a device-side performance model, which is one more reason in the gap between the
measured throughput and the prediction.

Finally, the realization is specific to the accelerator and its
interconnect, so the same architecture can land at a different optimum on
different hardware. A p6-b200
node\footnote{https://aws.amazon.com/ec2/instance-types/p6/} pairs eight
NVIDIA B200 GPUs with up to $14.4$\,TB/s of
bidirectional intra-node NVLink and a fourth-generation Elastic Fabric
Adapter delivering $3.2$\,Tbps ($400$\,Gbps per GPU) between nodes. On an
H200 platform the per-device peak FLOPs, the HBM capacity, and the NVLink
bandwidth are all lower, which shifts both the memory cap and the
throughput-optimal layout. Because intra-node collectives ride NVLink and
NVSwitch while inter-node collectives ride EFA RDMA, whether the expert
all-to-all is confined to a node or spills across the fabric can swing MFU
sharply. The optimum MOSAIC returns is therefore tied to the device
peak FLOPs and the bandwidth constants of the target platform, and it must
be resolved, not merely rescaled, when the hardware changes.

\vskip 0.2in

\end{document}